\documentclass[acmtog, nonacm]{acmart}
\pdfoutput=1

\usepackage{multirow}
\usepackage{longtable}
\usepackage{lscape}

\AtBeginDocument{%
  \providecommand\BibTeX{{%
    \normalfont B\kern-0.5em{\scshape i\kern-0.25em b}\kern-0.8em\TeX}}}

\setcopyright{acmcopyright}
\copyrightyear{2020}
\acmYear{2020}
\acmDOI{10.1145/1122445.1122456}

\acmJournal{TOG}
\acmVolume{37}
\acmNumber{4}
\acmArticle{111}
\acmMonth{8}

\begin{document}

\title{A survey of joint intent detection and slot-filling models in natural language understanding}

\author{Henry Weld}
\email{hwel4188@uni.sydney.edu.au}
\affiliation{
  \institution{The University of Sydney}
  \city{Sydney}
  \country{Australia}
}

\author{Xiaoqi Huang}
\email{xhua7314@uni.sydney.edu.au}
\affiliation{
  \institution{The University of Sydney}
  \city{Sydney}
  \country{Australia}
 }

\author{Siqu Long}
\email{slon6753@uni.sydney.edu.au}
\affiliation{
  \institution{The University of Sydney}
  \city{Sydney}
  \country{Australia}
}

\author{Josiah Poon}
\email{Josiah.Poon@sydney.edu.au}
\affiliation{
  \institution{The University of Sydney}
  \city{Sydney}
  \country{Australia}
}

\author{Soyeon Caren Han}
\email{caren.han@sydney.edu.au}
\affiliation{
  \institution{The University of Sydney}
  \city{Sydney}
  \country{Australia}
}

\begin{abstract}
Intent classification and slot filling are two critical tasks for natural language understanding. Traditionally the two tasks have been deemed to proceed independently. However, more recently, joint models for intent classification and slot filling have achieved state-of-the-art performance, and have proved that there exists a strong relationship between the two tasks.
This article is a compilation of past work in natural language understanding, especially joint intent classification and slot filling. We observe three milestones in this research so far:
Intent detection to identify the speaker’s intention, slot filling to label each word token in the speech/text, and finally, joint intent classification and slot filling tasks. In this article, we describe trends, approaches, issues, data sets, evaluation metrics in intent classification and slot filling. We also discuss representative performance values, describe shared tasks, and provide pointers to future work, as given in prior works. To interpret the state-of-the-art trends, we provide multiple tables that describe and summarise past research along different dimensions, including the types of features, base approaches, and dataset domain used. 

\end{abstract}




\maketitle

\section{Introduction} \label{sec introduction}
The efficacy of virtual assistants becomes more important as their popularity rises. Central to their performance is the ability for the electronic assistant to understand what the human user is saying, in order to act, or reply, in a way that meaningfully satisfies the requester. 

The human-device interface may be text based, but is now most frequently voice, and will probably in the near future include image or video. To put the understanding of human utterances within a framework, within the natural language processing (NLP) stack lies spoken language understanding (SLU). SLU starts with automatic speech recognition (ASR), the task of taking the sound waves or images of expressed language, and transcribing to text. Natural language understanding (NLU) then takes the text and extracts the semantics for use in further processes - information gathering, question answering, dialogue management, request fulfilment, and so on.

The concept of a hierarchical semantic frame has developed to represent the levels of meaning within spoken utterances. At the highest level is a domain, then intent and then slots. The domain is the area of information the utterance is concerned with. The intent (a.k.a. goal in early papers) is the speaker's desired outcome from the utterance. The slots are the types of the words or spans of words in the utterance that contain semantic information relevant to the fulfilment of the intent. An example is given in Table \ref{table:frameeg} for the domain \textit{movies}. Within this domain the example has intent \textit{find\_movie} and the individual tokens are labelled with their slot tag using the IOB tagging format. 

\begin{table}[t]
\begin{tabular}{|l|l|l|l|l|l|l|}
\hline
\textbf{query}  & find & recent & comedies & by & james & cameron  \\ \hline
\textbf{slots}  & O    & B-date & B-genre   & O  & B-dir  & I-dir\\ \hline
\textbf{intent} & \multicolumn{6}{c|}{find\_movie} \\ \hline
\textbf{domain} & \multicolumn{6}{c|}{movies}     \\ \hline
\end{tabular}
\caption{An example of an utterance as semantic frame with domain, intent and IOB slot annotation (from \cite{hakkani2016multi})}
\label{table:frameeg}
\end{table}

The NLU task is thus the extraction of the semantic frame elements from the utterance. NLU is important - it is central to devices that desire a spoken interface with humans - for example, conversational agents, instruction in vehicles (driverless or otherwise), Internet of Things (IoT), personal assistants, online helpdesks/chatbots, robot instruction, and so on. Improving the quality of the semantic detection will improve the quality of the experience for the user, and from here it draws its importance and popularity as a research topic.

In many data sets, and indeed real world applications, the domain is limited; it is concerned only with hotel bookings, or air flight information, for example. In these cases the domain level is generally not part the analysis. However in wider ranging applications, for example the SNIPS data set discussed later, or the manifold personal voice assistants which are expected to field requests from various domains, inclusion of the domain detection in the problem can lead to better results. However, for the purposes of this survey we will treat the domain as ancillary.

This leaves us with intent and slot identification. What does the human user want from the communication, and what semantic entities carry the details? The two sub-tasks are known as intent detection and slot filling. The latter may be a misnomer as the task is more correctly slot labelling, or slot tagging. Slot filling is more precisely giving the slot a value of a type matching the label. For example, a slot labelled ``B-city'' could be filled with the value ``Sydney''. Intent detection is usually approached as a supervised classification task, mapping the entire input sentence to an element of a finite set of classes. Slot filling then is a labelling of the sequence of tokens in the utterance, making it within the sequence-to-sequence (seq2seq) class of problems. 

While early research looked at the tasks separately, or put them in a series pipeline, it was quickly noted that the slot labels present and the intent class should and do influence each other in ways that solving the two tasks simultaneously should garner better results for both tasks. This has been the at the centre of NLU over recent years though work on the single tasks has continued.

A joint model which simultaneously addresses each sub-task must, to be successful, capture the joint distributions of intent and slot labels, with respect also to the words in the utterance, their local context, and the global context in the sentence. A joint model has the advantage over pipeline models that it is less susceptible to error propagation, and over separate models in general that there is a only a single model to train and fine tune.
 
A drawback is that a large annotated corpus is usually required, though this is also true of separate models. The model may also be relatively complicated and take time to train. It has also been observed that joint models may not generalise well to unseen data, due to the variety of natural language expressions of similar intent. In real world applications the domains and label sets may change over time.

In many ways the development of the field has followed a similar path to other areas of NLP, starting with classical (statistical or probabilistic) models. Neural networks were applied as computing power increased. In particular, due to the sequential nature of the slot labelling sub-task, recurrent neural networks (RNNs) have been a technology frequently used in the field. In more recent years the transformer architecture has debuted to address issues like long range dependency. As a result, attention has increased in importance. As far as feature creation goes, convolution, word embeddings, and pre-trained language models have all been applied, amongst many other methods. The use of external knowledge bases has been observed in more recent papers.

The most regularly used data sets are two freely available sets - ATIS and SNIPS. A common experiment is implied by the literature, from which the reported results are compared in this survey. In addition, one of the aims of this survey is to address further standardisation, in terms of the parameters of the experiment and the evaluation metrics used.

The approaches to the joint task have been manifold and have shown excellent results in standard supervised training/test experiments. As new techniques make what may appear to be incremental increases to the state of the art it is perhaps time to recast the measures of success in the field. Rather than just developing new, more challenging annotated data sets, increasingly important must be the development of unstructured semantic detection in new domains.

The motivation for this survey is to take stock of the state of the field in 2020 following a surge of ideas and approaches over recent years, particularly in the joint task. We collect information on the approaches pursued so far and the issues encountered and addressed. With the survey completed we propose some future directions for the field.

%

In summary, this survey address three major questions:
\begin{itemize}
    \item Q1: How do these joint models achieve and balance two aspects, intent classification and slot filling?
    \item Q2: Have syntactic clues/features been fully exploited or does semantics override this consideration?
    \item Q3: Can successful models in one supervised domain be made more generalisable to new domains or languages or unseen data?
\end{itemize}


\subsection{Scope}

The focus of this survey is on extraction of the intent and slots of single utterances. The separate tasks are covered and then the joint task is addressed in detail. Papers on the following aspects will be reviewed but the information considered as ancillary only:

\begin{itemize}
    \item Multi-domain data sets with annotated domain. Extraction of the domain is a classification task like intent detection, albeit less granular. Inclusion of the domain identification task may aid the two sub-tasks of interest and this will be mentioned;
    \item Dialogue action. Dialogue action is the identification of the next action to be taken by a dialogue management system once intent and slots have been identified. In some cases dialogue action is a direct substitute for intent and in others a mapping is made from intent and slot labels to the action. We review papers that include a strong focus on intent and slot detection;
    \item Multi-intent data sets. An utterance may have multiple intent (for example, flight booking and hotel booking). More work has been done in pure intent detection on this aspect than in the joint task. We will consider it in the pure intent task in particular and make comments on expansion to the joint task; 
    \item Automatic speech recognition (ASR). Some papers begin with the ASR step and look at error propagation from ASR to intent or slot prediction. We don't consider this aspect.
\end{itemize}

\begin{table*}[t]
\caption{Historical overview of intent detection papers}
\label{tab:overvew-intent}
\begin{tabular}{ccp{5.5cm}p{9cm}}

\toprule
\textbf{Year} & \textbf{\# papers} & \textbf{Feature engineering} & \textbf{Technologies}  \\
\midrule

2011 & 3 & Dependency parse & SVM, DBN, multi-layer NN, AdaBoost  \\

2012 & 1 & Bag-of-words & C4.5, RF, NB, KNN, Linear SVM  \\

2013 & 1 & n-gram & SVM, SVM-HMMs  \\

2015 & 4 & n-gram, word2vec, & LSTM, RNN, ensemble, RF, clustering, SVM, AdaBoost, NN, J48, FFN  \\

2016 & 1 & CNN & RF  \\

2018 & 7 & GloVe, word2vec, character embedding, grammatical features, dependency parse, knowledge base, POS, CRF, Regex, PCFG-ML, fastText & (Bi)CNN, (Bi)LSTM, (Bi)GRU, ensemble, Capsule networks, attention, (Bi)RNN, adversarial networks, gradient reversal layer, SVM, J48, Logistic regression, PPN, RF, Gaussian Naïve Bayes, KNN, NB, softmax regression \\

2019 & 4 & n-gram, character, word2vec, CNN, BiLSTM & BiLSTM, attention, Ridge, KNN, MLP, passive aggressive, RF, linear SVC, SGD, nearest centroid, multinominal NB, Bernoulli NB, K-means, CNN, BiGRU, density-based novelty detection algorithm, local outlier factor \\

2020 & 1 & BERT, word2vec, CNN & Siamese, triple loss \\

\bottomrule

\end{tabular}
\end{table*}

\subsection{Related surveys}

\cite{tur2011spoken} is a complete summary of the SLU field at the advent of the neural era (2011). \cite{wang2016recent} concentrate on models that jointly address sub-tasks in dialogue systems, including NLU, dialogue management (DM) and Natural Language Generation (NLG). They cover the early models in the joint task but predate the works explicitly tying NLU to dialogue action covered here.

\cite{tur2018deeplearning} concentrates on goal-oriented conversational language understanding but within that field provide an excellent precursor to this survey, covering the state of the art to 2017 in the two sub-tasks and 2016 in the joint task. \cite{hou2019taskoriented} give a small overview of the separate and joint tasks. \cite{liu2019intentdetection} provides a good survey of intent detection methods up to 2018 including multi-intent detection and evaluation methods.

Tangentially related surveys include \cite{serban2018corpora} which surveys dialogue data sets available for research, and \cite{deriu2020surveyevaluation} which gives an overview of evaluation methods for dialogue systems.

In this survey we bring the coverage of methods up to mid-2020 including the many applications of deep learning in the field. As well as a technological survey we look at issues addressed in each task and the joint task, and the approaches designed to address these issues. We also supply a summary of reported performance on the standard data sets.

\subsection{Structure of the survey}

The survey begins with a broad overview of the literature in Section \ref{sec overview}. We then give a detailed description of the methods for each sub-task (Sections \ref{sec intent} and \ref{sec slot}) and the joint task in Section \ref{sec joint}, along with the issues addressed and solutions proposed. In Section \ref{sec dataset} a survey of the data sets encountered takes place. In Sections \ref{sec metrics} and \ref{sec experiment} there is a description of the experiments and evaluation methods applied and a discussion of standardisation of these. A summary of the results achieved over the history of the field is given in Section \ref{sec performance}. We finish with a discussion of the challenges and opportunities for research in the field and give concluding remarks in Section \ref{sec conclusions}.

\begin{table*}[t]
\caption{Historical overview of slot labelling papers}
\label{tab:overvew-slot}
\begin{tabular}{ccp{5.5cm}p{9cm}}

\toprule
\textbf{Year} & \textbf{\# papers} & \textbf{Feature engineering} & \textbf{Technologies}  \\
\midrule

2011 & 1 & Neural network, observation feature vector & Deep learning, CRF  \\

2012 & 1 & n-gram, K-DCN & Kernel learning, deep learning, DCN, log-linear model  \\

2013 & 3 & Discriminative embedding, named entity, dependency parse, POS, SENNA, RNNLM, bag-of-words & DBN, RNN, RNN-LM  \\

2014 & 2 & RNN, lexicon feature & CRF, LSTM, regression model, deep learning  \\

2015 & 3 & Word embedding, named entity, word embedding & RNN, sampling approach, external memory  \\

2016 & 3 & Word embedding, context window, RNN, CNN & BiRNN, attention, LSTM, encoder-labeler, CNN  \\

2017 & 1 & Word embedding & (Bi)LSTM, encoder-decoder, focus mechanism, entity position-aware attention  \\

2018 & 5 & BiLSTM, word embedding, character, CNN, delexicalisation & CRF, MTL, segment tagging, NER, BiLSTM, attention, delexicalised sentence generation, DNN, reinforcement learning, GRU, pointer network  \\

2019 & 6 & Word embedding, web-data, expert feedback, contextual information, GloVe, POS, character, BERT & BiLSTM, BiGRU, different knowledge sources, context gate, MTL, CNN  \\

2020 & 2 & ResTDNN & Prior knowledge driven label embedding, CRF, TDNN, RNN  \\

\bottomrule

\end{tabular}
\end{table*}

\section{Overview of the literature} \label{sec overview}

Intent classification is a form of text classification where the text is a single sentence that comes from a spoken or written utterance. Much effort has been made to construct features which encapsulate the sentence, both semantically and syntactically, and the words within it. These features have been passed to classifiers from the suite of classical and, from 2011, deep learning methods, as outlined in Table \ref{tab:overvew-intent}. Issues around ambiguity, shortness of sentences, treatment of out-of-vocabulary words and emerging label sets are amongst those covered in the literature.

Slot tagging (see Table \ref{tab:overvew-slot}) is framed as a sequence labelling problem and in early years drew from methods for statistically modelling the dependencies within sequences, like conditional random fields (CRFs) and Hidden Markov models (HMMs). Around 2013 the strength of RNNs in this area had been observed and was applied to the task and developed over the ensuing years. Interestingly the use of CRFs returned, often as a post-RNN step, due to their efficacy at handling label dependency issues. As far as feature creation goes the general goal of the task is to use the semantic information within the words and various context windows from small to long-range within the sentence. Attention is used as one approach for eliciting useful context. Slot tagging has experimented with external knowledge bases for extra performance.

Methods used by both sub-tasks to extend their features include looking at meta-data from the data collection. Multi-task learning has also been used by both tasks to look for synergistic learning from other related tasks. Of course the joint task itself is an example of this synergistic approach. Both tasks have also considered methods for transfer learning to other languages and to data with new, unseen tag sets.

The two earliest papers (2008-9) addressing the joint task drew methods from classical NLP. Features were constructed from words, n-grams and suffixes, or from a semantic parsing of the utterances. A CRF or a support vector machine (SVM) was used for the analysis. In 2013 the first neural network was used though it really just constructed convolutional neural network (CNN) features for use in the CRF model from 2008. In 2014 a recursive neural network (RecNN), which works over trees, was applied to the dependency parse of the utterances. In 2015 the first completely neural network was devised, using a recurrent neural network (RNN, different to a recursive neural network) embedding of words, CNN representation of sentences, and a feed forward network (FFN) for the analysis.

By 2016 the RNN encoder-decoder architecture had been found to be useful for seq2seq tasks and started to make its impact in the joint task. Unidirectional and bi-directional Long Short Term Memory (LSTM) and Gated Recurrent Unit (GRU) cells were tested within circuits. Attention made its first appearance. On the input feature side K-SAN graphs were used as a knowledge base. 

In 2017 the field appeared to stay progress, with only character embedding being added to the input features and no improvement of performance results on the major data sets. Perhaps though, researchers were working on the many developments which exploded in 2018. Word embeddings were introduced - word2vec, GloVe and ELMo. The circuits were still largely RNN based. For new architectures a capsule neural network and bidirectional circuits were introduced. Here bidirectional refers to explicit influence paths through the circuit: intent2slot refers to intent information being used as part of slot prediction and slot2intent the opposite, slot information being used as part of intent prediction. 

In 2019 BERT debuted as a word embedding technology and ELMo fell away. More knowledge bases were used as input features. Work on pre-processing the data sets included delexicalisation, augmentation, and sparse word embeddings using a lasso method. In architecture RNN and attention continued to be used and CRF made a return to handle label dependency issues. Newly applied architectures included the transformer, and memory neural networks.

The indications from 2020 are that graph embeddings are being used more to capture slot-intent and word-slot-intent relationships.

\begin{table*}[t]
  \caption{Historical overview of joint task papers}
  \label{tab:overviewjoint}
  \begin{tabular}{ccp{5.5cm}p{9cm}}
    \toprule
    \textbf{Year} & \textbf{\# papers} & \textbf{Feature engineering} & \textbf{Technologies}  \\
    \midrule
    2008 & 1 & words/n-grams/suffixes & CRF  \\
    2009 & 1 & semantic tree & SVM \\
    2013 & 1 & CNN & CRF \\
    2014 & 1 & dependency parse & RecNN (diff to RNN) \\
    2015 & 1 & RNN words, CNN sentence, Bag of words & MLP \\    
    2016 & 6 & RNN, K-SAN & (Bi)LSTM, (Bi)GRU, encoder-decoder RNN, attention \\    
    2017 & 4 & character, word, CNN & BiLSTM \\ 
    2018 & 18 & word2vec, GloVe, ELMo, CNN sentence, attention sentence & BiLSTM, BiGRU, encoder-decoder RNN, Capsule NN, BiDirectional \\  
    2019 & 29 & BERT, GloVe, character, knowledge base (tuples), delexicalisation & memory NN, transformer, CRF, attention, BiDirectional  \\  
    2020 & 10 & BERT, Graph embedding & Graph S-LSTM, BiDirectional, GCN, Capsule\\ 
\bottomrule
\end{tabular}
\end{table*}

\begin{table*}[t]
\caption{Intent detection papers reviewed with addressed issue, approach and techniques}
\begin{tabular}{ p{3.5cm}p{4.9cm}p{8.4cm} }

\hline
\textbf{Paper} & \textbf{Addressed issue} & \textbf{Approach}  \\

\hline
\cite{gonzalez2011multifaceted} & Multi-faceted query intent prediction & Combined multifaceted (multi-label) intent classification \\ \hline

\cite{Sarikaya_2011} & Small/lack of labelled training data & Initialise FFN using DBN derived from trained stacked RBMs\\ \hline

\cite{tur2011sentence} & Short text query in web search & Simplified sentence structure as additional feature \\ \hline

\cite{Chen_2012} & Contextual/temporal information modeling & Semi-supervised co-training based on two independent features (text/metadata) \\ \hline

\cite{bhargava2013easy} & Small/lack of labelled training data & 1) Incorporating temporal information as additional feature; 2) Modeling temporal/session information as a sequence \\ \hline

\cite{ravuri2015recurrent} & OOV issue & Incorporating temporal information using RNN-based models with one-hot word embedding and n-gram hashing \\ \hline

\cite{Hasanuzzaman_2015} & Small/lack of labelled training data & Multi-objective ensemble learning with feature engineering (external resource used) \\ \hline

\cite{Purohit_2015} & Ambiguity in interpretation; Imbalanced data & Hybrid feature representation created by combining top-down processing using knowledge-guided patterns with bottom-up processing using a bag-of-tokens model \\ \hline

\cite{kanhabua2015learning} & Event-based web searching & Time-based and event-based clustering with click-through and standard statistical feature-based classification \\ \hline


\cite{hashemi2016query} & Complex feature engineering & CNN feature extracted vector representation \\ \hline

\cite{zhang2016mining} & Co-occurrence of words from different intents; word correlations addressing & Heterogeneous features of pairwise word correlation and POS information  \\ \hline

\cite{Firdaus_2018} & Exploring combination of deep learning architectures & Pre-trained embedding with ensemble of deep learning models \\ \hline

\cite{Xia_2018} & Emerging intents detection & Capsule-based architectures with zero-shot learning to discriminate emerging intents via knowledge transfer \\ \hline

\cite{costello2018multi} & Multi-domain/multi-lingual generalisation ability & Multi-layer ensemble models of different deep learning techniques \\ \hline

\cite{Masumura_2018} & Multi-task and multi-lingual joint modelling & Adversarial training method for the multi-task and multi-lingual joint modelling \\ \hline

\cite{Mohasseb_2018} & Grammar feature exploration & Grammar-based framework with 3 main features \\ \hline

\cite{Xie_2018} & Short text; Semantic feature expansion & Semantic Tag-empowered combined features \\ \hline

\cite{Qiu_2018} & Potential consciousness information mining & A similarity calculation method based on LSTM and a traditional machine learning method based on multi-feature extraction \\ \hline

\cite{kim2018ood} & OOD utterances & Multi-task learning \\ \hline

\cite{Cohan_2019} & Utilisation of naturally labelled data & Multitask learning based on joint loss \\ \hline

\cite{Shridhar_2019} & OOV issue; Small/lack of labelled training data & Subword semantic hashing \\ \hline

\cite{Wang_2019} & Learning of deep semantic information & Hybrid CNN and bidirectional GRU neural network with pre-trained embeddings (Char-CNN-BGRU) \\ \hline

\cite{Lin_2019} & Emerging intents detection & Maximise inter-class variance and minimise intra-class variance to get the discriminative feature \\ \hline

\cite{ren2020siamese} & Similar utterance with different intent & Triples of samples used for training \\ \hline

\cite{yilmaz2020kloos} & OOD utterances & KL divergence vector for classification \\ \hline

\hline
\end{tabular}
\end{table*}


\section{Intent Detection} \label{sec intent}

Intent detection is typically set up as a sentence classification problem. That is, a feature or features are constructed from the sentence and these are passed through a classification algorithm to predict a class for the sentence from a predefined set of classes. As a classification problem the techniques applied look to discover a well defined decision boundary between the features. Intent classification differs from classification tasks in other fields due to the nature of the data which are text sentences, coming from spoken language utterances. Hence, at least initially, the features should look to capture semantic information in the sentence. Beyond the semantic information within the words many approaches have been made to extend the feature set using internal (syntactic, word context) or external (meta-data, sentence context) information.

\subsection{Major areas of research}

Research into intent classification in SLU has generally come from four areas: search engines, question answering systems, dialogue systems, and text categorisation. Early search engines applied text similarity to select results for users. More recently, intent classification has been applied to understand the searcher's intent further and this approach has been proven to give better search results. However, web queries are usually short and informal, causing difficulties in classifying intents because of insufficient information. Similarly, answering questions from users also benefits from understanding the intents of questions to generate better quality responses. In dialogue systems it has been shown to be useful to identify intents of users in order to give appropriate responses to users. Moreover, intent classification can be applied in more general NLP tasks, such as text classification, sentiment analysis and scientific citation.

\subsection{Overview of technological approaches}

Before 2015, most papers focused on classical machine learning approaches, such as SVM, K-nearest neighbours and random forest.  Features used by these models were mainly generated by dependency parsing, word embedding and n-grams. One deep learning method explored early on was deep belief networks (DBN). In more recent years, with the success of deep learning in other areas, neural networks, especially RNNs, started to be widely used for this task. Attention mechanisms have been integrated in models for identifying which parts of sentences should contribute to the classification. Since intent classification is proposed to be integrated with web engineering, which requires the ability to understand short texts that contain less information, features used for training have been enriched by feature engineering using web metadata.

\subsection{Issues addressed in intent detection}

In this section we survey the issues encountered in the literature around intent detection, and the solutions proposed. The issues may be specific to the task, like ambiguity of semantic intent. They may be general machine learning issues like lack of training data, and dealing with new or adapting domains. They may be issues specific to the available data like imbalanced data, short sentences, or the out-of-vocabulary (OOV) issue. Feature creation to capture information extra to that in the words is considered to boost performance. Extending the range of the task to multiple intents or to identify out-of-domain sentences is covered.  

\subsubsection{Ambiguity in interpretation}

In essence, this issue is at the heart of intent classification; identifying the decision boundary between samples close together in feature space, yet belonging to different classes. This issue may be more prevalent with short texts, since they may include insufficient information and not follow correct grammar.

An early approach from \cite{Purohit_2015} was to propose a rich feature representation with an ensemble learning framework giving different perspectives on the classification. The feature creation is covered further in ensuing sections.

A more recent approach from \cite{ren2020siamese} proposed training triples of samples - an anchor sample, a positive sample in the same class and a negative sample from a different class. Combining convolutional and BERT encodings of each one and mapping them to Euclidean space with Siamese shared weights, an intermediate loss of the anchor-positive distance minus the anchor-negative distance is minimised. The Euclidean mapping of the anchor is used for classification.

This latter approach feeds in to the emerging field of contrastive learning and methods from there should be deployed in the NLU field.

\subsubsection{Lack of labelled training data or small training sets}

Collecting and labelling large amounts of data for training can be expensive. With small data sets, models are more likely to be over-trained. Further, the out-of-vocabulary (OOV) issue, where words appear in the test set that are not in the training set, is more likely to occur with them.

\cite{Sarikaya_2011} proposed a DBN-initialised neural network for intent classification to learn from unlabelled data and generate features for a feed-forward network. The feed forward network is then fine-tuned on labelled data which may be small in number but still give reasonable results. A DBN is a stack of Restricted Boltzmann Machines (RBMs). To train a DBN, the RBMs are trained layer-by-layer in sequence using parameters learned by previous layers. After training the stack of RBMs, the weights of the DBN are used to initialise the weights of a feed-forward neural network. This approach performed better than traditional machine learning models, such as maximum entropy and boosting, and similarly to SVM.

\cite{Hasanuzzaman_2015} tried to include temporal query understanding into web search query intent classification. They tackled two major issues: one is the inadequacy of limited training data while the other is the limited literal features able to be extracted from queries of short length (typically 3-4 words). They utilised external resources collected from the web that may help bolster temporal information, such as web snippets for queries and the most relevant year, date etc. Based on this 28 features were designed and extracted. They then proposed an ensemble learning solution framework defined as a multi-objective optimisation problem (MOO) and explored with 28 classifiers using different optimisation strategies. The utilisation of external resources and ensemble learning was intended to reduce bias to better handle the limited training data.


Methods for dealing with new unannotated domains and data sets by transfer of concept from existing data sets or models weights combined with few shot methodologies have been explored in the slot labelling and joint task area and are discussed later.

Sufficient data is essential for training a model. To work with unlabelled data, unsupervised training methods could be investigated in further research.

\subsubsection{Multi-domain/multi-lingual generalisability}

Most text classification models focus on only one language, one domain and also one task. Some models have been proposed to have better generalisability.

\cite{costello2018multi} developed a novel multi-layer ensembling approach that ensembles both different model initialisation and different model architectures to determine how multi-layer ensembling improves performance on multilingual intent classification. They constructed a CNN with character-level embedding and a bidirectional CNN with attention mechanism. In addition, they explored LSTM and GRU with or without character-level embedding and attention mechanism. When ensembling models, they use a majority vote with confidence approach.

\cite{Masumura_2018} proposed an adversarial training method for the multi-task and multi-lingual joint modelling to improve performance on minority data. The language-specific network can be shared between multiple tasks, where words in the input utterance are converted into language-specific hidden representations. Next, each word representation is converted into a hidden representation that uses BiLSTMs to take neighbouring word context information into account. Task-specific networks can be shared between multiple languages, where the language-specific hidden representations are converted into task-specific hidden representations. The proposed method combines a language-specific task adversarial network with a task-specific language adversarial network. 

\subsubsection{Emerging intents detection} \label{sec emergeintent}
In dynamic real world applications the intent set evolves. A method to detect and classify emerging intents is a desirable adjunct task.

\cite{hashemi2016query} proposed to use a CNN to extract query features for intent classification, which is trained based on word-level embeddings generated by word2vec trained on Google News. Query representations are taken after a max pooling layer. They perform clustering on these representations and observe that new examples far from the clusters could be used to identify emerging intents, though they do not perform that task.

\cite{Xia_2018} proposed two capsule-based architectures to detect emerging intents. They construct three capsules, SemanticCaps, DetectionCaps and Zero-shot DetectionCaps. SemanticCaps is based on a bidirectional RNN with multiple self-attention heads and is used to extract semantic features from utterances. Then, DetectionCaps aggregate the low-level information from SemanticCaps to high-level information in an unsupervised routing-by-agreement approach and obtain intent representations. For detecting emerging intents, the Zero-shot DetectionCap takes information from SemanticCaps and DetectionCaps to calculate vote vectors for information transferral. Then, the vote vectors are multiplied with similarity between embeddings of existing intents and emerging intents and summed to generate representations of emerging intent labels.

\subsubsection{Unseen intents}

Dealing with intents which are unseen in the training data is a related challenging task. \cite{Lin_2019} proposed a two-stage method to detect unseen intent labels. First, they used a Bi-LSTM to extract features of a sentence. Then, the forward output vector and the backward output vector were concatenated and the concatenation result was used as the input of the next stage. The model uses large margin cosine loss (LMCL) as the loss function, instead of softmax loss, which aims to maximise the decision margin. Thus, the inter-class variance is maximised and the intra-class variance is minimised. This is to ensure that the features extracted by Bi-LSTM can be more discriminative. In the second stage, the model takes the concatenation vector and applies a local outlier factor (LOF) to detect unseen intents, which is a density-based detection algorithm.

\paragraph{The OOV issue}
Most word embedding based approaches are dependent on vocabularies and may suffer to some extent from OOV issues, though small training data sets may be affected more. Using character n-grams is a common approach to handle unseen words based on the idea that similar words may come from a common root. Another is to replace all words below a chosen frequency in the training set with a special token, say UNKNOWN.

\cite{ravuri2015recurrent} proposed character n-grams as their word input encoding method with both RNN and LSTM, since they thought the OOV issues became more severe when using RNNs because an unknown word could propagate an effect to the consequent words. Similarly, \cite{Shridhar_2019} proposed sub-word semantic hashing inspired by the Deep Semantic Similarity Model for solving the OOV issue which comes with small training data sets. Before sub-word semantic hashing, sentences are transferred into lower case, pronouns in sentences are replaced by `-PRON-', and special characters except stop characters are removed. Then, classes with less sentences are oversampled by adding augmented sentences, which are generated using synonym replacement. After this, every token in each sentence is wrapped by two '\#' symbols and represented using trigrams. These sub-tokens are then vectorised using an inverse document frequency vector and the Euclidean norm. In the end, the vectors can be used in any intent classification model.

\subsubsection{Short text queries}

User queries for search engines are usually short (3-4 words) and lack context, so it is essential to extract more information from queries for successful classification. Syntactic features, such as POS tags, and also external knowledge sources have been used to enrich query features.

\cite{Hasanuzzaman_2015} included temporal query understanding into web search query intent classification and their model works well with the limited literal features for queries of short length. \cite{Purohit_2015} focused on intent understanding of social media text such as tweets. Some of these can be short, leading to ambiguity of interpretation and sparsity of relevant behaviours. They try to improve the expressiveness of data by utilising multiple patterns from knowledge sources and fuse the top-down knowledge-guided patterns with bottom-up frequency-based representation for feature formation. Based on this, they utilise an ensemble learning strategy to reduce the bias.

\cite{Xie_2018} proposed a model called Semantic Tag-empowered User Intent Classification (ST-UIC), based on a constructed semantic tag repository. This model uses a combination of four kinds of features including characters, non-key-noun part-of-speech tags, target words, and semantic tags. After pre-processing, characters and target word features are extracted for maintaining the contextual information. Then, key nouns are expanded using semantic tags and POS tags are used as features if a query does not contain target words. With this approach, representation can be enriched for short queries.

In contrast, \cite{tur2011sentence} tried to simplify the query input based on a dependency parser to generate simple and well-formed queries. They were motivated by the performance gain of existing statistical SLU models on simple, well-formed queries as well as the need for handling increased web search queries formed by key words. They simply kept the top level predicate and its dependants for the query simplification, and combined it with the sentence input for further classification using AdaBoost. The essence here is to try to provide the extracted key word pieces as auxiliary information, which proved to decrease the intent classification error rate. Because some semantic and syntactic information contained in the sentence are filtered out, when this simplified syntactic structure of sentence was used alone as input for classification, a decrease of performance was reported.

This issue mainly occurs with user queries for web searching. While not widely reported on in the joint task literature, short texts do occur in other data sets and the methods described above can be applied there. Rather than the rule based feature construction approach, knowledge graphs may be further investigated as a method for adding relevant external information.

\subsubsection{Other feature engineering}

Even when the utterances are longer the challenge is to extract  information relevant to the classification task. One early solution generating features using neural networks was by \cite{Sarikaya_2011}, who generated features for a feed-forward network using their DBN-initialised neural network. After training, the features generated by DBNs were found to be useful in discriminative classification tasks.

Many papers since have used standard word embeddings, RNN and CNN feature creation. In order to boost semantic understanding \cite{Wang_2019} proposed to use both in a model called Character-CNN-BGRU. Firstly this model uses character embeddings to represent sentences, rather than word embeddings. A CNN takes the character embedding as the input and extracts local features via max pooling after a convolutional layer. Meanwhile, a window feature sequence layer is added on the convolutional layer to obtain temporal information, which is important for the bidirectional gated recurrent unit (BiGRU). Finally, the output of the max pooling layer and the output of BiGRU are concatenated and passed to a softmax layer.

While semantic features are useful, methods to extract other usable information for classification have been developed.

\paragraph{Grammar feature exploration}
From the question answering field comes a feature creation technique based on grammar. Question words, such as "what", "how" and "where", and also domain specific grammar may indicate the class of a question. Based on this idea, \cite{Mohasseb_2018} proposed a grammar-based framework for question classification, which utilises three features: grammatical features, domain specific grammatical features and grammatical patterns. Grammatical features are used to parse a question into a sequence of grammatical terms. Domain specific features are used to identify the domains which grammatical terms in the sentence correspond to and tag them. After parsing and tagging each term in the question, the pattern is formulated. The classification task is processed using machine learning models, such as SVM or the J48 algorithm.

\subsubsection{Imbalanced data}

A data set is unlikely to have the same number of samples for each class and sometimes data can be quite imbalanced. Training a model using imbalanced data can cause poor performance on minority classes.

\cite{Purohit_2015} noticed that social media text corpora can have such imbalanced data. They suggested that two of their constructed features aid with imbalanced data in their data set. These are Contrast Patterns features, where they mine sequential patterns within each intent class then contrast them. \cite{Shridhar_2019} dealt with imbalanced data through oversampling by adding augmented sentences for classes with less samples during the pre-processing stage of their model.

ATIS, one of the major data sets for the joint task is very imbalanced in the intent aspect, and yet performance is excellent. The imbalanced data issue is rarely brought up in the joint task literature.

\subsubsection{Co-occurrence of words from different intents}

This is a particular form of ambiguity. Words important to different intents may co-occur in a query of a particular intent and how these words are positioned may convey crucial information for intent detection of the current query.

Based on this idea, \cite{zhang2016mining} proposed two types of heterogeneous information: (1) pairwise word feature correlations (2) POS tags of the queries. The pairwise feature correlations are calculated based on cosine similarity between each semantic feature pair and learned by CNNs with pooling layers. Here each dimension of the word vector is deemed to be a semantic feature. It tries to model the intent through these feature-level representations. Meanwhile the POS tags provide the word-level information about word categories. Their experiment results show that utilising the feature-level semantic representation outperform the baseline model using only word-level features and incorporating POS information with the feature-level representation significantly improves the performance.


\subsubsection{Contextual/temporal information modelling}

A single sentence in a conversation can be ambiguous, but the ambiguity can be eliminated if previous utterances are considered during intent classification. Meanwhile, web queries are sensitive to time and the intent carried by them may change over time; for example as world events wax and wane in importance. Therefore, some studies incorporated contextual and temporal information in intent classification.

With multi-turn dialogue \cite{bhargava2013easy} included the context from previous queries for the intent classification and slot filling of the current query. Each sub-task is treated separately, so this is not a joint model. For intent classification, they compared two approaches. The first is to construct a context sessions feature by simply using 1 for the type of intent inferred from the previous utterance set and 0 to all others. This is combined with a basic bag-of-n-gram feature for the current utterance and fed to an SVM for intent classification. The second approach is to treat the query sequence as a sequential tagging problem using SVM-HMMs with a Viterbi algorithm. The incorporation of intent from previous utterances as additional information showed a significant reduction in error rate.

\cite{Hasanuzzaman_2015} utilised external resources collected from the web that may help bolster temporal information, such as web snippets for queries and the most relevant year, date and other time indicators. Based on this, 28 features are designed and extracted. Another solution to generate features is referring to one of a set of contemporaneous events, known as event-based web searching. \cite{kanhabua2015learning} utilised the event-related log patterns that reveal both implicit and explicit temporal information needs, together with general lexical information such as named entities for feature representation. Classification is performed by SVM, AdaBoost, decision tree/J48, and a neural network.

\cite{Chen_2012} included temporal information in their model as well. They proposed a metadata feature for enhancing community query intent classification. The metadata feature included query topic, query time, and user experience indicated by the number of previous queries. They firstly explored supervised learning using the text and metadata features separately. The result showed that using both two features together outperformed the separate experiment for query intent classification. This drove them to use a co-training procedure, which is a semi-supervised learning framework that can utilise a small amount of annotated queries plus a large amount of unlabelled ones. During the training, two separate and independent classifiers are trained first based on the two features. Then the prediction with higher confidence from either of the classifiers will be used as the label for the unlabelled query for training until a stopping criteria is reached. Utilising the predictions of the two separate classifiers as labels for further training requires that the two features are conditionally independent and sufficient for classification. Experimenting with SVM, higher micro and macro F1 scores were achieved from semi-supervised co-training compared to supervised learning with a combination of the two features.

Rather than temporal information \cite{Qiu_2018} proposed the construction of multiple features from user metadata, regex extraction of named entities, and probabilistic context free grammar of composite entities.

Contextual and temporal information may be not suitable for all data sets, but, when available, future research can attempt to use it with their models. The multi-turn dialogue approach will be explored further under the joint task. Graphs may also be explored for integration of contemporaneous topics and events.

\subsubsection{Target variation}

Typically in intent classification the pre-defined intents are enumerated and a cross-entropy loss is calculated. \cite{Qiu_2018} rather calculated LSTM embeddings of the training sentences and averaged them within samples of the same intent. For test prediction they calculate a similarity measure of the LSTM embedded test sample with the training samples averaged by intent and choose the closest.

A label can be composed from several words; "play\_music" is composed of "play" and "music", for example. Future research may investigate whether some words in the sentence correspond to "play" and some other words correspond to "music".

\subsubsection{Generalisability via ensembles}

Overfitting of models to the training data distribution leads to poor generalisability. One method to address this is to use ensembles of models to synergistically exploit the benefits of each. Deep learning architectures, such as LSTM, GRU and CNN, have been used frequently in intent classification. \cite{Firdaus_2018} proposed to combine those deep learning architectures. They used GloVe and word2vec for word embedding. To start exploring combinations of different models, they constructed CNN, LSTM and GRU individually. Then, four ensemble models were built, which are CNN-LSTM, CNN-GRU, LSTM-GRU and CNN-LSTM-GRU. In these models, predictions from individual models are combined using a MLP model. \cite{Qiu_2018} also used an ensemble of classical methods, being random forest, SVM, Naïve Bayes and softmax regression, with the ensemble outperforming the components.

\paragraph{Out-of-domain utterances}

Not all utterances made to SLU devices contain an intent related to the purpose of the device. They may be incidental conversation or have intent that the device is not meant to fulfil.

\cite{kim2018ood} augmented a dataset with such utterances and performed a multi-task learning approach to perform classification of in domain utterances and detection of out-of-domain (OOD) utterances simultaneously. A loss function which maximises the intent accuracy while accepting a value of false acceptance rate (1-recall) for OOD utterances below a threshold is back-propagated. Including the second task boosts the intent detection performance.

\cite{yilmaz2020kloos} also constructed augmented data sets with OOD utterances. They constructed a vector of KL divergence values for subsequent pairs of intent probabilities determined from hidden states of a unidirectional LSTM fed with word embeddings. The KL vectors are fed to an SVM, Naïve Bayes and Logistic Regression for OOD classification. Logistic regression gives the best results.

\subsubsection{Multifaceted query intent prediction} \label{sec intmulti}

Most annotated queries in training data sets express only one intent. In real life however, queries may contain more than one intent. For example, the query ``find Beyonce’s movie and music'' has two intents, `find\_movie' and `find\_music'. An NLU system should be able to handle multi-intent queries.

One straightforward solution is to use the top couple of predictions from existing single label classifiers. Another solution is having binary classifiers for every label in single classifiers. \cite{gonzalez2011multifaceted} used this approach. In their model, each utterances has multiple facets, each a class with at least two categorical labels. For instance, in the Task facet the intent can be 'Informational', 'Not Informational', or 'Ambiguous'. They experimented with linear SVMs for intent classification of different combination of the facets. Compared with the corresponding single facet classification, additional supervision from multiple labels led to improvement of the overall performance. This additional information was found beneficial to small categories (classes with few samples in the corpus) with the recall of those small categories improving. 

Treating multi-labels as atomic labels has been explored in some studies. This approach may suffer data sparsity problem, but has good classification accuracy. Based on this, \cite{xu2013exploiting} proposed two approaches to exploit the information shared among different intent combinations. The first one is adding class features, which is to add n-gram features for combined intent appropriate to the separated intents. For example, the multi-intent label, 'buy\_game\#play\_game', should have added two features for the embedded intents 'buy\_game' and 'play\_game'. The other is adding hidden variables to identify segments belonging to each intent. Instead of using existing segmentation algorithms, they add a layer of hidden states corresponding to each word. This can indicate the embedded intent which a word most strongly aligns to. Following that, a perceptron layer performs the classification.

Understanding queries with multiple intents can make conversations with dialogue systems more natural and smooth, but there is not much work in this area. Further research could explore more approaches on modelling relations among label combinations.

\section{Slot filling} \label{sec slot}

Slot filling is the second critical task in natural language understanding. It is the attachment of a label to each token in an utterance. The label describes the type of semantic information contained in the word represented by the token. A span is a contiguous set of words which together make up a semantic unit, for example ``new york city'' is a single span represented in BIO notation with the labels \textit{B-city I-city I-city}. Slot filling is treated as a sequence labelling task. The task is learning not just slot label distributions for words but also what slot labels typically co-occur in utterances (label dependency), and in what order. Reference to context in both directions of a word should be included to maximise performance.

\subsection{Major areas of research}

Slot filling is a key task in dialogue systems, to interpret natural language from users, from which the system can judge what information to retrieve or what task to complete for the user. Slot filling models are also integrated with online shopping websites, whose core is a task-oriented dialog system. Product search queries can be better understood and shopping assistance can be provided to customers. The field of question answering has provided research to extract the semantic features within queries. 

\subsection{Overview of technological approaches}

Traditionally, generative models (for example, HMMs \cite{wang2005spoken}) which capture the joint probability distribution of the utterance tokens and their slot labels, and discriminative models (like CRFs) that estimate slot label conditional probabilities given the observed token sequence, were used to address this problem. With the success of deep learning, researchers experimented with putting components, for example, CRFs and brief networks, within a deep structure. 

Since 2013, RNNs have been increasingly popular in this field. In RNNs each word can access information from the previous words. Later, bi-directional RNNs were applied to utilise both the past and future context. However, the distance between words is linear in RNNs, and the vanishing gradient problem may occur, meaning that long-term dependencies cannot be learnt by the model. As a result, LSTM cells began to be used more frequently because of their ability to forget unimportant information and more successfully model longer dependencies. However, even LSTMs can underperform with very long sentences. Other models are thus incorporated to capture label dependency, which refers to the situation that some slots appear in context with some other slots more frequently, and perform sequence-level optimisation. The combination of CRF layers attached to RNNs is often seen. 

Another issue is that RNNs are typically not able to process multiple words simultaneously due to their sequential nature. Thus, attention mechanisms which can take more tokens into account simultaneously than RNNs are used. Additional features are also incorporated to improve the performance of RNNs, such as named entity features, segment features and external memory. Integrating extra knowledge from different sources is also considered an effective approach. Some recent slot filling models also attempt to handle unseen semantic labels and multiple domain tasks by adapting neural CRFs and label embedding.

\subsubsection{Exploring new models in detail}

Models that have proved reliable for sequence labelling in other fields have been adopted in addressing the slot filling problem. In particular, deep learning models have been applied in slot filling. 

\cite{li2012use} made an early deep model by constructing a stacked model, in their case a deep convex network (DCN), and extended it to the kernel version (K-DCN) for domain and intent classification tasks. With the kernel approach, the number of layers can be increased. Later, they attempted to solve slot filling problems using K-DCN as feature extractors.

RNNs with different architectures have been explored in many studies, considering their promising performance in sequence modeling elsewhere. In 2013, \cite{mesnil2013investigation} compared recurrent neural networks, including Elman-type and Jordan-type networks and bi-directional Jordan-type RNNs. Two years later, \cite{mesnil2015recurrent} implemented Elman-type and Jordan-type networks and also their variations. Both Elman and Jordan-type networks are constructed with a 3-word context window. Moreover, a bi-directional Jordan-type network was implemented which takes both past and future information into account.

\cite{yao2013recurrent} adopted Recurrent Neural Network Language Models (RNN-LMs) to predict slot labels rather than words. This model used an Elman architecture RNN which can remember past words. Originally, the output of the training model is exactly the input word sequence, but in the new model, the outputs are sequences of labels instead. Future words, named entities, syntactic features and word-class information were also integrated in the analysis.

In the first models using RNNs in NLU, only words preceding the current word were considered. However, words occurring after the current word can also provide useful information. Therefore, \cite{vu2016bidirectional} use bi-directional Elman-type networks with a 3-word context window. In their network, the BiLSTM generates forward output and backward output, which are combined for making predictions. The model adopts the ranking loss function, so the model is not forced to learn a pattern for the artificial class O. In an extension \cite{vu2016sequential} proposed a bi-directional sequential CNN for slot labelling which considers both previous contextual words with preserved order, surrounding context and also past and future information. To label a word, two matrices are separately generated for the previous and future context words of the word. These are combined to form a matrix for the current word. Then, the two matrices for the past and future information are passed to corresponding vanilla sequential CNNs. The output of networks are concatenated with the matrix for the current word. After that two matrices can be combined using a weighted sum of the forward and the backward hidden layer or by concatenating. For training, the ranking loss function is again applied.

\cite{korpusik2019comparison} perform a useful comparison of BiGRU, CNN and BERT based models with the then new BERT outperforming the other candidates. 

The incorporation of newer technologies for slot labelling, particularly seemingly suitable attention based methods like the Transformer, has been subsumed by the joint task in more recent years as will be explored in Section \ref{sec joint}.

\begin{table*}[t]
\caption{Slot filling papers reviewed with addressed issue and approach}
\begin{tabular}{ p{3.5cm}p{4.9cm}p{8.4cm} }

\hline
\textbf{Paper} & \textbf{Addressed issue} & \textbf{Approach} \\


\hline
\cite{yu2011sequence} & Long-range state dependency & Deep learning, CRF  \\

\hline
\cite{li2012use} & To extend DCN & Kernel learning, deep learning, DCN, log-linear model  \\

\hline
\cite{deoras2013deep} & Data sparsity problem (CRF) & Deep belief network (DBN)  \\

\hline
\cite{mesnil2013investigation} & To explore RNN & RNN  \\

\hline
\cite{yao2013recurrent} & To explore RNN-LM & RNN-LM  \\

\hline
\cite{yao2014recurrent} & Label dependencies, label bias problem & RNN, CRF  \\

\hline
\cite{yao2014spoken} & Gradient diminishing and exploding problem, label dependencies, label bias problem & LSTM, regression model, deep learning  \\

\hline
\cite{liu2015recurrent} & Label dependencies & RNN, sampling approach  \\

\hline
\cite{mesnil2015recurrent} & To explore RNN & RNN  \\

\hline
\cite{peng2015recurrent} & Vanishing and exploding gradient  & RNN, external memory  \\

\hline
\cite{kurata2016leveraging} & Label dependencies & LSTM, encoder-labeler  \\

\hline
\cite{vu2016bidirectional} & To explore past and future information & Bi-directional RNN, ranking loss function  \\

\hline
\cite{vu2016sequential} & To explore CNN & CNN  \\

\hline
\cite{zhu2017encoder} & To explore attention mechanism & Bi-directional LSTM, LSTM, encoder-decoder, focus mechanism  \\

\hline

\hline
\cite{dai2018elastic} & Unseen slots & CRF  \\

\hline
\cite{gong2019deep} & To explore MTL & MTL, segment tagging, NER  \\

\hline
\cite{louvan2018exploring} & To explore MTL & MTL, NER, bi-LSTM, CRF  \\

\hline
\cite{shin2018slot} & To better labelling common words & Encoder-decoder attention, delexicalised sentence generation  \\

\hline
\cite{wang2018new} & Imbalanced data & DNN, reinforcement learning  \\

\hline
\cite{zhao2018improving} & OOV & GRU, attention, pointer network  \\

\hline
\cite{gong2019deep} & To explore MTL & MTL, segment tagging, NER  \\

\hline
\cite{kim2019slot} & To extend original SLU to H2H conversations & Bi-LSTM, different knowledge sources  \\

\hline
\cite{shen2019progressive} & Continual learning & Bi-LSTM, context gate  \\

\hline
\cite{veyseh2019improving} & Restricted utilising contextual information & MTL, Bi-LSTM \\

\hline
\cite{korpusik2019comparison} & Architecture comparison & BiGRU, CNN, BERT \\

\hline
\cite{louvan2019leveraging} & Low resource data set & MTL \\

\hline
\cite{zhu2020prior} & Data sparsity problem & Prior knowledge driven label embedding, CRF  \\

\hline
\cite{zhang2020deeptime} & Long range dependency, vanishing gradient & TDNN  \\

\hline
\end{tabular}
\end{table*}

\subsection{Issues addressed by slot labelling papers}

Like intent detection, issues found in slot filling can be task specific, more general machine learning issues, data set issues, or may involve feature engineering or methodological approaches for better results. Slot filling introduces issues typical to seq2seq tasks like label dependency and long distance dependency. 

\subsubsection{Label dependency}

There are dependencies between slot labels, meaning that some slots appear more commonly with some other slots in the same utterance. For example, in a travel data set it is highly probable that \textit{B-FromCity} and \textit{B-ToCity} are in the same sentence. Capturing such label dependencies would help find the best slot combinations and generate better prediction results.

One approach is to integrate CRFs and regression models. \cite{yao2014spoken} applied an LSTM for the slot labelling task and included a regression model to capture label dependencies. \cite{yao2014recurrent} proposed a recurrent conditional random ﬁeld (R-CRF) which integrates recurrent neural networks and a CRF to explicitly model the dependencies between semantic labels and achieve sequence-level optimisation. The R-CRF can use the RNN activations as features of the CRF. Similar to CRFs, this model can use sequence-level optimisation as well.

Another solution utilises the encoder-decoder architecture, which had been applied for machine translation previously and is able to encode the global information of the input sentence. \cite{kurata2016leveraging} proposed the encoder-labeler LSTM. This architecture encodes the sentence to a fixed length vector. Then, the encoding vectors are used as the input of another LSTM, the labeler LSTM, which considers label dependencies. The labeler LSTM can predict the slot label conditioned on the encoded sentence information. With such a method, the model is able to label slots utilising the whole sentence information. \cite{zhu2017encoder} proposed a BiLSTM-LSTM encoder-decoder model, where a sentence is encoded using a bi-directional LSTM and the encoded sentence is then decoded using a uni-directional LSTM. At the same time, they developed a focus mechanism for this model because of the alignment limitation of attention mechanisms.

Moreover, \cite{liu2015recurrent} used continuous vectors to represent possible output labels, which are fed to recurrent connections. These continuous vectors are also fed to hidden layers, so every hidden layer can utilise word input, previous hidden states and predicted output labels. Moreover, both true labels and the predicted output labels can be fed to some layers in a fashion decided by a sampling approach for robustness. Because the previous predicted labels are used as input to the next step, error propagation should be studied further.

\subsubsection{Long range dependency}

Long-range dependency (LRD), also called long memory or long-range persistence, is a phenomenon that may arise in the analysis of spatial or time series data. It relates to the rate of decay of statistical dependence of two points with increasing time interval or spatial distance between the points. While the short text queries described in the intent section should not suffer from this, utterances in SLU data sets can be much longer.

The approach from \cite{yu2011sequence} is a deep structured CRF which is made up of several simple CRFs. Lower layers can generate frame-level marginal posterior probabilities. Then the higher layer takes these probabilities along with the observation sequence of the previous layer. In the end, the highest level will make the final predictions. In the training process, layers are trained separately for efficiency. For each layer, the parameters are determined once the layer is trained.

\cite{zhang2020deeptime} addressed the shortcomings of RNN being the long range dependency issue and vanishing or exploding gradient. They used a deep stacking of time delay neural networks for feature creation. These create convolutional features from context windows of varying size and varying step sizes through the sentences. The features are then passed through a final RNN and classified.

The Transformer architecture, which utilises self attention across utterances, and memory networks which can store longer range information, are now studied in the joint task, as described in the next section. Focused research using these methods for only the slot task may uncover useful methods for use in the joint task.

\subsubsection{The label bias problem}

Label bias is a seq2seq issue related to maximum entropy Markov models (MEMM). In MEMM, the states with a single outgoing transition can ignore their observation, meaning that the model can tend to stay in a state which is unlikely to happen. Most approaches to this problem combine CRFs with RNNs.

\cite{yao2014spoken} applied LSTM cells, which contain a gate which can forget unimportant information, and also incorporated a regression model to model label dependencies. In order to avoid the label bias problem, the regression model took non-normalised scores before softmax. The R-CRF of \cite{yao2014recurrent} also addresses label bias. Similarly, \cite{mesnil2015recurrent} explored RNNs with multiple different architectures and proposed to apply Viterbi encodings and recurrent CRFs to eliminate the label bias problem.

\subsubsection{Learning common words}

The surrounding words of one slot in different sentences are usually similar. For example, the word "to" is highly likely to be lie between \textit{B-FromCity} and \textit{B-ToCity}. Therefore, \cite{kurata2016leveraging} thought that learning the common words around slots in different ways may be helpful for slot filling.

\cite{shin2018slot} introduced a model which can jointly generate delexicalised sentences and predict labels using the encoder-decoder framework with input alignment. In the delexicalised sentences, words are replaced by their corresponding slot labels. For example, to delexicalise the sentence “i want to fly from baltimore to dallas”, the words `baltimore' and `dallas' should be replaced by \textit{B-FromCity} and \textit{B-ToCity}. This approach is based on the fact that different words that correspond to the same slot usually play a similar semantic and syntactic role in the sentence, which allows the model to learn the common words surrounding the slots.

\subsubsection{Low resource data sets}

A more general machine learning issue is that methods generally rely on the presence of annotated training data which is costly to produce. \cite{louvan2019leveraging} tested models which also trained on a large freely available annotated data set for a similar task (NER for example) in a multi-task learning environment (see Section \ref{sec slotmtl}. They then used just 10\% of the available training data from slot tagging data sets to train the slot labelling task. They varied this proportion and observed when it reached 40\% that the auxiliary task stopped having much effect.

\subsubsection{Diminishing and exploding gradients}

In neural networks with multiple layers, or long RNN sequences, the gradient may become vanishingly small, preventing the weights from changing value during back-propagation. Alternatively, large gradients may self-propagate and lead to unstable networks.

\cite{yao2014spoken} combined an LSTM and a regression model for slot filling. LSTM are partially designed to address these issues. To avoid the gradient diminishing and exploding problem, the memory cells within them are linearly activated and propagated between different time steps.

\cite{peng2015recurrent} introduced an external memory to overcome the limitations on memory capacity of simple RNNs and therefore the diminishing and exploding gradient problems can be addressed. The new model is call RNN-EM, where the hidden layer has an additional input that comes from the external memory. A weight vector is used to retrieve the content from the external memory, which is determined according to the similarity between the contents and the hidden layer.

\subsubsection{Data sparsity problem}

Data sparsity can be an issue when there are a large number of discrete features. If those discrete feature are represented using matrices, those matrices can be large and sparse, which may lead to a model ignoring the relations among features.

\cite{deoras2013deep} applied deep belief networks (DBN) for semantic tagging integrated with lexical, named entity, dependency parser based syntactic features and part of speech (POS) tags. A DBN is a stack of Restricted Boltzmann Machines (RBMs), where the input of one layer is the output of the previous one and each layer applies a sigmoid activation function on their inputs. In this model, features are embedded into vectors at the first layer and passed to the next layer due to the large input layer. During the pre-training process, the parameters of neurons are learnt using the online version of conjugate gradient (CG) optimisation on several small batches. Compared to CRF, DBN is more general.

\cite{zhu2020prior} noticed that the data sparsity problem can also occur with labels because labels are usually encoded using one-hot vectors, so they also proposed a label embedding which is constructed using prior knowledge including atomic concepts, slot descriptions, and slot exemplars. An atomic concept assumes that each slot can be represented as a set of atoms. Slot descriptions are the textual description for slots in natural language. Slot exemplars are to extract label embeddings for slot labels which contain values of each slot and their neighbour contexts.


\subsubsection{Continuous learning}

With new data becoming available quickly, it is desirable to have a retrained model incorporating the new data. However, the training process could be time and cost consuming and keeping all the existing data can introduce redundancy. Thus, there is a problem of how to continuously learn from new data.

\cite{shen2019progressive} proposed a ProgModel, consisting of a context gate. This gate aims to transfer previously learned knowledge to a small expanded component, which is placed after the hidden state of the new model. The training procedure is progressively conducted at each batch. Therefore, the model can learn faster from the new training data without forgetting the previous expressions.

\subsubsection{Imbalanced data}

Training data sets can be imbalanced, dominated by some tags while only containing a small number of examples of other tags. This can lead to poorer performance in the minority tags. One solution from the slot labelling literature comes from \cite{wang2018new} who design a deep reinforcement learning (DRL) based augmented tagger with a deep neural network, which includes a training part and an inference part. While the whole data set is used in the training part, only partial data with unsatisfactory performance will be evaluated by the augmented tagger.

Similarly to intent classification, augmentation via generating sentences which contain the minor tags could be researched.

\subsubsection{Unseen labels}

As with intent classification models, slot labelling models can rely heavily on the training data and will struggle to correctly assign a label which does not appear in the training data.

\cite{zhao2018improving} proposed a seq2seq model together with a pointer network to solve this problem. This model predicts slot values by jointly learning to copy a word which may be out-of-vocabulary (OOV) from an input utterance through a pointer network, or generate a word within the vocabulary through a seq2Seq model with attention.

\cite{dai2018elastic} proposed an elastic conditional random field (eCRF) which can utilise semantic meaning in slot embedding for open-ontology slot filling. The model has a slot description encoder which takes all slot descriptions as input, and outputs distributed representations for slots. Meanwhile, a BiLSTM is used to extract features from utterances. Then, the eCRF labeller, which is a potential function containing two terms for semantic similarity of the slot descriptions and the extracted contextual features and interactions between the slot labels, is applied.

\subsubsection{Exploring multi-task learning} \label{sec slotmtl}

Multi-task learning (MTL) is the idea that similar auxiliary tasks can assist a main task. Some papers have used this idea setting slot labelling as the main task.


\cite{louvan2018exploring} attempted to perform slot filling and named entity recognition (NER) jointly in a multi-task framework considering that most slot values are also named entities and NER has high state-of-the-art performance. They mentioned that a better slot tagging result can be achieved if NER is at a lower level. Similarly, \cite{gong2019deep} investigated hierarchical multi-task learning to perform low-level tasks first, namely named entity tagging and segment tagging, and then the high-level task, that is slot labelling, can make use of the results from the low levels with cascade and residual connections. \cite{louvan2019leveraging} then performed a more wide ranging experiment with two auxiliary tasks (NER and semantic tagging (SemTag)), and a comparison of a training on auxiliary task(s) followed by fine-tuning on the main task approach versus the previously explored hierarchical approach. They found generally the hierarchical approach gave better results and that parsimoniously using only one auxiliary task (NER) worked better.

Most models utilise contextual information, but they use it in a restricted manner, for example, self-attention. Therefore, \cite{veyseh2019improving} proposed a multi-task setting to train a model to incorporate the contextual information in two different levels which are representation level and task-specific level. This multi-task setting includes the slot filling as the main task and two auxiliary tasks. The first one is to increase consistency between the word representation and its context and another one is to enhance task specific information in contextual information.


\subsubsection{Extending to human-to-human conversations}

As the main thrust of this survey shows, task oriented language understanding in human-to-machine (H2M) conversations has been extensively studied. An interesting twist is to perform slot tagging in human-to-human (H2H) conversations. Here the agent is a third party listening in, not being directly asked to perform a task.

\cite{kim2019slot} focused on slot filling in H2H conversations and explored LSTMs with different knowledge sources. First, the character embedding and the word embedding are concatenated for each word. Then, the embeddings are passed to a bi-directional LSTM, which will be used for making final predictions. Furthermore, there is an additional model which can utilise knowledge from multiple sources, including sentence embeddings for H2H conversation, contextual information and H2M expert feedback. The sentence embeddings are generated by a sentence level embedding model trained using tweets with URL and web search queries to the same URL. The contextual information is extracted from previous utterances in the conversation. Also, this model uses pre-trained slot filling models for H2M conversations on similar domains as the expert model. The knowledge from three sources is encoded into vectors which are then combined and aggregated with the output of the bi-directional LSTM. 

Slot tagging introduces the problem of generating a sequence of hidden labels to a sequence of word tokens, qualitatively different to the intent classification task. Together, in SLU, the two sub-tasks contribute to a better representation of the semantics of an utterance than each one separately. In the next section we consider the joint task where both are addressed in one model.


\section{Joint intent and slot models} \label{sec joint}

The joint task marries the objectives of the two sub-tasks. As most papers point out there is a relationship between the slot labels we should expect to see conditional on the intent, and vice versa. A statistical view of this is that a model needs to learn the joint distributions of intent and slot labels. The model should also pay regard to the distributions of slot labels within utterances, and one would expect to inherit approaches to label dependency from the slot-labelling sub-task. Approaches to the joint task range from implicit learning of the distribution, through explicit learning of the conditional distribution of slot labels over the intent label, and vice versa, to fully explicit learning of the full joint distribution.

The joint task should also expect to inherit most of the issues of each sub-task and we find this is true. The mass of research now appears in the joint task and as such newer methods tried there have not been tried previosly in the sub-tasks.

\subsection{Major areas of research}

Research in the joint task has largely come from the personal assistant or chatbot fields. The chatbot is usually task-oriented within a single domain, while the personal assistant may be single or multi-domain.

Other areas to contribute papers are IoT instruction, robotic instruction (there is also a different concept of intent in robotics to describe what action the robot is attempting), and in vehicle dialogue for driverless vehicles. These areas also need to filter out utterances not applied to the device. 

Researchers have also drawn data from question answering systems, for example \cite{zhang2016joint} who annotated a Chinese question dataset from Baidu Knows.

\subsection{Overview of technological approaches}

In this section we give an overview of papers which focused on the joint task itself, measuring their efficacy largely on performance against state-of-the-art. Generally they are using new technologies as they became available, or new architectures to make the learning more explicit. 

\subsubsection{Classical methods}
The earliest work on the joint task used a tri-level CRF with the three layers being token features, slot labels and intent labels. \cite{jeong2008triangular} showed this architecture performed better than performing the two sub-tasks in a pipeline. Other early statistical models used a maximum entropy model (MEM) for intent and a CRF for slot labelling (\cite{wang2010strategies}), and a multilayer HMM (\cite{celikyilmaz2012joint}.

\subsubsection{Recursive neural networks}

The earliest attempt at a neural model to address the joint task was in \cite{guo2014joint} which used recursive neural networks (RecNNs) (different to recurrent neural networks (RNN)). RecNNs work over trees, in this case the constituency parse tree of the utterance, with leaves corresponding to the words (represented by word vectors). A neural network is applied at each node of the tree, recursively upwards to the root, computing a state for each node. At each node the states from children nodes are combined with a weight vector representing the node's syntactic type. Individual slot label classifiers are applied to each leaf using a combination of the word vectors of itself and its neighbours and the state vectors along the path from the leaf to the root. The state at the root is passed to an intent classifier. A combined loss over the slots and intent (and domain) is back-propagated. An optional post-processing, Viterbi decoded Markov layer is applied to the slots. Results were close to, but below, the then state-of the art for the tasks treated separately.

\subsubsection{Recurrent neural networks}

In 2016 the power of the RNN circuit for seq2seq tasks was explored in multiple papers. Features representing tokens are passed in temporal sequence to RNN units which have a hidden state. Intermediate hidden states may be used for slot labelling. The final hidden state is an embedding of the entire utterance and may be used for intent prediction. The classic encoder-decoder, which produces a sequential output, is the most commonly used architecture (\cite{liu2016attention}. Issues with the original RNN cells are addressed by LSTM and GRU cells. Bidirectional RNNs, where the input sequence is passed in in both forward and backward direction, address issues with unidirectional capturing of context.

Other architectures include (a joint loss is back-propagated unless mentioned):
\begin{itemize}
    \item a two layer LSTM with the top layer hidden states informing slot labelling and the first layer final state informing intent classification \cite{zhou2016hierarchical};
    \item the slot tagging task is softmax classifiers applied to the output of a simple BiLSTM using the concatenated hidden states. A special token is added to encapsulate the whole utterance for use in intent classification \cite{hakkani2016multi};
    \item a Bi-LSTM encoder decoder but with separate losses for intent and slot prediction (\cite{zheng2017intent};
    \item rather than seq2seq \cite{kim2017onenet} perform a global slot prediction (learning the joint distribution) from a matrix of the hidden states to a matrix of slot tag probabilities for each word, intent is predicted from a sum of hidden states;
    \item \cite{wen2018jointly} propose to use both a hierarchical (multi-layer) and a contextual (BiLSTM or LSTM) approach, investigating various combinations and using differing layers for intent and slot prediction;
    \item an ensemble using both BiLSTM and BiGRU fed to separate MLPs whose outputs are fused then projected and a softmax applied to predict intent and slots concurrently is proposed by \cite{firdaus2018deep}.
\end{itemize}

For RNNs the input is typically token feature by token feature in temporal sequence. \cite{hakkani2016multi} compared that to using context windows with superior results. However \cite{zheng2017intent} showed inferior results with context windows.

One critical observation made of many purely recurrent models is that the sharing of the information between the two sub-tasks is implicit. That is, while the sub-tasks are addressed jointly, it is often only through back-propagation of a joint loss.

\subsubsection{Attention}
Attention is an obvious technique for forcing an interaction between information from the two sub-tasks, in a learned way. Some attention constructions may still be seen as an implicit way of sharing information, but stronger methods start to force explicit learning.

In early papers a basic concept of attention used was the weighted sum of Bi-RNN hidden states as an input to slot and intent prediction (\cite{liu2016attention}). Then \cite{goo2018slot} used a stronger, more explicit attention. The base circuit is a BiLSTM taking word vectors in sequence and using a different learned weighted sum of the intermediate states of the BiLSTM for each slot prediction (the slot attention) and the final state for intent detection. The new addition is a slot gate which takes the current slot attention vector and combines it with the current intent vector in an attention operation. The output of the slot gate feeds the slot prediction. This circuit is an early example of intent2slot, a path through the circuit where intent prediction information is also fed explicitly to the slot prediction element. Another variation on intent2slot is provided in \cite{li2019conditional}.

\cite{qin2019stack} also use an intent2slot architecture but with BERT encoding and and using stack propagation. Rather than a gate like \cite{goo2018slot}, the intent detection itself directly feeds the slot filling. Also the intent detection is performed at token level and the final intent is taken by vote. \cite{wang2020sasgbc} too use an intent2slot gate with BERT embeddings.

\cite{yu2018novel} in a sense provide the dual approach to that of \cite{goo2018slot}, providing an attended slot prediction as the main input into intent prediction. The attention is additive on the weighted hidden states of a BiLSTM encoder and the weighted sum of the predicted slot labels. We call this explicit feed of slot information to the intent slot2intent.

\cite{zhang2019novel} extend the intent2slot gate of \cite{goo2018slot} with a pair of slot gates, one carrying the global intent information to the slot task, and one taking it to each slot location individually. \cite{zhang2019joint} also apply intent2slot but only to tokens determined to be not labelled `O'.

\cite{li2018self} introduce self-attention to the BiLSTM architecture to force a stronger learning ``at the semantic level'' between the slots and the intent. A first self-attention layer performs attention on word and convolutional character embeddings. This is concatenated with the word embeddings and fed to a BiLSTM layer. The final state informs intent detection. Self attention is performed between the intermediate states of the BiLSTM. This self attention is combined with the intent prediction which is then combined with the intermediate states to perform slot tagging. \cite{chen2019self} starts with word and character embeddings from a BiLSTM layer, then performs multi-head self-attention on these, followed by a BiLSTM encoder whose final state informs intent prediction. Another multi-head self-attention on the second BiLSTM hidden states, combined with the masked intent prediction, feed a CRF for slot prediction.

In \cite{chen2019wais} a BiLSTM layer takes the word inputs. State attention is performed as follows. For slots each hidden state is combined with the softmax of a weighted sum of all the hidden states passed through a feed forward network. Intent detection takes the last hidden state in combination with a similar weighted sum of the intermediate states. A similar formulation is used for word attention by weighting sums of word vectors rather than hidden states. All these features are combined in a fusion layer to inform the two tasks.

\cite{xu2020model} use a standard encoder-decoder LSTM which incorporates a length variable attention, that is attention of a sub-sequence of learned width over the hidden states.

\paragraph{Transformer}

The transformer architecture \cite{vaswani2017attention}, a non-recurrent model useful for capturing global dependencies via multi-head self-attention (among other strengths) appears in \cite{thido2019cross} to construct contextual embeddings of the word tokens. In their model attention is applied between all these to inform the intent prediction sub-task where it gives a superior result.

\cite{zhang2019using} also use the transformer architecture. They pass word embeddings to a 3 level transformer layer, then extract a global output to inform intent detection and token level output to pass to a CRF for slot detection. Differently to \cite{thido2019cross}, a special token is added to represent the whole utterance. Both these models only use the bidirectional encoder of \cite{vaswani2017attention}. 

\subsubsection{Hierarchical models}

A hierarchical model passes information learned to be relevant through ordered levels. While this flow is explicit it is often unidirectional. For example, \cite{lee2018coupled} supply a hierarchical approach with slot, intent and domain levels. Each element of the intent level is represented as the vector sum of the components in the slot layer coming from the same utterance. 

\cite{zhang2019capsule} provided relevant feedback from the highest level back to the lowest in their capsule network solution, a novel approach that sought to explicitly capture the words-slots-intent hierarchy. A capsule represents of a group of neurons whose output can be used for predictions at the next level; word capsules can be used to make slot label predictions, and so on. The hierarchy is learned using a routing-by-agreement mechanism: the prediction is only endorsed when there is strong agreement from the incoming capsule. The authors also propose a mechanism whereby a strong intent message at the highest level can be fed back to the earlier levels to help them in their task. This explicit and direct feedback is stronger than the implicit or indirect joint learning typically found in RNN  models. \cite{staliunaite2020capsule} extended this work to a multi-task setting with extra mid-level capsules for NER and POS labels, with mixed results.

\onecolumn
\begin{longtable}[t]{ p{3.3cm} p{5.7cm} p{7cm} } 
\caption{Joint task papers reviewed with addressed issue, approach and techniques} \\

\hline
\textbf{Paper} & \textbf{Addressed issue} & \textbf{Approach}  \\

\hline
\cite{jeong2008triangular} & Joint solution of related tasks & Tri-layer CRF, extra layer for classification \\ \hline

\cite{wang2010strategies} & Small training sets & MEM and CRF, joint task versus pipeline \\ \hline

\cite{celikyilmaz2012joint} & Small training sets & Tri-level HMM, bolstered features \\ \hline

\cite{xu2013convolutional} & Automated feature creation & CNN features into TriCRF \\ \hline

\cite{guo2014joint} & Incorporate discrete constituency parse of utterance & RecNN on word vecs and parse tree  \\ \hline

\cite{shi2015contextual} & Context from multi-turn dialogue & RNN (token) and CNN (sentence) features, MLP\\ \hline

\cite{zhou2016hierarchical} & Hierarchical task relationship & RNN, LSTM \\ \hline

\cite{hakkani2016multi} & Seq2seq, joint model, architectures & BiLSTM \\ \hline

\cite{chen2016syntax} & Incorporate language knowledge & K-SAN attention network, GRU \\ \hline

\cite{zhang2016joint} & Apply RNN to intent & GRU \\ \hline

\cite{liu2016attention} & Employ encoder-decoder with attention & Encoder-decoder with attention \\ \hline

\cite{liu2016joint} & Real time analysis & LSTM, MLP \\ \hline

\cite{zheng2017intent} & NLP in navigation dialogue & BiLSTM encoder decoder, seq2seq \\ \hline

\cite{ma2017jointly} & No long term memory, linearity & LSTM, sparse attention \\ \hline

\cite{kim2017onenet} & Error propagation, information sharing between tasks & Word and character RNN embedding \\ \hline

\cite{yang2017endtoend} & Noisy NLU outputs & Dialogue act unit after NLU \\ \hline

\cite{goo2018slot} & Learn relationship between slot and intent attention vectors & Slot gate, BiLSTM \\ \hline

\cite{pan2018multiple} & Multiple utterance dialogue & Utterance to utterance attention \\ \hline

\cite{wen2018jointly} & Using hierarchy and context & Two layer (Bi)LSTM \\ \hline

\cite{wang2018chinese} & Capturing local semantic information & CNN, BiLSTM encoder decoder \\ \hline

\cite{firdaus2018deep} & Domain dependence & Ensemble model, GRU \\ \hline

\cite{shen2018user} & Slow training time & Progressive multi-task model using user information \\ \hline

\cite{li2018joint} & Correlation of different tasks & Multi-task model incl. POS tag \\ \hline

\cite{li2018self} & Sharing semantic information & Self-attention \\ \hline

\cite{zhang2018attention} & Tagging strategy & Token tags include intent and slot \\ \hline

\cite{zhang2019capsule} & Hierarchical structure & Capsule network with rerouting (feedback) \\ \hline

\cite{zhao2018joint} & Spatial (context) and serial (order) information & Encoder-decoder, CNN \\ \hline

\cite{wang2018bimodel} & slot2intent and intent2slot & Bi-directional architecture \\ \hline

\cite{siddhant2018unsupervised} & Unsupervised learning & ELMo on unused utterances, BiLSTM \\ \hline

\cite{yu2018novel} & Use sequence labelling output for intent & Cross attention, BiLSTM, CRF \\ \hline

\cite{lee2018coupled} & Hierarchical vector approach & Learn vectors representing elements of frame \\ \hline

\cite{jung2018learning} & Model relationship between text and its semantic frame & Vector representation of frame \\ \hline

\cite{ray2018robust} & Rare, OOV words & Paraphrasing input utterances \\ \hline

\cite{liu2019cm} & Unidirectional information flow & Memory network \\ \hline

\cite{shen2019interpreting} & Poor generalisation in deployment & Sparse word embedding (prune useless words) \\ \hline

\cite{ray2019interative} & Slots which take many values perform poorly & Delexicalisation \\ \hline

\cite{wang2019effective} & Language knowledge base, history context & Attention over external knowledge base, multiturn history \\ \hline

\cite{li2019conditional} & Implicit knowledge sharing between tasks & BiLSTM, multi-task (DA) \\ \hline

\cite{gupta2019simple} & Speed & Non-recurrent and label recurrent networks \\ \hline

\cite{gupta2019casa} & Multi-turn dialogue, using context & Token attention, previous history \\ \hline

\cite{chen2019self} & Capturing intent-slot correlation & Multi-head self attention, masked intent \\ \hline

\cite{chen2019bert} & Poor generalisation & BERT \\ \hline

\cite{bhasin2019unified} & Learning joint distribution & CNN, BiLSTM, cross-fusion, masking \\ \hline

\cite{thido2019cross} & Lack of annotated data, flexibility & Language transfer, multitasking, modularisation \\ \hline

\cite{zhang2019novel} & Key verb-slot correlation & Key verb in features, BiLSTM, attention \\ \hline

\cite{zhang2019using} & Learning joint distribution & Transformer architecture \\ \hline

\cite{daha2019deep} & Efficient modelling of temporal dependency & Character embedding and RNN \\ \hline

\cite{dadas2019deep} & Lack of annotated data, small data sets & Augmented data set \\ \hline

\cite{chen2019wais} & Learning joint distribution & Word embedding attention \\ \hline

\cite{e2019novel} & Learning joint distribution & Bidirectional architecture, feedback \\ \hline

\cite{zhang2019joint} & Poor generalisation & BERT encoding, multi-head self attention \\ \hline

\cite{qin2019stack} & Weak influence of intent on slot & Use intent prediction instead of summarised intent info in slot tagging \\ \hline

\cite{gangadharaiah2019joint} & Multi-intent samples & Multi-label classification methods \\ \hline

\cite{firdaus2019multi} & Multi-turn dialogue history, learning joint distribution & RNN, CRF \\ \hline

\cite{pentyala2019multi} & Optimal architecture & BiLSTM, different architectures \\ \hline

\cite{castellucci2019multi} & Non-recurrent model, transfer learning & BERT, language transfer \\ \hline

\cite{schuster2019crosslingual} & Low resource languages & Transfer methods with SLU test case \\ \hline

\cite{okur2019natural} & Natural language & Locate intent keywords, non-other slots \\ \hline

\cite{xu2020model} & Only good performance in one sub-task & Joint intent/slot tagging, length variable attention \\ \hline

\cite{bhasin2020parallel} & Learning joint distribution & Multimodal Low-rank Bilinear Attention Network \\ \hline

\cite{firdaus2020deep} & Learning joint distribution & Stacked BiLSTM \\ \hline

\cite{zhang2020graph} & Limitations of sequential analysis & Graph representation of text \\ \hline

\cite{wang2020new} & Non-convex optimisation & Convex combination of ensemble of models \\ \hline

\cite{wang2020sasgbc} & BERT issues with logical dependency (I before B) & CRF and self attention over BERT \\ \hline

\cite{ni2020NLUIoT} & Model transfer, IoT & Pipeline structure from medical analogue \\ \hline

\cite{krone2020learning} & Unseen labels & Few-shot meta-learning \\ \hline

\cite{bhatiya2020metalearning} & Unseen labels, language transfer & Few-shot meta-learning \\ \hline

\cite{tang2020endtoendmask} & Linear chain CRF limitations & GCN based CRF \\ \hline

\cite{staliunaite2020capsule} & Extend capsule network & Capsule network with MTL \\ \hline


\hline
\end{longtable}


\twocolumn

\subsubsection{Bi-directional models}

A model where there is a pipeline from one sub-task to the other may be seen as unidirectional. A bi-directional model, different to the bi-directionality seen in RNNs, has an explicit path from slot processing into intent prediction and also from intent processing into slot prediction. This can form two parallel paths through the circuit, often with a fusion layer or a joint loss.

\cite{wang2018bimodel} proposes the first such bi-directional circuit. In this paper each path is a a BiLSTM and the hidden states from each path are shared with the other, another form of explicit influence between the tasks. An optional LSTM decoder is supplied on each side. Interestingly the loss is not a joint loss but the circuit alternates between predicting intent for a batch, and back-propagating intent loss, then predicting slots for the same batch and back-propagating slot loss. They call this asynchronous training.

\cite{bhasin2019unified} also uses bi-directional paths. Starting with GloVe word embeddings, an intent path converts them to convolutional features which are concatenated then projected. The slot path passes the word vectors through a BiLSTM with a CRF on top with the results also projected. Three types of fusion of the paths (after reshaping/broadcasting) were tested: addition, average or concatenation.

\cite{e2019novel} also consider bi-directionality. They start with a BiLSTM encoder. A weighted sum of intermediate states for each step (the slot contexts) feeds a slot sub-net, while the weighted final hidden state (the intent context) feeds an intent sub-net. These two interact in either a slot2intent fashion (slot affects intent) or intent2slot. The outputs then feed a softmax intent classifier and a CRF respectively. In slot2intent mode a learned combination of the slot contexts and intent context then feed the intent sub-net, where they are combined with the intent context for prediction. In intent2slot mode the intent context is combined with the slot contexts to form a slot informed intent context. This is then fed to the slot sub-net where it is combined with the slot contexts to feed the CRF for prediction. As may be expected intent2slot gave better slot results and slot2intent gave better intent results. That only one can be applied is a weakness of the architecture.


\subsubsection{Memory networks}
\cite{liu2019cm} consider that even with the inclusion of feedback that the circuit of \cite{zhang2019capsule} is still overly unidirectional. To overcome this they consider the use of memory networks to the joint task. As they see the typical interaction as a pipeline from words to slots to intent, they alternate interaction from slots to intent and vice versa via multiple blocks of memory nets. The network begins with GloVe word embeddings and max pooled convolutional character embeddings. These feed the first memory block, which constructs slot features, intent features and hidden states. Further memory blocks in the stack take the previous block's hidden states as inputs. The memory blocks perform three operations, which also strive to capture local context and global sequential patterns:

\begin{itemize}
    \item Deliberate Attention: a slot memory (with number of cells equal to number of slot labels) and intent memory (ditto for number of intent labels) are randomly initialised then updated. At each word position each memory is updated as a weighted sum of the other memory and of the block hidden states for the current word. Diffusion of influence between slots and intents thus takes place and can inform the hidden states for the next word.
    \item Local Calculation: this is a recurrent process receiving the input embeddings or previous block's hidden states. It calculates slot representation and intent representations as interactions between its inputs and the slot and intent memories. It is an LSTM network. 
    \item Global Recurrence: a BiLSTM layer on top which encodes global sequential interactions.
\end{itemize}

After the stacked blocks a final prediction takes place. Slots are labelled via a CRF on the final hidden states and slot representations. Intent is via an average of the final hidden states and intent representations. 

\subsubsection{Meta-studies of flow architectures}

In an approach which considers both feature creation and architecture \cite{pentyala2019multi} give an interesting generalisation of multi-task learning architectures then apply it to the joint task. For example a three sub-task parallel architecture would take samples with training labels for each sub-task, develop universal features, task specific features, and grouped features, concatenate them and then feed them to task specific decoders. Series architectures are also given. Their base circuit uses word and character embeddings and is a standard BiLSTM encoder feeding an LSTM decoder for slots and a softmax classifier on the final hidden states for intent. No attention or slot gating occur. The base circuit is then adjusted to match some of the series and parallel architectures. \cite{firdaus2020deep} also look at varieties of series architectures from multi-task circuit design.

\subsubsection{Graph networks}

Graph networks can be used to address shortcomings of limited context windows suffered by RNNs and CRFs, as they can learn global relationships between words and labels.

\cite{zhang2020graph} use a graph S-LSTM network to overcome perceived shortcomings of RNNs, being lack of parallelisation (due to sequential nature), weak local context use, and lack of long range detection. The graph has as nodes the word representations and sentence representation from an LSTM, hence the network simultaneously works on the whole sentence. Only word nodes within a context window are connected by edges. The sentence node is connected to all word nodes. Messages are passed between the nodes to enable global coordination. The final node states for each slot go through a convolution unit and self attention before being used for slot filling. The final sentence node state is used directly for intent detection.

\cite{tang2020endtoendmask} see shortcomings of linear chain CRFs as being limited context and only applicable to the slot sequence. They construct a graph based CRF graph convolutional network which learns relationships between words, slot labels and intent labels. BERT embeddings are passed through a BiLSTM which feed the GCN for prediction. A weighted joint loss is back-propagated.

\subsubsection{Importing methods from analogous fields}

\cite{bhasin2020parallel} propose an interesting analogy; that the relationship between intent and slots is similar to that between the query and image in visual question answering. Thus they borrow an idea from the latter field - Multimodal Low Rank Bilinear (MLB) fusion, between the features of each part.

\cite{ni2020NLUIoT} also propose an analogy with the joint task of clinical domain detection and entity recognition in medical literature. Coming from the IoT field they also propose a pipeline structure where intent is detected first and then slots determined in a closed domain setting.

\subsection{Feature creation and enhancement}

As discussed in the sub-task sections, feature creation is a critical part of the design of circuits in NLU as it ideally should capture, at least, semantic information of the individual tokens, their context, and of the entire sentence. Then, any other information that may be used to enhance the result may be considered, including meta-data and syntactic information.

\subsubsection{Token embedding}

The earliest models used features familiar from methods like POS tagging and containing one-hot word embedding, n-grams, affixes etc. (\cite{jeong2008triangular}). \cite{celikyilmaz2012joint} incorporated entity lists from sites such as IMDB (movie titles) or Trip Advisor (hotel names).

Neural models enable the embedding of diverse natural language without such feature engineering. The first neural features were convolutional embeddings of the utterance words in \cite{xu2013convolutional}, which fed to a statistical model after \cite{jeong2008triangular}. \cite{shi2015contextual} was the first to use RNN based token embeddings but also combined those into a CNN based sentence embedding.

\cite{ma2017jointly} used for input at each step a convolution of the current word and the previously predicted slot labels. \cite{wang2018chinese} used multiple convolutional features of the embedding words but also maintained the order of the words within the convolutions. These were then fed to an RNN layer.

The gamut of word embedding methods have been used including word2vec (\cite{pan2018multiple, wang2018chinese}), fastText (\cite{firdaus2020deep}), GloVe (\cite{zhang2016joint, liu2019cm, dadas2019deep, okur2019natural, bhasin2019unified, thido2019cross, pentyala2019multi, bhasin2020parallel}), ELMo \cite{zhang2020graph} and \cite{krone2020learning} (pre-print only), BERT (\cite{zhang2019joint, qin2019stack, ni2020NLUIoT} and \cite{chen2019bert, castellucci2019multi, krone2020learning} (pre-print only). 
\cite{firdaus2018deep} and \cite{firdaus2019multi} used concatenated GloVe and word2vec embeddings to capture more word information.

While BERT displays impressive performance, \cite{wang2020sasgbc} identify a limitation (logical dependency for slot filling) and counter it by feeding it to an intent2slot gate, an attention layer and a CRF.

\cite{gupta2019simple} tested ten different word contextualisation embeddings from four different method groups (feed forward, CNN, attention, LSTM) with different depths.

\cite{kim2017onenet} were the first to use a combination of character and word embedding. Others also used this (\cite{liu2019cm, chen2019self, firdaus2019multi, pentyala2019multi}. On the other hand, \cite{daha2019deep} use only character embedding.

Pre-computed syntactic features, for example POS tags for each token using the nltk library (\cite{firdaus2018deep} have been included with word embeddings.

\cite{zhang2019novel} take from the service robotics field the importance of a key verb in an instruction in informing the slot labels. The key verb is deduced from a dependency parsing. A feature is constructed from the training data to encode a priori dependencies between words and key verbs. The circuit takes the key verb feature and concatenates it with each word's one hot encoding. These are passed to a BiLSTM layer to produce token embeddings.

\subsubsection{Sentence embedding}
The use of the final hidden state in an RNN  as the sentence embedding was used frequently (\cite{zhou2016hierarchical, liu2016attention, wang2018chinese}). Sentences were also embedded by using a special token for the whole sentence in \cite{hakkani2016multi, zhang2019using}, as a max pooling of the RNN hidden states (\cite{zhang2016joint}, as a learned weighted sum of Bi-RNN hidden states (\cite{liu2016attention}), as an average pooling of RNN hidden states (\cite{ma2017jointly}), as a convolutional combination of the input word vectors (\cite{zhao2018joint, bhasin2019unified}, and as self-attention over BERT word embeddings (\cite{zhang2019joint}).

\cite{ma2017jointly} also apply a sparse attention mechanism which evaluates word importance over a batch and applies weights within each sample utterance for the intent detection.

\cite{daha2019deep} used an extra <TAGG> token after the end-of-sentence <EOS> token for sentence encapsulation and see better intent prediction performance. \cite{okur2019natural} encode both a <BOU> and <EOU> token at the beginning and end of the utterance in their BiLSTMs.

\subsection{Target variations}

The targets are typically the annotated intent and slot labels. \cite{zhang2018attention} construct a single tag for each token which incorporates the slot tag and the sentence intent. Their circuit then just performs a single seq2seq task and the sentence intent is deduced by a majority vote of the intent portion of the predicted tags. \cite{xu2020model} use the same single tag set. \cite{qin2019stack} perform the intent detection at token level though separate to the slot prediction, and the final intent is taken by vote.

\cite{lee2018coupled} works with learned embeddings of slot labels, intents and domains where the sum of slot label embeddings for an utterance is close to the intent embedding in vector space. A network can then be trained to map tokens to vectors close to the slot labels and intent for the utterance.

\cite{jung2018learning} proposes a vector embedding of the entire semantic frame (intent, slot labels, slot values) as the target. In training the utterance and the semantic frame are input and vectorised. A semantic frame vector is output. The distance between the output vector and input frame vector is minimised. In testing the text is input and a vector is output and the nearest semantic frame vector is chosen.

\cite{okur2019natural} proposed an extra token tag for intent keywords, for example the word ``play'' in an utterance with intent \textit{PlayMusic}. In one of their models only intent keywords and non-Other slot tokens contribute to intent detection.

\subsection{Issues addressed and solutions proposed}

\subsubsection{Narrowness of approach}
The use of features constructed only from the tokens in the sentences may be too narrow an approach. External knowledge about the words' places in the language, or the syntactic structure of the sentence, or of co-occurrence statistics amongst word and labels may aid the task. Methods to incorporate extra elements have been developed.

\paragraph{Knowledge bases}

Knowledge bases are constructs containing information or statistical priors that may be useful to the task at hand. They may be constructed independent of the task, or as a preliminary step using information from the training data. They have been used for feature construction, as features themselves, and to be consulted via attention.

\cite{chen2016syntax} was the first to use an extra knowledge base to inform the joint task. They use a K-SAN input, being a structured knowledge network. Two K-SANs are constructed, one taking a dependency parse of the utterance (syntactic), and the other an Abstract Meaning Representation (AMR) graph (semantic). Each representation is tested separately. A CNN encodes the representation into a vector, while a separate CNN encodes the sentence itself into another vector. Attention is applied between the two vectors and the results combined to give a ``knowledge guided representation'' of the utterance. This is included as an input to a GRU RNN cell along with the word encodings in sequence. A second RNN just takes the utterance words as input. A weighted sum of the hidden states of the two RNNs is used for prediction.

\cite{wang2019effective} incorporate the ConceptNet\footnote{http://conceptnet.io} framework as a knowledge base source. (Head, Relation, Tail) triples are extracted for each word in the utterance. The TransE model (\cite{bordes2013transe}) for embedding multi-relational data is used to encode the knowledge. Attention is applied between words and the knowledge base encoding.


\cite{qin2020agif} capture the interaction between multiple intents, and slots, with a graph representation. For multi-intent  a score is calculated for each intent and those above a threshold are returned. The graphs use graph attention networks. Tokens are encoded by a BiLSTM and then multiple intents are predicted. These slot path takes the token embeddings through an LSTM which provides a feature for each token which interacts with the intent predictions and the slot-intent graph to make slot predictions.

The inclusion of knowledge embedded in graph representations, or networks that perform tasks on such graphs has borne fruit in the very recent literature. Further research in this area could include other types of such graphical representations and incorporate information not just from the current training set or external knowledge bases but some combination of the two, or data from several training sets.

\subsubsection{Multi-turn dialogue}

Typically in NLU only the current single utterance is analysed. Temporal information or previous utterance context or previous dialogue action are not considered. However as noted in the intent and slot sections using such information in the model can lead to better performance.

There are multiple data sets available which contain a multi-turn dialogue around a single intent or set of related intents. In these cases incorporating the history from previous turns can be incorporated. \cite{shi2015contextual} fed a sentence embedding along with the predicted intent and domain labels of previous turns into the intent prediction for the current turn. \cite{pan2018multiple} calculate attention between the BiGRU embeddings of successive utterances which make up a single sample and contribute to a single intent. \cite{wang2019effective} similarly use attention between the BiLSTM encoding of each utterance to the previous utterances in the history.

\cite{gupta2019casa} look at multiple contextual inputs in multi-turn dialogues for the current utterance. For the current utterance they apply token2token attention and sentence2token attention at the input. Information from previous turns, including intents, slots and dialogue actions can then be attached.

While it is sensible for the research to focus on single utterance analysis it should be noted that SLU devices are often listening to all dialogue, filtering out-of-domain utterances using methods discussed in Section \ref{sec emergeintent}, and that incorporating lead in dialogue can be useful to the joint task.

\paragraph{Multi-task learning}

Looking for synergies with related tasks has been an approach in the two sub-tasks and has been actively applied in the joint task. As described earlier the full semantic frame contains three levels - domain, intent and slots. Simultaneously solving the domain with the other layers has been explored \cite{shi2015contextual, hakkani2016multi}.

\cite{shen2018user} introduced an extra task to predict tags for known user information from metadata (for example location, timestamp). The metadata task is preliminary and thus informs the BiLSTM word embedding. The results of the preliminary task feed the regular joint task training and the BiLSTM word embeddings are updated.

\cite{li2018joint} works on the theory that adding a further sequential task (POS tagging) will aid the joint tasks. A single LSTM layer takes word embeddings and performs an intent and slot prediction at each step, feeding those predictions with the LSTM hidden state to a next-word POS tagger. A joint loss across all tasks is calculated. The results show that the extra task helps improve intent detection.

\cite{yang2017endtoend} claim that noisy SLU output can be mitigated by making it part of an end-to-end network including dialogue action prediction in the dialogue manager, with errors back-propagating from the dialogue manager refining the NLU prediction. The hidden states of a BiLSTM SLU model also feed a second BiLSTM which performs the dialogue action prediction. A joint loss across all tasks is back-propagated. In related work, \cite{li2019conditional} also tied together an SLU network and a network to predict the next dialogue action. They use a stronger NLU segment to improve overall results. A joint loss across intent, slots and actions was back-propagated and performance exceeded the SLU model alone. \cite{gupta2019casa} use dialogue action in a multi-turn data set. \cite{firdaus2020deep} incorporate dialogue action, typically as the first task in a multi-task pipeline, rather than the last.

\cite{staliunaite2020capsule} incorporated POS and NER tagging simultaneously with slot tagging and intent detection using a capsule network, however the results were generally poorer when both NER and POS were included rather than just one, and mixed for different data sets indicating a generalisability issue.

The method of using SLU as fine-tuning with pre-training on another task, or vice versa, has shown improvements in the SLU performance. However the results of \cite{staliunaite2020capsule}, echoing those of \cite{louvan2019leveraging} on slot tagging, indicate a parsimonious approach to adding extra tasks simultaneously more often yields a better result.

\subsubsection{Generalisability}

\paragraph{Domain dependence}
An issue found is that a model trained successfully on one domain or data set does not perform as well on a different domain or data set, implying it has simply learned statistical properties of the training data set. One issue suggested by \cite{firdaus2018deep} is that the language in the data sets is not particularly ``natural''. Though their ensemble model with syntactic POS features performed well on ATIS it is unclear it generalised to a second data set.

\cite{firdaus2018deep} propose to design a domain invariant model by using an ensemble of word embeddings in an ensemble circuit with a BiGRU unit and a BiLSTM unit. While together they outperform each unit used alone, the circuit didn't transfer well to a new dataset. This approach of using multiple methods in one circuit for generalisability appears to rely too much on chance than good design.

\cite{shen2019interpreting} looked at the drop off of performance of state-of-the-art architectures when deployed. Some issues that cause drop off in performance are personalised language of users not matching the training data, and the cost of annotated training sets (and hence their limited size and spread). Focusing on the vocabulary they propose a sparse vocabulary embedding which they apply to two existing architectures and show improved results. The embedding uses lasso regularisation to penalise words useless to the tasks. They apply the method to the networks of \cite{liu2016attention} and \cite{goo2018slot} and find that while using sparse vocabulary that intent accuracy increases but slot f1 decreases. They qualitatively discuss these results with observations on what words/structures help the two sub-tasks and the joint task.

\cite{zhang2019joint} use BERT encoding, claiming that a pre-trained model should address the poor generalisability of models that perform their own embedding. They use a two step decoder where the first step decodes intent which feeds the intent classifier and also the second decoder which works on slot labelling. The intent decoder performs multi-head self attention on the BERT encodings. In the slot decoder a BERT embedding for a word is concatenated with the attended intent in training only if it is a ``real slot'', i.e. non-'O'; otherwise it is concatenated with a random vector. Each concatenation feeds a softmax classifier for the token. A joint loss is back-propagated. The results are good for both ATIS and SNIPS.

\paragraph{Non-English data and transfer learning}

NLU is eventually required in many languages, most of which do not have the large annotated training datasets required. An aspect of generalisability of models is thus whether they can be used outside the language on which they are trained.

Papers have used the same architecture for both English and non-English data sets to give comparative studies across languages. \cite{jeong2008triangular} used ATIS and a Korean banking dataset. \cite{zhang2016joint} used ATIS and Chinese questions collected from Baidu Knows. \cite{pan2018multiple} work only with a Chinese data set where word boundaries are not clearly identified. 

Other papers considered the transfer of the model from English to other languages to address lack of annotated data in those languages. \cite{thido2019cross} consider a simple weight transfer from an English model for use in German. \cite{castellucci2019multi} (pre-print only) consider transfer learning from English to Italian.

\cite{schuster2019crosslingual} study transfer to low resource languages, in this case from English to Spanish and Thai. The circuit is a basic BiLSTM with CRF. They evaluate three different cross-lingual transfer methods: (1) translating the training data, (2) using cross-lingual pre-trained embeddings (CoVE), and (3) using a multilingual machine translation encoder as contextual word representations. They find that using cross-lingual transfer well outperforms training on limited data from the low resource language. The work is extended by \cite{liu2019zerocross}, \cite{bhatiya2020metalearning} and \cite{qin2020cosdaml} but moves into cross-lingual transfer theory and out of the scope of this survey.

This issue of generalisability is still very much open and in demand by endusers. Methods discussed in Section \ref{sec lack} for using few-shot methods to boost performance of existing models in new domains or data sets warrant further investigation.

\subsubsection{Limited training data}

Annotated training data is costly in time and resources to produce. With new domains and applications for SLU appearing, with existing domains changing, and with colloquial language shifting, there is a need for methods to perform well with limited training data. 

\paragraph{Small data sets}

In an early statistical model \cite{wang2010strategies} test two-pass (pipeline, intent then slot) versus one-pass (simultaneous solving) for a small training set. They show that intent classification is much better in the two pass model while token level slot f1 suffers slightly. \cite{tam2015rnnbased} proposed using an RNN network to learn the word/label dependency distributions from available training data. For intent, the intent label is attached to each word in an utterance. Synthetic samples are then generated for use in training. They showed that this can lead to better results for slot tagging using a CRF on three data sets but that the results for intent were inconsistent.

\cite{dadas2019deep} propose a data set augmentation scheme which generates new training samples from existing ones via three methods: labelled word replacement from an external synonym lexicon; random replacement of outside words with a synonym; and ``sequence order mutation'' - change of order of spans for utterances with one labelled span. They showed that augmentation can improve the slot f1 result, more so for smaller data sets, but has little effect on intent accuracy. There is a further literature on data set augmentation for SLU which we will not cover here.

\paragraph{Lack of annotated data} \label{sec lack}

As new domains appear it takes time and cost to develop annotated data sets for training. \cite{shen2018user} address this by training on user metadata as a preliminary step. They show they can achieve higher slot f1 scores on smaller training sets and with less epochs than only using the intent and slot annotations. \cite{siddhant2018unsupervised} construct an unlabelled utterance data set collected from ASR interactions with their agent. They train an ELMo style word embedding on this data set. For the joint task they find their embedding outperforms fastText. As well as language transfer, \cite{bhatiya2020metalearning} also address the transfer to new label domains with minimal samples available via a few-shot meta-learning approach.

\paragraph{Unseen labels}

\cite{krone2020learning} (pre-print only) address the issue of unseen test classes by applying two few-shot algorithms: model agnostic meta-learning (MAML) and prototypical networks, in combination with three word embeddings - GloVe, BERT and ELMo. They find the prototypical network algorithm performs best, that joint training significantly improves slot filling span based F1, and that ELMo and BERT share the spoils from the word embeddings.

\subsubsection{The OOV issue}
Out-of-vocabulary words in the test set, that is words that do not appear in the training set, may lead to lower test performance. Similarly the use of rare words in the training set may introduce unwanted bias. This issue is related to generalisability and also to changing vocabulary from user to user, or over time.

\cite{zhang2016joint} set all words that only appear once in the training set to an unknown UNK token. Then new words in the test set are also set to the UNK token. They also replace all numbers with a generic DIGIT token. This is also applied by \cite{li2018self} and \cite{zhang2020graph}. \cite{ray2018robust} perform a paraphrasing of input utterances to cater for rare or OOV words, or for unusually phrased requests. The paraphrasing is performed by an encoder-decoder RNN and is performing a kind of translation. The paraphrase can be applied to any downstream model. \cite{chen2019bert} propose BERT embeddings as a sop to rare or OOV words. BERT uses word-piece encoding to provide a meaningful embedding for all words.

\cite{ray2019interative} address the issue of networks having trouble with slots with large semantic variability - that is, there are many values the slot can take during training and many unseen values during testing/deployment. They call these out-of-distribution (OOD) slots. They propose a new delexicalisation method. This replaces values in OOD slot locations with default values in pre-processing.

\subsubsection{Obfuscation and speed}
Taking a contrary view, \cite{gupta2019simple} consider how joint modelling may obfuscate, or hide, information and may also be unnecessarily slow. They propose a modularised network with separated tasks after a common word contextualisation pre-processing. The modularisation enables easier analysis of results. They perform speed analysis within their model suite.

\cite{wang2020new} propose a convex combined multiple model approach to counter limitations of non-convex optimisation, one of which is slow speed of convergence due to being stuck near non-optimal solutions. Each network in the circuit has the same structure but different initialised weights. A convex combination of label predictions from each network is used as the label prediction for each slot and the intent. Both a local loss function for each network and a global loss function on the combination are back-propagated. The networks are BiLSTMs with a context layer. The convex combination outperforms single classifiers. The speed improvements are significant.

\subsubsection{Real time learning}
In \cite{liu2016joint}, the authors consider real time analysis where the whole utterance isn't analysed but a prediction is made at each time step. In this RNN the intent is predicted at each step and used as context to the slot prediction (as well as a next word language model). Thus the current slot prediction is conditional on the input words to that point, the previous slot predictions and the previous intent predictions. The recurrent unit is an LSTM but the current intent and slot predictions use MLPs on the current hidden state.

\subsubsection{Label dependency}
This is an issue covered in the slot filling section and the methods used there including CRFs and encoder-decoder seq2seq models have been used in the joint task. We note further that the use of CRFs after a deep learning solution became popular again from 2018 (see Table \ref{tab:overviewjoint}) to counter this issue (\cite{yu2018novel, zhang2019using, e2019novel, firdaus2019multi}). \cite{wang2020sasgbc} use a CRF to counter a label dependency limitation for slot-filling in using BERT due to its non-recurrent nature.

\cite{chen2019self} claim earlier models do not perform slot filling realistically enough (so reflecting the language priors) nor explore intent-slot correlation well. They propose to use a CRF for the former and a masked intent prediction as an input to the CRF for the latter. The mask is ``a conditional probability distribution of slot given intent, obtained from training data''. \cite{bhasin2019unified} also use a CRF with masking, prior conditional probabilities of slot/intent co-occurrence obtained from training data, for slot prediction.

\subsubsection{Handling multi-labels}
The multi-label issue was addressed for intent classification in Section \ref{sec intmulti}. With the move into neural models similar methods have been applied.

\cite{wen2018jointly} simply removed multi-label samples from their data set. \cite{dadas2019deep} tried using both the first label as the only label, and merging labels to a compound label. \cite{qin2020agif} consider multi-intent data sets, including their own extension of SNIPS to multi-intent. For multi-intent a score is calculated for each intent and those above a threshold are returned.

\cite{gangadharaiah2019joint} studied both sentence level and token level multi-intent detection. For ATIS, they split the compound multi-labels, giving about 2\% of the data set with multi-labels. They also use an internal data set with 52\% of the samples having multi-labels. Although the assignment method is unclear a sentence may be assigned multi-labels during prediction, and these are then assigned to individual tokens in the sentence to aid with slot filling.

\section{Data sets} \label{sec dataset}

\begin{table*}[t]
  \caption{Major data sets used in the literature, single turn in English unless noted, Train-Val-Test gives the number of utterances}
  \label{tab:data set}
  \begin{tabular}{cccccl}
    \toprule
    Name & Public & Train-Val-Test & Num Intents & Num Slots & Domain,Notes\\
    \midrule
    ATIS & Y & 4478/500/893 & 21 & 128 & air travel\\
    SNIPS-NLU & Y & 13084/700/700 & 7 & 72 & personal assist.\\
    FRAMES & Y & 20006/-/6598 & 24 & 136 & hotel, multiturn\\
    CQUD & N & 3286 & 43 & 20 & Chinese, question answering\\
    TREC & Y & 5500/-/500 & 6(50) & - & question classification\\    
    TRAINS & N & 5355/-/1336 & 12 & 32 & problem solving, multiturn\\    
    Microsoft Cortana & N & 10k/1k/15k & 10-20 & 15-63 & personal assist., multi-domain\\ 
    Facebook & Y & 30521/4181/8621 & 12 & 11 & multi-lingual task oriented\\
    SRTS FrameNet & N & 2803/-/312 & 12 & 61 & robotics\\  
    Alexa & N & 264000/-/- & 246 & 3409 & 17 domains\\  
    DSTC2 & Y & 4790/1579/4485 & 13 & 9 & multi-turn, restaurant search\\  
    DSTC4 & Y & 5648/1939/3178 & 87 & 68 & multi-turn, Skype tour guide dialogues\\  
    DSTC5 & Y & 27528/3441/3447 & 84 & 533 & dialogue with social robots\\  
    CMRS & N & 2901/969/967 & 5 & 11 & Chinese, meeting room reservations\\  
    CU-Move & N & 57584/-/- & 5 & 38 & in-vehicle dialogue\\  
    AMIE & N & 3418/-/- & 10 & 7 & in-vehicle dialogue\\  
    TeleBank & N & 2238/-/- & 25 & 17 & Korean, banking\\
    MIT MOVIE\_ENG & Y & 8798/97/2443 & - & 25 & movies, slot only\\
    MIT RESTAURANT & Y & 6894/766/1521 & - & 17 & restaurants, slot only\\
\bottomrule
\end{tabular}
\end{table*}

\subsection{Introduction}

A summary of the most commonly used data sets is presented in Table \ref{tab:data set}. Here we cover the most commonly used data sets, ATIS and SNIPS, popular due to their easy availability and ubiquity of use allowing comparison between models. We then briefly cover the other data sets.

\subsection{The Air Travel Information System (ATIS)}
ATIS was introduced in 1990 in \cite{hemphill1990atis} and its history is instructive in understanding some of the conventions of the field. The domain is air travel information including ``information about flights, fares, airlines, cities, airports, and ground services''. The first release, ATIS-0, collected 740 evaluable samples. Each sample contained a sound file of a single utterances question, a transcription of the question, a set of tuples constituting the answer, and the SQL query that produced the tuples. 

Tokens were generated according to Standard Normal Orthographic Representation (SNOR) rules: whitespace-separated lexical tokens, case insensitive alphabetic text, spelled letters are represented with the letter followed by a fullstop (e.g., “a. b. c.”), no non-alphabetic characters (except apostrophes for contractions and possessives and hyphens for hyphenated words and fragments). The average length of the SNOR translated utterances was 11.3 tokens. 

 Extensions to the data set were made available in subsequent years ATIS-1 (\cite{pallett1992atis1}, ATIS-2 (\cite{hirschman1992atis2}) and ATIS-3 (\cite{dahl1994expanding}) in late 1993 to mid-1994.

The set which evolved to become the standard ATIS for NLU analysis was drawn from the annotated samples in ATIS-2 and 3. \cite{he2003datadriven} were the first to use the combined set for language understanding. \cite{raymond2007generative} used the same set but tweaked the annotation to something more closely resembling the ATIS set used today. The set contains 4978 training samples and 893 test samples. In more recent years with the advent of neural net models, 500 of the training samples are set aside as a validation set.

\cite{tur2010left} work towards formalising the ATIS data set, using the same samples as \cite{he2003datadriven} and \cite{raymond2007generative}. The intents listed by \cite{tur2010left} are not the current ones as they list 17 intents each of which have non-zero frequency in train and test set.

In later releases some joint intents are included to give 21 intents. Also in later releases the SNOR rules are relaxed. For example punctuation is allowed (``st. louis''), utterances are all lowercase, numbers are allowed for times and years but not dates. We note that in the version of the data set used today that the intents are highly imbalanced with 75\% of the samples in a single intent.

\cite{tur2010left} perform an AdaBoost classification on word n-gram features for intent classification and then separately a CRF method to label slots and then perform a classification of error types into 6 types for intent and 5 types for slots. They then suggest research directions based on these errors, being: use of parsers to identify head words or clauses; a priori information (knowledge bases); and, methods to enable long distance pattern identification, as opposed to more local, shorter patterns. They also measure the high mis-annotation rate (2.5\% for intent and 8.4\% for slots).

In 2018 \cite{bechet2018atis} performed the next analysis specifically to question the usefulness of ATIS. They ran a set of different methods from a boosted tree ensemble to a BiLSTM net on ATIS slot tagging with and without named entity tag labelling. They use the same data as \cite{raymond2007generative}, which removes issues with position labels (B,I,O) by collapsing semantic spans as single tokens. For example, `san jose' is a single token not two. While this weakens their approach the results are worth looking at.

They chose their five best models and cluster the predicted slots according to:
\begin{itemize} 
\item agree/correct (AC) - all models get the slot correct and agree on the answer;
\item non-agreement/error (NE) - all models got wrong but there is no agreement on the errors; 
\item agree/error (AE) - all models got the wrong slot but they all made the same error;
\item non-agreement/correct (NC) - models don’t agree on the solution but at least one is correct. 
\end{itemize}

These clusters suggest future directions for research. While AC is `solved', AE and NE are open problems (aspects of the data set not captured by the models), and NC are useful for model comparison between those that got them right and those that did not.

They also highlight issues with the data set - bad annotations, ambiguity ``where slots could be labelled with different labels'', and repetition errors where ``only the first mention of an entity is labelled'', e.g. in ``show flight and prices Kansas city to Chicago on next Wednesday arriving in Chicago by 7pm'' Chicago is only labelled once.

They estimate that about 2.5\% of the utterances are erroneously slot-tagged and conclude that ATIS is at the end of its useful life for analysis.


\cite{niu2019rationally} performed the next deep analysis of the ATIS data set and extensively reviewed the shortcomings of the data set. They have subsequently re-annotated the data set fixing what they deem errors.

Even without this re-annotated version of ATIS results reported in the literature show that the test intent accuracy being achieved is now above 99\% and slot f1 above 98\%. It appears that the models to date have successfully captured the joint distributions of words, slots and intents in the data set. Further models may only make improvements at the edges and while useful may be hidden by what appear non-significant increase in the evaluation measures.  

\subsection{SNIPS}

The SNIPS Natural Language Understanding data set and its creation are fully described in \cite{coucke2018snips}. It contains 15884 utterances (train 13084, development 700, test 700) in 7 balanced intent classes. In training there are 72 slot labels and a vocabulary size of 11241 words. The average sequence length is 9.05. Unlike ATIS, SNIPS covers different domains - weather, restaurants and entertainment. \cite{liu2019cm} show an interesting visualisation that the slot labels used for different domains form largely disjoint sets. These differences have made it a useful counterpoint for experimentation in NLU and models addressing both ATIS and SNIPS successfully show they can handle imbalanced data. However, the reported test results for SNIPS too are excellent - intent accuracy above 99\% and slot f1 around 98\%. 

\subsection{Other data sets}
Microsoft have several non-publicly available sets which have been used by Microsoft researchers. For example FRAMES is multi-turn dialogues around hotel bookings. The Microsoft Cortana personal voice assistant data sets have at least six domains - weather, calendar, communication, reminder, alarm, places. Other software houses with data sets include Facebook (public) and Alexa (private).

Some competitions have applicable data sets,  for example the DSTC 2, 3 and 5 competitions have been used in papers. These often contain multi-turn dialogues. Also the Chinese competition based CCKS data set has been used for research.

The TRAINS data set, a collection of problem-solving dialogues, has been used in four papers. Data sets from diverse but relevant fields have been FrameNet from robotics, CU-Move and AMIE from in vehicle communication, and from question answering CQUD (from Baidu Knows), Yahoo and TREC (only intent annotated).

Non-English data sets have been generated, for example \cite{bellomaria2019almawave} derived an Italian data set starting by translating SNIPS and then using Italian words for tokens like cities or movie names. 

\subsection{Discussion}

It is argued that ATIS and SNIPS have reached near to the end of their useful lives as benchmarks for the joint task. Excellent test results show that the methods developed in this survey can successfully learn the joint distributions of intent and slot labels, and slot labels with each other, in a supervised learning setting. They appear to be set to continue being the benchmarks due to the ability to compare a new approach to previous ones, though this should be tempered by the use of non-standard experimental set up discussed in Section \ref{sec experiment}.

They are useful for study because they are single utterance, have reasonable numbers of intents and slots, are task focused (so have a clear intent), have reasonable utterance lengths. Challenges include mis-annotation, OOV issues and perhaps the level of unnaturalness of the language.

In their defence they provide differences - class imbalance versus imbalance, single versus multi-domain - and a model that scores well on both can claim to have some generalised ability. However as noted in the literature the greater generalisability of such supervised learning models to new domains is in question.

It is probable that more naturally conversational data should be tested. To avoid costly annotation this should be largely unannotated, encouraging research in zero or few shot methods. Such methods can still be tested on ATIS and SNIPS (as in \cite{krone2020learning} (pre-print only). Metrics for measuring the efficacy of such models in the absence of annotation need to be considered.

We further note that all the few and zero shot papers reviewed use annotated datasets for evaluation, hence still need to be transferred to new unseen datasets.

\section{Evaluation metrics} \label{sec metrics}

\subsection{Intent classification}

\subsubsection{Intent accuracy}

For intent classification the widely used metric of accuracy is most commonly used for evaluation. Accuracy is the ratio of the number of correct predictions of intent to the total number of sentences.

Some utterances in the ATIS data set have more than one intent label. Most researchers, since they are not doing multiple label detection, consider the combined label as a new label type, e.g. atis\_airfare\#atis\_flight\_time. \cite{zhang2020graph} note ``some researchers (\cite{liu2016attention}; \cite{li2018self}) count an utterance as a correct classification if any ground truth label is predicted. Others (\cite{goo2018slot}; \cite{e2019novel}) require that all of these intent labels have to be correctly predicted if an utterance is to be counted as a correct classification.''

\subsection{Error rate}

Some papers, \cite{Mohasseb_2018} for example, instead use error rate, the ratio of wrongly classified samples to the total number of samples or 100\% - accuracy, to measure intent classification performance.

\subsubsection{Intent precision, recall and F1}

Less frequently (e.g. \cite{li2019conditional}) precision, recall and F1 are used to evaluate intent prediction. For an intent class $C$, 

\begin{itemize}
  \item TP is the number of True positives, intents which are correctly classified as of class $C$.
  \item FP is the number of False positives, intents which belong to other classes but are incorrectly classified as class $C$.
  \item FN is the number of False negatives, intents of class $C$ which are incorrectly classified as other classes.
\end{itemize}

Two approaches are used; micro-averaged and macro-averaged. In the micro-averaged approach, the TP, FP and FN are summed across all classes:

\begin{equation}
Precision\ =\ \frac{\sum{TP} }{\sum{TP} + \sum{FP}}
\end{equation}

\begin{equation}
Recall\ =\ \frac{\sum{TP} }{\sum{TP} + \sum{FN}}
\end{equation}

In the macro-averaged approach, the precision and recall are computed for each class first, then the average across all $n_C$ classes is reported.

\begin{equation}
Precision\ =\ \frac{1}{n_C} \sum{\frac{TP}{TP + FP}}
\end{equation}

\begin{equation}
Recall\ =\ \frac{1}{n_C} \sum{\frac{TP}{TP + FN}}
\end{equation}

For both approaches, F1 is computed as:

\begin{equation}
F1 = \frac{2 \times precision \times recall}{precision + recall}
\end{equation}



A variation used for multi-label identification are precision and recall at the top-k predictions. Here precision is the ratio of correct labels in the top k predictions divided by k and recall is the ratio of correct labels in top k predictions over the total number of correct labels. This is used by \cite{zhang2016mining}.

\subsubsection{Tests for significance}

Standard tests for significance of difference between two models are used. Welch's t-test is to test the hypothesis that two populations have the same mean. \cite{Firdaus_2018} and \cite{firdaus2019multi} used this with the p-value threshold set to 0.05. Other papers use the student t-test for similar purposes.

McNemar's test is to test paired binary classified data to evaluate how well two tests agree with each other. \cite{jeong2008triangular} used this for a classification of ATIS intents into two domains.

\subsection{Slot labelling evaluation}

\subsubsection{Span slot precision, recall and F1}

A span (sometimes called a chunk) refers to a sequence of words with the same class. For example the labelling B-MISC I-MISC I-MISC is a span of class MISC.

For a class $C$ we can thus define at the span level:

\begin{itemize}
  \item TP is the number of True positives, the number of spans of class $C$ which are wholly correctly predicted.
  \item FP is the number of False positives, the number of spans of a different class which are incorrectly predicted as of class $C$.
  \item FN is the number of False negatives, the number of spans of class $C$ which are incorrectly predicted, partially or wholly, to another class.
\end{itemize}

Micro-averaged and macro-averaged precision and recall and F1 can then be calculated, similarly to the previous intent section.

In most papers slot F1 is reported as the span based micro-averaged F1 over all classes excluding O. The conlleval.py\footnote{https://github.com/sighsmile/conlleval} script is regularly used (\cite{deoras2013deep, liu2019cm, daha2019deep}) to calculate F1 score, precision and recall with micro-averaging. 

\subsubsection{Token-based slot precision, recall and F1}

In this evaluation metric, TP, FP and FN are calculated at the token level. For slot label $L$ (e.g B-MISC):

\begin{itemize}
  \item TP is the number of True positives, is the number of tokens which are correctly predicted as label $L$.
  \item FP is the number of False positives, is the number of tokens which are from another label but incorrectly predicted as label $L$.
  \item FN is the number of False negatives, is the number of tokens of label $L$ which are incorrectly labelled.
\end{itemize}

The formulas for precision, recall and F1 are the same as the span-based. \cite{li2019conditional} use token based slot measures.

\subsubsection{Slot accuracy}

Slot accuracy is the ratio of the number of correctly labelled slots to the total number of slots. This is used in \cite{yu2011sequence} where it is referred as word labelling accuracy (WLA).

\subsection{Semantic accuracy}

A sentence is correctly analysed if both the intent is correctly predicted and all the slots (including O labels) are correctly predicted.  Semantic accuracy is then the number of correctly analysed sentences divided by the number of sentences.

\subsection{Other accuracy measures}

Other classifications are done outside the joint task which are outside the scope of this paper. For example if a domain is predicted a domain accuracy is measured, and if a dialog act is predicted a dialog act accuracy is measured (\cite{celikyilmaz2012joint} used this to evaluate their dialogue model).

\subsection{Qualitative evaluation} \label{sec qualitative}

\cite{jeong2008triangular} used Hinton diagrams to visualise relationships between intent and slot labels arising from weights in their tri-CRF. Attention heat maps are used for similar purposes by \cite{ma2017jointly, qin2020agif}.

\cite{firdaus2019multi} provided t-SNE plots of their intent features to illustrate their effectiveness in prediction. 

\section{Experimental Setup} \label{sec experiment}

The standard experiment trains on annotated utterances, creates features, and learns to predict an intent and slot labels for each utterance. A held out, unseen test set is used for evaluating performance.

The experimental setup varies for different papers. Further, many do not clearly state their setup with respect to data sets and hyper-parameters. In those papers which do specify the setup for data sets, most utilised the train-test split, where usually 80\% of observations were treated as training data and the remainder were for testing. The training data may be split further to make a validation set. Alternatively, papers use 5-fold or 10-fold cross-validation for evaluation. Results are sometimes reported to be averaged over a number of runs.

For parameter tuning, dropout rates ranged from 0.003 to 0.5 and the size of hidden states was normally between 100 and 200, with as low as 64. Some models indicate the use of the Adam optimisation method with a learning rate between 0.0001 and 0.01. \cite{vu2016bidirectional} used 0.02 as their initial learning rate in the first ten epochs and then halved it for the last 15 epochs. Similarly, \cite{ravuri2015recurrent} halved the learning rate once the cross entropy loss decreased less than 0.01 per example on the held out set. Moreover, a few papers mentioned that they set parameters randomly in the beginning, and then apply 5-fold validation when tuning parameters. Word embedding dimensions vary from 64 to 1024.

The same problem of lack of reporting occurred with the number of epochs. In papers which stated the number of epochs, most models were trained for less than 50 epochs, with some of these training models using early stopping. \cite{gupta2019simple} allowed unlimited number of epochs with a stopping criteria, meanwhile \cite{Lin_2019} specified the maximum epoch can be 200 and applied early stopping as well. \cite{qin2019stack} used 300 epochs with no early stopping. Further, some models report the final epoch results while other report the best results. In the joint task there is an effort to standardise the number of epochs for the benchmark data sets to 10 for ATIS and 20 for SNIPS (with early stopping strategy permitted), to allow for comparison between models. This was initialised by \cite{goo2018slot} who also reran the experiments of \cite{hakkani2016multi} and \cite{liu2016attention} under that regime. Table \ref{NLU performance std epoch} contains the results for experiments using this number of epochs.

Considering that many papers did not clearly state their experimental setup, it may bring difficulties in replicating the models and obtaining results similar to those shown in the papers. Therefore, it is recommended that papers include detailed information about setup in the experiments section.

Furthermore, standardisation of the experiment is worth consideration. As discussed in Section \ref{sec performance}, a standard number of epochs on the standard data sets allows for a level of comparison between models. Results should be reported at this level. This should not limit results for different experimental setup being reported. 

\section{Performance summary} \label{sec performance}

To summarise performance in the joint task we list the models and their reported test results for the ATIS and SNIPS data-sets for the three standard evaluation metrics (if available) in Table \ref{NLU performance}. Papers are included in this table if at least one of their results is better than the benchmarks reported in papers from the previous calendar year. Several interesting patterns can be observed based on the results available: (1) The overall improvement on Slot F1 and Semantic accuracy for SNIPS over time (from around 87.3 in 2016 to almost 98.78 in 2019 for Slot F1 and from 73.2 in 2016 to 93.6 in 2020 for Semantic accuracy) is much more significant than ATIS (from 93.96 in 2014 to 98.75 in 2019 for Slot F1 and from 78.9 in 2016 to 91.6 in 2020 for Semantic accuracy), while the Intent accuracy performs in the opposite way (from 78.9 in 2016 to 91.6 in 2020 for ATIS and from 96.7 in 2016 to 99.98 in 2019 for SNIPS. (2) For those models that reported Slot F1 and Intent accuracy on both data-sets, 17 out of 26 perform better in Slot F1 for ATIS and in Intent accuracy for SNIPS. (3) Before 2019, all best performance for Semantic accuracy come from ATIS while from 2019 and a shift to SNIPS starts from 2019 ending up with all best Semantic accuracy in SNIPS. 


However, as mentioned in the previous section, there is a wide variety in the number of epochs for which neural models are allowed to run. In order to make fair comparison, we further extracted those models that use 10 epochs for ATIS and 20 epochs for SNIPS and provide the test results in Table \ref{NLU performance std epoch}. These results are either from papers who follow this etiquette, or from the reproduction by \cite{goo2018slot}, or are replicated by us using the GitHub code supplied by the authors when indicated by $\dagger\dagger$. In the latter case we also confirm the consistent calculation of intent and semantic accuracy as well as span-based slot f1. A similar pattern to before can be observed, that a significant improvement in Slot F1 and Semantic accuracy for SNIPS has been made since 2019 and most of the models perform better in Slot F1 for ATIS and Intent accuracy for SNIPS. These patterns could be related to the various distribution of slot and intent labels and different nature of domains of the two data sets with regard to different architectures of the models. 

In summary, we note that the results are now excellent for the two most commonly used data sets, and any fruitful newer developments that may be lost in results that appear to not significantly increase the results for these datasets. Just as SNIPS grew to become standard, and offered different aspects to ATIS (balanced data, multi-domain), it is probable that a new data set should become part of the SLU reporting canon. It should address the issues of unlabelled data and emerging domains as these problems should be addressed by newer models.

\begin{table*}[t]
\caption{Natural language understanding (NLU) performance on ATIS and SNIPS-NLU data sets (\%). * denotes ATIS 10 epoch, SNIPS 20 epoch, $\dagger$ indicates GitHub available}
\begin{tabular}{|l||c|c|c||c|c|c|}
\hline
\multicolumn{1}{|c||}{\multirow{2}{*}{\textbf{Model}}} & \multicolumn{3}{c||}{\textbf{ATIS}}  & \multicolumn{3}{c|}{\textbf{SNIPS}}  \\ \cline{2-7} 
\multicolumn{1}{|c||}{} & \textbf{Slot f1} & \textbf{Intent acc} & \textbf{Semantic acc} & \textbf{Slot f1} & \textbf{Intent acc} & \textbf{Semantic acc} \\ 
\hline
\cite{jeong2008triangular} Joint 2 & 94.42 & 93.07 & &    & & \\ 
\hline
\cite{xu2013convolutional} CNN TriCRF & 95.42 & 94.09 & &    & & \\ 
\hline
\cite{guo2014joint} RecNN+Viterbi & 93.96 & 95.4 & &    & & \\ 
\hline
\cite{shi2015contextual} RNN Joint + NE & 96.83 & 95.4 & &    & & \\ 
\hline
\cite{hakkani2016multi} in \cite{goo2018slot}* & 94.3 & 92.6 & 80.7  & 87.3   & 96.9 & 73.2  \\ 
\hline
\cite{chen2016syntax} K-SAN Syntax & 95.38   & & 84.32 &   & & \\ 
\hline
\cite{zhang2016joint} W+N & 96.89   & 98.32     &   &   & & \\ 
\hline
\cite{liu2016attention} in \cite{goo2018slot}*   & 94.2   & 91.1 & 78.9   & 87.8   & 96.7 & 74.1 \\
\hline
\cite{goo2018slot} Slot-Gated (Full Atten.)* $\dagger$ & 94.8   & 93.6 & 82.2 & 88.8   & 97.0 & 75.5 \\ 
\hline
\cite{goo2018slot} Slot-Gated (Intent Atten.)* $\dagger$ & 95.2   & 94.1 & 82.6 & 88.3   & 96.8 & 74.6   \\ 
\hline
\cite{wang2018chinese} Attention and aligned  & 97.76 & 97.17 &    &  & &   \\ 
\hline
\cite{firdaus2018deep}  & 98.02  & 98.43 &    &  & &   \\ 
\hline
\cite{li2018self} $\dagger$  & 96.52  & 98.77 &    &  & &   \\ 
\hline
\cite{li2018joint}  & 94.81  & 98.54 &    &  & &   \\ 
\hline
\cite{wang2018bimodel} $\dagger$ & 96.89  & 98.99 &    &  & &   \\ 
\hline
\cite{yu2018novel} ACJIS Model & 96.43  & 98.57     &    &  & &   \\ 
\hline
\cite{siddhant2018unsupervised} ELMo  & 95.62   & 97.42 & 87.35 & 93.9   & 99.29 & 85.43   \\
\hline
\cite{e2019novel} SF First (with CRF) *? $\dagger$  & 95.8    & 97.8   & 86.8   & 91.4    & 97.4 & 80.6   \\
\hline
\cite{zhang2019capsule} Capsule *? $\dagger$  & 95.2  & 95.0  & 83.4  & 91.8   & 97.3 & 80.9     \\ 
\hline
\cite{gupta2019simple} CNN 3L, 5 kern., label recur.  & 96.95    & 98.36 & & 94.22    & 99.1     &  \\
\hline
\cite{gupta2019simple} LSTM 1L, label recur.  & 97.37    & 98.36 & & 93.83    & 98.68    &  \\
\hline
\cite{gupta2019simple} CNN 3L, 5 kern., label recur.*  & 95.27    & 97.37 & & 92.3    & 97.57 &  \\
\hline
\cite{chen2019self}  & 96.54    & 98.91   & & 93.94  & 99.71 &   \\
\hline
\cite{zhang2019using} & 95.1    & 97.2   & & 93.3   & 98.9 &    \\
\hline
\cite{daha2019deep} BiLSTM-CRF  & 95.6   & 96.6 & 86.2 & 94.6     & 97.4     & 87.2   \\
\hline
\cite{liu2019cm} CM-Net with GloVe  & 96.2   & 99.1 & & 97.15   & 99.29   &  \\
\hline
\cite{liu2019cm} CM-Net with BERT  & & & & 97.31     & 99.32  &  \\
\hline
\cite{qin2019stack} Our Model $\dagger$  & 95.9    & 96.9 & 86.5  & 94.2  & 98 & 86.9   \\
\hline
\cite{qin2019stack} Model+BERT  & 96.1     & 97.5 & 88.6  & 97  & 99 & 92.9   \\
\hline
\cite{firdaus2019multi} HCNN+CRF, word+char embed's & 97.32   & 99.09     & & 94.38  & 98.24    &    \\
\hline
\cite{castellucci2019multi} & 95.7 & 97.8    &  88.2  & 96.2 & 99    &  91.6  \\
\hline
\cite{zhang2019joint} & 98.75     & 99.76    & & 98.78 & 99.98 &    \\
\hline
\cite{pentyala2019multi} Base & 95.4    & 96.1    & & 94.8 & 98 &    \\
\hline
\cite{pentyala2019multi} Base+BERT & 95.8    & 96.6    & & 94.5 & 97.6 &    \\
\hline
\cite{chen2019bert} BERT $\dagger$ & 96.1    & 97.5    &  88.2 & 97 & 98.6 &  92.8  \\
\hline
\cite{chen2019bert} BERT+CRF $\dagger$ & 96    & 97.9    &  88.6  & 96.7 & 98.4     &  92.6  \\
\hline
\cite{firdaus2020deep} BLSTM+atten+Multi:DAC+ID+SF & 98.11    & 99.06    & & & &    \\
\hline
\cite{wang2020sasgbc} SASGBC & 96.69  & 98.21    &  91.6 & 96.43 & 98.86 &  92.57  \\
\hline
\cite{wang2020new} CMA-BLSTMS n-128 & 96.89   & 98.88    & & & &  \\
\hline
\cite{tang2020endtoendmask} fully-E@EMG-CRF & 96.4   & 99.0  & 89.6 & 97.2 & 99.7 & 93.6 \\
\hline

\end{tabular}
\label{NLU performance}
\end{table*}

\begin{table*}[t]
\caption{Natural language understanding (NLU) performance on ATIS and SNIPS-NLU data sets (\%) using ATIS 10 epoch, SNIPS 20 epoch. $\dagger\dagger$ reproduced by this paper}
\begin{tabular}{|l||c|c|c||c|c|c|}
\hline
\multicolumn{1}{|c||}{\multirow{2}{*}{\textbf{Model}}} & \multicolumn{3}{c||}{\textbf{ATIS}}  & \multicolumn{3}{c|}{\textbf{SNIPS}}  \\ \cline{2-7} 
\multicolumn{1}{|c||}{} & \textbf{Slot f1} & \textbf{Intent acc} & \textbf{Semantic acc} & \textbf{Slot f1} & \textbf{Intent acc} & \textbf{Semantic acc} \\ 
\hline
\cite{hakkani2016multi} in \cite{goo2018slot}* & 94.3 & 92.6 & 80.7  & 87.3   & 96.9 & 73.2  \\ 
\hline
\cite{liu2016attention} in \cite{goo2018slot}*   & 94.2   & 91.1 & 78.9   & 87.8   & 96.7 & 74.1 \\
\hline
\cite{goo2018slot} Slot-Gated (Full Atten.)* $\dagger$ & 94.8   & 93.6 & 82.2 & 88.8   & 97.0 & 75.5 \\ 
\hline
\cite{goo2018slot} Slot-Gated (Intent Atten.)* $\dagger$ & 95.2   & 94.1 & 82.6 & 88.3   & 96.8 & 74.6   \\ 
\hline
\cite{li2018self} $\dagger\dagger$ & 94.82 & 97.00 & 84.00 & 88.93 & 97.71 & 76.43  \\ 
\hline
\cite{wang2018bimodel} $\dagger\dagger$ & 95.14 & 96.08 & 84.87 & 88.46 & 96.71 & 75.39  \\
\hline
\cite{e2019novel} SF First (with CRF) *? $\dagger$  & 95.8    & 97.8   & 86.8   & 91.4    & 97.4 & 80.6   \\
\hline
\cite{zhang2019capsule} Capsule *? $\dagger$  & 95.2  & 95.0  & 83.4  & 91.8   & 97.3 & 80.9     \\ 
\hline
\cite{gupta2019simple} CNN 3L, 5 kern., label recur.*  & 95.27    & 97.37 & & 92.3    & 97.57 &  \\
\hline
\cite{qin2019stack} Our Model $\dagger\dagger$  & 93.15    & 95.9 & 80.4  & 90.88  & 97.14 & 79.71   \\
\hline
\cite{chen2019bert} BERT $\dagger\dagger$ & 95.54    & 97.54    &  87.35 & \textbf{96.91} & 98.43 &  \textbf{92.43}  \\
\hline
\cite{chen2019bert} BERT+CRF $\dagger\dagger$ & \textbf{96.03}    & \textbf{97.76}  &  \textbf{88.47}  & 96.60 & \textbf{98.57}     &  92.14 \\
\hline

\end{tabular}
\label{NLU performance std epoch}
\end{table*}

\begin{table*}
\caption{Slot F1 Scores of slot filling models on ATIS}
\begin{tabular}{ |l|c| } 

\hline
\textbf{Model} & \textbf{ATIS}  \\

\hline
\cite{li2012use} Log-linear K-DCN & 91.88  \\

\hline
\cite{deoras2013deep} DBN + Sntc & 96.00  \\

\hline
\cite{mesnil2013investigation} Bi-directional Jordan-RNN & 93.98  \\

\hline
\cite{yao2013recurrent} RNN + Lex + NE & 96.60  \\

\hline
\cite{yao2014recurrent} R-CRF Model 2 & 96.65  \\ 

\hline
\cite{yao2014spoken} Deep LSTM & 95.08  \\

\hline
\cite{liu2015recurrent} RNN trained with sampled label linearly decreasing $P_i$  & 97.87  \\ 

\hline
\cite{mesnil2015recurrent} Hybrid & 95.06  \\

\hline
\cite{peng2015recurrent} RNN-EM & 95.25  \\

\hline
\cite{liu2016attention} Attention Encoder-Decoder NN (with aligned inputs) & 95.78  \\

\hline
\cite{kurata2016leveraging} Encoder-labeler Deep LSTM (W) & 95.47  \\

\hline
\cite{vu2016bidirectional} 5xR-biRNN & 95.56  \\

\hline
\cite{vu2016sequential} R-bi-sCNN & 95.61  \\

\hline
\cite{zhu2017encoder} BLSTM-LSTM (Focus) & 95.79  \\

\hline
\cite{gong2019deep} DCMTL & 95.83  \\

\hline
\cite{louvan2018exploring} MTL, different supervision level & 95.94  \\

\hline
\cite{wang2018new} DRL based Augmented Tagging System ($\gamma = 0.9$) & 97.86  \\

\hline
\cite{shen2019progressive} c-ProgModel & 93.91  \\

\hline
\cite{zhang2020deeptime} SC-TDNN-C & 95.73  \\

\hline
\end{tabular}
\label{table:f1-scores}
\end{table*}

\section{Critical Discussion and Conclusions} \label{sec conclusions}
Section \ref{sec introduction} presented the following three questions concerning joint intent detection and slot filling:

\begin{itemize}
    \item Q1: How do these joint models achieve and balance two aspects, intent classification and slot filling?
    \item Q2: Have syntactic clues/features been fully exploited or does semantics override this consideration?
    \item Q3: Can successful models in one supervised domain be made more generalisable to new domains or languages or unseen data?
\end{itemize}

\bibliographystyle{ACM-Reference-Format}
\bibliography{NLU-Survey2020}


\begin{thebibliography}{147}


\ifx \showCODEN    \undefined \def \showCODEN     #1{\unskip}     \fi
\ifx \showDOI      \undefined \def \showDOI       #1{#1}\fi
\ifx \showISBNx    \undefined \def \showISBNx     #1{\unskip}     \fi
\ifx \showISBNxiii \undefined \def \showISBNxiii  #1{\unskip}     \fi
\ifx \showISSN     \undefined \def \showISSN      #1{\unskip}     \fi
\ifx \showLCCN     \undefined \def \showLCCN      #1{\unskip}     \fi
\ifx \shownote     \undefined \def \shownote      #1{#1}          \fi
\ifx \showarticletitle \undefined \def \showarticletitle #1{#1}   \fi
\ifx \showURL      \undefined \def \showURL       {\relax}        \fi
\providecommand\bibfield[2]{#2}
\providecommand\bibinfo[2]{#2}
\providecommand\natexlab[1]{#1}
\providecommand\showeprint[2][]{arXiv:#2}

\bibitem[\protect\citeauthoryear{B{\'e}chet and Raymond}{B{\'e}chet and
  Raymond}{2018}]%
        {bechet2018atis}
\bibfield{author}{\bibinfo{person}{Fr{\'e}d{\'e}ric B{\'e}chet} {and}
  \bibinfo{person}{Christian Raymond}.} \bibinfo{year}{2018}\natexlab{}.
\newblock \showarticletitle{{Is ATIS too shallow to go deeper for benchmarking
  Spoken Language Understanding models?}}. In
  \bibinfo{booktitle}{\emph{{InterSpeech 2018}}}. \bibinfo{publisher}{{ISCA}},
  \bibinfo{address}{Hyderabad, India}, \bibinfo{pages}{1--5}.
\newblock
\urldef\tempurl%
\url{https://hal.inria.fr/hal-01835425}
\showURL{%
\tempurl}


\bibitem[\protect\citeauthoryear{Bellomaria, Castellucci, Favalli, and
  Romagnoli}{Bellomaria et~al\mbox{.}}{2019}]%
        {bellomaria2019almawave}
\bibfield{author}{\bibinfo{person}{Valentina Bellomaria},
  \bibinfo{person}{Giuseppe Castellucci}, \bibinfo{person}{Andrea Favalli},
  {and} \bibinfo{person}{Raniero Romagnoli}.} \bibinfo{year}{2019}\natexlab{}.
\newblock \bibinfo{title}{Almawave-SLU: {A} new dataset for {SLU} in Italian}.
\newblock
\newblock
\showeprint[arxiv]{1907.07526}
\urldef\tempurl%
\url{http://arxiv.org/abs/1907.07526}
\showURL{%
\tempurl}


\bibitem[\protect\citeauthoryear{Bhargava, Celikyilmaz, Hakkani-T{\""u}r, and
  Sarikaya}{Bhargava et~al\mbox{.}}{2013}]%
        {bhargava2013easy}
\bibfield{author}{\bibinfo{person}{Aditya Bhargava}, \bibinfo{person}{Asli
  Celikyilmaz}, \bibinfo{person}{Dilek Hakkani-T{\""u}r}, {and}
  \bibinfo{person}{Ruhi Sarikaya}.} \bibinfo{year}{2013}\natexlab{}.
\newblock \showarticletitle{Easy contextual intent prediction and slot
  detection}. In \bibinfo{booktitle}{\emph{2013 IEEE International Conference
  on Acoustics, Speech and Signal Processing}}. IEEE,
  \bibinfo{publisher}{{IEEE}}, \bibinfo{address}{Vancouver, Canada},
  \bibinfo{pages}{8337--8341}.
\newblock


\bibitem[\protect\citeauthoryear{Bhasin, Natarajan, Mathur, Jeon, and
  Kim}{Bhasin et~al\mbox{.}}{2019}]%
        {bhasin2019unified}
\bibfield{author}{\bibinfo{person}{Anmol Bhasin}, \bibinfo{person}{Bharatram
  Natarajan}, \bibinfo{person}{Gaurav Mathur}, \bibinfo{person}{Joo~Hyuk Jeon},
  {and} \bibinfo{person}{Jun-Seong Kim}.} \bibinfo{year}{2019}\natexlab{}.
\newblock \showarticletitle{Unified Parallel Intent and Slot Prediction with
  Cross Fusion and Slot Masking}. In \bibinfo{booktitle}{\emph{Natural Language
  Processing and Information Systems}},
  \bibfield{editor}{\bibinfo{person}{Elisabeth M{\'e}tais},
  \bibinfo{person}{Farid Meziane}, \bibinfo{person}{Sunil Vadera},
  \bibinfo{person}{Vijayan Sugumaran}, {and} \bibinfo{person}{Mohamad Saraee}}
  (Eds.). \bibinfo{publisher}{Springer International Publishing},
  \bibinfo{address}{Cham}, \bibinfo{pages}{277--285}.
\newblock
\showISBNx{978-3-030-23281-8}


\bibitem[\protect\citeauthoryear{{Bhasin}, {Natarajan}, {Mathur}, and
  {Mangla}}{{Bhasin} et~al\mbox{.}}{2020}]%
        {bhasin2020parallel}
\bibfield{author}{\bibinfo{person}{Anmol {Bhasin}}, \bibinfo{person}{Bharatram
  {Natarajan}}, \bibinfo{person}{Gaurav {Mathur}}, {and}
  \bibinfo{person}{Himanshu {Mangla}}.} \bibinfo{year}{2020}\natexlab{}.
\newblock \showarticletitle{Parallel Intent and Slot Prediction using MLB
  Fusion}. In \bibinfo{booktitle}{\emph{2020 IEEE 14th International Conference
  on Semantic Computing (ICSC)}}. \bibinfo{publisher}{{IEEE}},
  \bibinfo{address}{San Diego, USA}, \bibinfo{pages}{217--220}.
\newblock


\bibitem[\protect\citeauthoryear{Bhathiya and Thayasivam}{Bhathiya and
  Thayasivam}{2020}]%
        {bhatiya2020metalearning}
\bibfield{author}{\bibinfo{person}{Hemanthage~S. Bhathiya} {and}
  \bibinfo{person}{Uthayasanker Thayasivam}.} \bibinfo{year}{2020}\natexlab{}.
\newblock \showarticletitle{Meta Learning for Few-Shot Joint Intent Detection
  and Slot-Filling}. In \bibinfo{booktitle}{\emph{Proceedings of the 2020 5th
  International Conference on Machine Learning Technologies}} (Beijing, China)
  \emph{(\bibinfo{series}{ICMLT 2020})}. \bibinfo{publisher}{Association for
  Computing Machinery}, \bibinfo{address}{New York, NY, USA},
  \bibinfo{pages}{86–92}.
\newblock
\showISBNx{9781450377645}
\urldef\tempurl%
\url{https://doi.org/10.1145/3409073.3409090}
\showDOI{\tempurl}


\bibitem[\protect\citeauthoryear{Bordes, Usunier, Garcia-Duran, Weston, and
  Yakhnenko}{Bordes et~al\mbox{.}}{2013}]%
        {bordes2013transe}
\bibfield{author}{\bibinfo{person}{Antoine Bordes}, \bibinfo{person}{Nicolas
  Usunier}, \bibinfo{person}{Alberto Garcia-Duran}, \bibinfo{person}{Jason
  Weston}, {and} \bibinfo{person}{Oksana Yakhnenko}.}
  \bibinfo{year}{2013}\natexlab{}.
\newblock \showarticletitle{Translating Embeddings for Modeling
  Multi-relational Data}.
\newblock In \bibinfo{booktitle}{\emph{Advances in Neural Information
  Processing Systems 26}}, \bibfield{editor}{\bibinfo{person}{C.~J.~C. Burges},
  \bibinfo{person}{L.~Bottou}, \bibinfo{person}{M.~Welling},
  \bibinfo{person}{Z.~Ghahramani}, {and} \bibinfo{person}{K.~Q. Weinberger}}
  (Eds.). \bibinfo{publisher}{Curran Associates, Inc.}, \bibinfo{address}{Lake
  Tahoe, USA}, \bibinfo{pages}{2787--2795}.
\newblock
\urldef\tempurl%
\url{http://papers.nips.cc/paper/5071-translating-embeddings-for-modeling-multi-relational-data.pdf}
\showURL{%
\tempurl}


\bibitem[\protect\citeauthoryear{Castellucci, Bellomaria, Favalli, and
  Romagnoli}{Castellucci et~al\mbox{.}}{2019}]%
        {castellucci2019multi}
\bibfield{author}{\bibinfo{person}{Giuseppe Castellucci},
  \bibinfo{person}{Valentina Bellomaria}, \bibinfo{person}{Andrea Favalli},
  {and} \bibinfo{person}{Raniero Romagnoli}.} \bibinfo{year}{2019}\natexlab{}.
\newblock \bibinfo{title}{Multi-lingual Intent Detection and Slot Filling in a
  Joint BERT-based Model}.
\newblock
\newblock
\showeprint[arxiv]{1907.02884}
\urldef\tempurl%
\url{http://arxiv.org/abs/1907.02884}
\showURL{%
\tempurl}


\bibitem[\protect\citeauthoryear{Celikyilmaz and Hakkani-Tur}{Celikyilmaz and
  Hakkani-Tur}{2012}]%
        {celikyilmaz2012joint}
\bibfield{author}{\bibinfo{person}{Asli Celikyilmaz} {and}
  \bibinfo{person}{Dilek Hakkani-Tur}.} \bibinfo{year}{2012}\natexlab{}.
\newblock \showarticletitle{A Joint Model for Discovery of Aspects in
  Utterances}. In \bibinfo{booktitle}{\emph{Proceedings of the 50th Annual
  Meeting of the Association for Computational Linguistics (Volume 1: Long
  Papers)}}. \bibinfo{publisher}{Association for Computational Linguistics},
  \bibinfo{address}{Jeju Island, Korea}, \bibinfo{pages}{330--338}.
\newblock
\urldef\tempurl%
\url{https://www.aclweb.org/anthology/P12-1035}
\showURL{%
\tempurl}


\bibitem[\protect\citeauthoryear{Chen, Zhang, and Mark}{Chen
  et~al\mbox{.}}{2012}]%
        {Chen_2012}
\bibfield{author}{\bibinfo{person}{Long Chen}, \bibinfo{person}{Dell Zhang},
  {and} \bibinfo{person}{Levene Mark}.} \bibinfo{year}{2012}\natexlab{}.
\newblock \showarticletitle{Understanding user intent in community question
  answering}. In \bibinfo{booktitle}{\emph{Proceedings of the 21st
  International Conference Companion on World Wide Web - {WWW}
  {\textquotesingle}12 Companion}}. \bibinfo{publisher}{{ACM} Press},
  \bibinfo{address}{Lyon, France}, \bibinfo{pages}{823–828}.
\newblock
\urldef\tempurl%
\url{https://doi.org/10.1145/2187980.2188206}
\showDOI{\tempurl}


\bibitem[\protect\citeauthoryear{Chen, Zeng, and Lou}{Chen
  et~al\mbox{.}}{2019a}]%
        {chen2019self}
\bibfield{author}{\bibinfo{person}{Mengyang Chen}, \bibinfo{person}{Jin Zeng},
  {and} \bibinfo{person}{Jie Lou}.} \bibinfo{year}{2019}\natexlab{a}.
\newblock \bibinfo{title}{A Self-Attention Joint Model for Spoken Language
  Understanding in Situational Dialog Applications}.
\newblock
\newblock
\showeprint[arxiv]{1905.11393}
\urldef\tempurl%
\url{http://arxiv.org/abs/1905.11393}
\showURL{%
\tempurl}


\bibitem[\protect\citeauthoryear{Chen, Zhuo, and Wang}{Chen
  et~al\mbox{.}}{2019b}]%
        {chen2019bert}
\bibfield{author}{\bibinfo{person}{Qian Chen}, \bibinfo{person}{Zhu Zhuo},
  {and} \bibinfo{person}{Wen Wang}.} \bibinfo{year}{2019}\natexlab{b}.
\newblock \bibinfo{title}{{BERT} for Joint Intent Classification and Slot
  Filling}.
\newblock
\newblock
\showeprint[arxiv]{1902.10909}
\urldef\tempurl%
\url{http://arxiv.org/abs/1902.10909}
\showURL{%
\tempurl}


\bibitem[\protect\citeauthoryear{Chen and Yu}{Chen and Yu}{2019}]%
        {chen2019wais}
\bibfield{author}{\bibinfo{person}{Sixuan Chen} {and} \bibinfo{person}{Shuai
  Yu}.} \bibinfo{year}{2019}\natexlab{}.
\newblock \showarticletitle{WAIS: Word Attention for Joint Intent Detection and
  Slot Filling}. In \bibinfo{booktitle}{\emph{Proceedings of the AAAI
  Conference on Artificial Intelligence}}, Vol.~\bibinfo{volume}{33}.
  \bibinfo{publisher}{AAAI Press}, \bibinfo{address}{Honolulu, USA},
  \bibinfo{pages}{9927--9928}.
\newblock
\urldef\tempurl%
\url{https://doi.org/10.1609/aaai.v33i01.33019927}
\showDOI{\tempurl}


\bibitem[\protect\citeauthoryear{{Chen}, {Hakanni-Tür}, {Tur}, {Celikyilmaz},
  {Guo}, and {Deng}}{{Chen} et~al\mbox{.}}{2016}]%
        {chen2016syntax}
\bibfield{author}{\bibinfo{person}{Yun-Nung {Chen}}, \bibinfo{person}{Dilek
  {Hakanni-Tür}}, \bibinfo{person}{Gokhan {Tur}}, \bibinfo{person}{Asli
  {Celikyilmaz}}, \bibinfo{person}{Jianfeng {Guo}}, {and} \bibinfo{person}{Li
  {Deng}}.} \bibinfo{year}{2016}\natexlab{}.
\newblock \showarticletitle{Syntax or semantics? Knowledge-guided joint
  semantic frame parsing}. In \bibinfo{booktitle}{\emph{2016 IEEE Spoken
  Language Technology Workshop (SLT)}}. \bibinfo{publisher}{{IEEE}},
  \bibinfo{address}{San Diego, USA}, \bibinfo{pages}{348--355}.
\newblock


\bibitem[\protect\citeauthoryear{Cohan, Ammar, van Zuylen, and Cady}{Cohan
  et~al\mbox{.}}{2019}]%
        {Cohan_2019}
\bibfield{author}{\bibinfo{person}{Arman Cohan}, \bibinfo{person}{Waleed
  Ammar}, \bibinfo{person}{Madeleine van Zuylen}, {and} \bibinfo{person}{Field
  Cady}.} \bibinfo{year}{2019}\natexlab{}.
\newblock \showarticletitle{Structural Scaffolds for Citation Intent
  Classification in Scientific Publications}. In
  \bibinfo{booktitle}{\emph{Proceedings of the 2019 Conference of the North
  {A}merican Chapter of the Association for Computational Linguistics: Human
  Language Technologies, Volume 1 (Long and Short Papers)}}.
  \bibinfo{publisher}{Association for Computational Linguistics},
  \bibinfo{address}{Minneapolis, Minnesota}, \bibinfo{pages}{3586--3596}.
\newblock
\urldef\tempurl%
\url{https://doi.org/10.18653/v1/N19-1361}
\showDOI{\tempurl}


\bibitem[\protect\citeauthoryear{Costello, Lin, Mruthyunjaya, Bolla, and
  Jankowski}{Costello et~al\mbox{.}}{2018}]%
        {costello2018multi}
\bibfield{author}{\bibinfo{person}{Charles Costello}, \bibinfo{person}{Ruixi
  Lin}, \bibinfo{person}{Vishwas Mruthyunjaya}, \bibinfo{person}{Bettina
  Bolla}, {and} \bibinfo{person}{Charles Jankowski}.}
  \bibinfo{year}{2018}\natexlab{}.
\newblock \bibinfo{title}{Multi-Layer Ensembling Techniques for Multilingual
  Intent Classification}.
\newblock
\newblock
\showeprint[arxiv]{1806.07914}~[cs.CL]


\bibitem[\protect\citeauthoryear{Coucke, Saade, Ball, Bluche, Caulier, Leroy,
  Doumouro, Gisselbrecht, Caltagirone, Lavril, Primet, and Dureau}{Coucke
  et~al\mbox{.}}{2018}]%
        {coucke2018snips}
\bibfield{author}{\bibinfo{person}{Alice Coucke}, \bibinfo{person}{Alaa Saade},
  \bibinfo{person}{Adrien Ball}, \bibinfo{person}{Th{\'{e}}odore Bluche},
  \bibinfo{person}{Alexandre Caulier}, \bibinfo{person}{David Leroy},
  \bibinfo{person}{Cl{\'{e}}ment Doumouro}, \bibinfo{person}{Thibault
  Gisselbrecht}, \bibinfo{person}{Francesco Caltagirone},
  \bibinfo{person}{Thibaut Lavril}, \bibinfo{person}{Ma{\"{e}}l Primet}, {and}
  \bibinfo{person}{Joseph Dureau}.} \bibinfo{year}{2018}\natexlab{}.
\newblock \bibinfo{title}{Snips Voice Platform: an embedded Spoken Language
  Understanding system for private-by-design voice interfaces}.
\newblock
\newblock
\showeprint[arxiv]{1805.10190}
\urldef\tempurl%
\url{http://arxiv.org/abs/1805.10190}
\showURL{%
\tempurl}


\bibitem[\protect\citeauthoryear{Dadas, Protasiewicz, and Pedrycz}{Dadas
  et~al\mbox{.}}{2019}]%
        {dadas2019deep}
\bibfield{author}{\bibinfo{person}{Slawomir Dadas}, \bibinfo{person}{Jaroslaw
  Protasiewicz}, {and} \bibinfo{person}{Witold Pedrycz}.}
  \bibinfo{year}{2019}\natexlab{}.
\newblock \showarticletitle{A Deep Learning Model with Data Enrichment for
  Intent Detection and Slot Filling}. In \bibinfo{booktitle}{\emph{2019 {IEEE}
  International Conference on Systems, Man and Cybernetics, October 6-9,
  2019}}. \bibinfo{publisher}{{IEEE}}, \bibinfo{address}{Bari, Italy},
  \bibinfo{pages}{3012--3018}.
\newblock
\urldef\tempurl%
\url{https://doi.org/10.1109/SMC.2019.8914542}
\showDOI{\tempurl}


\bibitem[\protect\citeauthoryear{{Daha} and {Hewavitharana}}{{Daha} and
  {Hewavitharana}}{2019}]%
        {daha2019deep}
\bibfield{author}{\bibinfo{person}{Fatima {Daha}} {and}
  \bibinfo{person}{Saniika {Hewavitharana}}.} \bibinfo{year}{2019}\natexlab{}.
\newblock \showarticletitle{Deep Neural Architecture with Character Embedding
  for Semantic Frame Detection}. In \bibinfo{booktitle}{\emph{2019 IEEE 13th
  International Conference on Semantic Computing (ICSC)}}.
  \bibinfo{publisher}{{IEEE}}, \bibinfo{address}{Newport Beach, USA},
  \bibinfo{pages}{302--307}.
\newblock


\bibitem[\protect\citeauthoryear{Dahl, Bates, Brown, Fisher, Hunicke-Smith,
  Pallett, Pao, Rudnicky, and Shriberg}{Dahl et~al\mbox{.}}{1994}]%
        {dahl1994expanding}
\bibfield{author}{\bibinfo{person}{Deborah~A. Dahl}, \bibinfo{person}{Madeleine
  Bates}, \bibinfo{person}{Michael Brown}, \bibinfo{person}{William Fisher},
  \bibinfo{person}{Kate Hunicke-Smith}, \bibinfo{person}{David Pallett},
  \bibinfo{person}{Christine Pao}, \bibinfo{person}{Alexander Rudnicky}, {and}
  \bibinfo{person}{Elizabeth Shriberg}.} \bibinfo{year}{1994}\natexlab{}.
\newblock \showarticletitle{Expanding the Scope of the {ATIS} Task: The
  {ATIS}-3 Corpus}. In \bibinfo{booktitle}{\emph{{H}uman {L}anguage
  {T}echnology: Proceedings of a Workshop held at {P}lainsboro, {N}ew {J}ersey,
  {M}arch 8-11, 1994}}. \bibinfo{publisher}{Association for Computational
  Linguistics}, \bibinfo{address}{Plainsboro, USA}, \bibinfo{pages}{43--48}.
\newblock
\urldef\tempurl%
\url{https://www.aclweb.org/anthology/H94-1010}
\showURL{%
\tempurl}


\bibitem[\protect\citeauthoryear{{Dai}, {Zhang}, {Ou}, {Wang}, and
  {Feng}}{{Dai} et~al\mbox{.}}{2018}]%
        {dai2018elastic}
\bibfield{author}{\bibinfo{person}{Yinpei {Dai}}, \bibinfo{person}{Yichi
  {Zhang}}, \bibinfo{person}{Zhijian {Ou}}, \bibinfo{person}{Yanmeng {Wang}},
  {and} \bibinfo{person}{Junlan {Feng}}.} \bibinfo{year}{2018}\natexlab{}.
\newblock \bibinfo{title}{Elastic CRFs for Open-ontology Slot Filling}.
\newblock
\newblock
\showeprint[arxiv]{1811.01331}


\bibitem[\protect\citeauthoryear{Deng, Tur, He, and Hakkani-Tür}{Deng
  et~al\mbox{.}}{2012}]%
        {li2012use}
\bibfield{author}{\bibinfo{person}{Li Deng}, \bibinfo{person}{Gokhan Tur},
  \bibinfo{person}{Xiaodong He}, {and} \bibinfo{person}{Dilek Hakkani-Tür}.}
  \bibinfo{year}{2012}\natexlab{}.
\newblock \showarticletitle{Use of kernel deep convex networks and end-to-end
  learning for spoken language understanding}. In
  \bibinfo{booktitle}{\emph{2012 IEEE Spoken Language Technology Workshop
  (SLT)}}. \bibinfo{publisher}{{IEEE}}, \bibinfo{address}{Miami, USA},
  \bibinfo{pages}{210--215}.
\newblock


\bibitem[\protect\citeauthoryear{Deoras and Sarikaya}{Deoras and
  Sarikaya}{2013}]%
        {deoras2013deep}
\bibfield{author}{\bibinfo{person}{Anoop Deoras} {and} \bibinfo{person}{Ruhi
  Sarikaya}.} \bibinfo{year}{2013}\natexlab{}.
\newblock \showarticletitle{Deep Belief Network based Semantic Taggers for
  Spoken Language Understanding}. In \bibinfo{booktitle}{\emph{Interspeech}}.
  \bibinfo{publisher}{ISCA}, \bibinfo{address}{Lyon, France},
  \bibinfo{pages}{2713--2717}.
\newblock


\bibitem[\protect\citeauthoryear{Deriu, Rodrigo, Otegi, Echegoyen, Rosset,
  Agirre, and Cieliebak}{Deriu et~al\mbox{.}}{2020}]%
        {deriu2020surveyevaluation}
\bibfield{author}{\bibinfo{person}{Jan Deriu}, \bibinfo{person}{Alvaro
  Rodrigo}, \bibinfo{person}{Arantxa Otegi}, \bibinfo{person}{Guillermo
  Echegoyen}, \bibinfo{person}{Sophie Rosset}, \bibinfo{person}{Eneko Agirre},
  {and} \bibinfo{person}{Mark Cieliebak}.} \bibinfo{year}{2020}\natexlab{}.
\newblock \showarticletitle{Survey on evaluation methods for dialogue systems}.
\newblock \bibinfo{journal}{\emph{Artificial Intelligence Review}}
  \bibinfo{volume}{53} (\bibinfo{year}{2020}).
\newblock
\showISBNx{1573-7462}
\urldef\tempurl%
\url{https://doi.org/10.1007/s10462-020-09866-x}
\showDOI{\tempurl}


\bibitem[\protect\citeauthoryear{E, Niu, Chen, and Song}{E
  et~al\mbox{.}}{2019}]%
        {e2019novel}
\bibfield{author}{\bibinfo{person}{Haihong E}, \bibinfo{person}{Peiqing Niu},
  \bibinfo{person}{Zhongfu Chen}, {and} \bibinfo{person}{Meina Song}.}
  \bibinfo{year}{2019}\natexlab{}.
\newblock \showarticletitle{A Novel Bi-directional Interrelated Model for Joint
  Intent Detection and Slot Filling}. In \bibinfo{booktitle}{\emph{Proceedings
  of the 57th Annual Meeting of the Association for Computational
  Linguistics}}. \bibinfo{publisher}{Association for Computational
  Linguistics}, \bibinfo{address}{Florence, Italy},
  \bibinfo{pages}{5467--5471}.
\newblock
\urldef\tempurl%
\url{https://doi.org/10.18653/v1/P19-1544}
\showDOI{\tempurl}


\bibitem[\protect\citeauthoryear{Firdaus, Bhatnagar, Ekbal, and
  Bhattacharyya}{Firdaus et~al\mbox{.}}{2018a}]%
        {firdaus2018deep}
\bibfield{author}{\bibinfo{person}{Mauajama Firdaus}, \bibinfo{person}{Shobhit
  Bhatnagar}, \bibinfo{person}{Asif Ekbal}, {and} \bibinfo{person}{Pushpak
  Bhattacharyya}.} \bibinfo{year}{2018}\natexlab{a}.
\newblock \showarticletitle{A Deep Learning Based Multi-task Ensemble Model for
  Intent Detection and Slot Filling in Spoken Language Understanding}. In
  \bibinfo{booktitle}{\emph{Neural Information Processing}},
  \bibfield{editor}{\bibinfo{person}{Long Cheng}, \bibinfo{person}{Andrew
  Chi~Sing Leung}, {and} \bibinfo{person}{Seiichi Ozawa}} (Eds.).
  \bibinfo{publisher}{Springer International Publishing},
  \bibinfo{address}{Cham}, \bibinfo{pages}{647--658}.
\newblock
\showISBNx{978-3-030-04212-7}


\bibitem[\protect\citeauthoryear{Firdaus, Bhatnagar, Ekbal, and
  Bhattacharyya}{Firdaus et~al\mbox{.}}{2018b}]%
        {Firdaus_2018}
\bibfield{author}{\bibinfo{person}{Mauajama Firdaus}, \bibinfo{person}{Shobhit
  Bhatnagar}, \bibinfo{person}{Asif Ekbal}, {and} \bibinfo{person}{Pushpak
  Bhattacharyya}.} \bibinfo{year}{2018}\natexlab{b}.
\newblock \showarticletitle{Intent Detection for Spoken Language Understanding
  Using a Deep Ensemble Model}.
\newblock In \bibinfo{booktitle}{\emph{Lecture Notes in Computer Science}}.
  \bibinfo{publisher}{Springer International Publishing},
  \bibinfo{address}{Cham}, \bibinfo{pages}{629--642}.
\newblock
\urldef\tempurl%
\url{https://doi.org/10.1007/978-3-319-97304-3_48}
\showDOI{\tempurl}


\bibitem[\protect\citeauthoryear{Firdaus, Golchha, Ekbal, and
  Bhattacharyya}{Firdaus et~al\mbox{.}}{2020}]%
        {firdaus2020deep}
\bibfield{author}{\bibinfo{person}{Mauajama Firdaus}, \bibinfo{person}{Hitesh
  Golchha}, \bibinfo{person}{Asif Ekbal}, {and} \bibinfo{person}{Pushpak
  Bhattacharyya}.} \bibinfo{year}{2020}\natexlab{}.
\newblock \showarticletitle{A Deep Multi-task Model for Dialogue Act
  Classification, Intent Detection and Slot Filling}.
\newblock \bibinfo{journal}{\emph{Cognitive Computation}}  \bibinfo{volume}{12}
  (\bibinfo{year}{2020}).
\newblock
\showISBNx{1866-9964}
\urldef\tempurl%
\url{https://doi.org/10.1007/s12559-020-09718-4}
\showDOI{\tempurl}


\bibitem[\protect\citeauthoryear{Firdaus, Kumar, Ekbal, and
  Bhattacharyya}{Firdaus et~al\mbox{.}}{2019}]%
        {firdaus2019multi}
\bibfield{author}{\bibinfo{person}{Mauajama Firdaus}, \bibinfo{person}{Ankit
  Kumar}, \bibinfo{person}{Asif Ekbal}, {and} \bibinfo{person}{Pushpak
  Bhattacharyya}.} \bibinfo{year}{2019}\natexlab{}.
\newblock \showarticletitle{A Multi-Task Hierarchical Approach for Intent
  Detection and Slot Filling}.
\newblock \bibinfo{journal}{\emph{Knowledge-Based Systems}}
  \bibinfo{volume}{183} (\bibinfo{year}{2019}), \bibinfo{pages}{104846}.
\newblock
\showISSN{0950-7051}
\urldef\tempurl%
\url{https://doi.org/10.1016/j.knosys.2019.07.017}
\showDOI{\tempurl}


\bibitem[\protect\citeauthoryear{Gangadharaiah and Narayanaswamy}{Gangadharaiah
  and Narayanaswamy}{2019}]%
        {gangadharaiah2019joint}
\bibfield{author}{\bibinfo{person}{Rashmi Gangadharaiah} {and}
  \bibinfo{person}{Balakrishnan Narayanaswamy}.}
  \bibinfo{year}{2019}\natexlab{}.
\newblock \showarticletitle{Joint Multiple Intent Detection and Slot Labeling
  for Goal-Oriented Dialog}. In \bibinfo{booktitle}{\emph{Proceedings of the
  2019 Conference of the North {A}merican Chapter of the Association for
  Computational Linguistics: Human Language Technologies, Volume 1 (Long and
  Short Papers)}}. \bibinfo{publisher}{Association for Computational
  Linguistics}, \bibinfo{address}{Minneapolis, Minnesota},
  \bibinfo{pages}{564--569}.
\newblock
\urldef\tempurl%
\url{https://doi.org/10.18653/v1/N19-1055}
\showDOI{\tempurl}


\bibitem[\protect\citeauthoryear{Gong, Luo, Zhu, Ou, Li, Zhu, Zhu, Duan, and
  Chen}{Gong et~al\mbox{.}}{2019}]%
        {gong2019deep}
\bibfield{author}{\bibinfo{person}{Yu Gong}, \bibinfo{person}{Xusheng Luo},
  \bibinfo{person}{Yu Zhu}, \bibinfo{person}{Wenwu Ou}, \bibinfo{person}{Zhao
  Li}, \bibinfo{person}{Muhua Zhu}, \bibinfo{person}{Kenny~Q. Zhu},
  \bibinfo{person}{Lu Duan}, {and} \bibinfo{person}{Xi Chen}.}
  \bibinfo{year}{2019}\natexlab{}.
\newblock \showarticletitle{Deep Cascade Multi-task Learning for Slot Filling
  in Online Shopping Assistant}. In \bibinfo{booktitle}{\emph{Proceedings of
  the AAAI Conference on Artificial Intelligence}}, Vol.~\bibinfo{volume}{33}.
  \bibinfo{publisher}{AAAI Press}, \bibinfo{address}{Honolulu, USA},
  \bibinfo{pages}{6465--6472}.
\newblock
\urldef\tempurl%
\url{https://doi.org/10.1609/aaai.v33i01.33016465}
\showDOI{\tempurl}


\bibitem[\protect\citeauthoryear{Gonz{\'a}lez-Caro and
  Baeza-Yates}{Gonz{\'a}lez-Caro and Baeza-Yates}{2011}]%
        {gonzalez2011multifaceted}
\bibfield{author}{\bibinfo{person}{Cristina Gonz{\'a}lez-Caro} {and}
  \bibinfo{person}{Ricardo Baeza-Yates}.} \bibinfo{year}{2011}\natexlab{}.
\newblock \showarticletitle{A Multi-faceted Approach to Query Intent
  Classification}. In \bibinfo{booktitle}{\emph{String Processing and
  Information Retrieval}}, \bibfield{editor}{\bibinfo{person}{Roberto Grossi},
  \bibinfo{person}{Fabrizio Sebastiani}, {and} \bibinfo{person}{Fabrizio
  Silvestri}} (Eds.). \bibinfo{publisher}{Springer Berlin Heidelberg},
  \bibinfo{address}{Berlin, Heidelberg}, \bibinfo{pages}{368--379}.
\newblock
\showISBNx{978-3-642-24583-1}


\bibitem[\protect\citeauthoryear{Goo, Gao, Hsu, Huo, Chen, Hsu, and Chen}{Goo
  et~al\mbox{.}}{2018}]%
        {goo2018slot}
\bibfield{author}{\bibinfo{person}{Chih-Wen Goo}, \bibinfo{person}{Guang Gao},
  \bibinfo{person}{Yun-Kai Hsu}, \bibinfo{person}{Chih-Li Huo},
  \bibinfo{person}{Tsung-Chieh Chen}, \bibinfo{person}{Keng-Wei Hsu}, {and}
  \bibinfo{person}{Yun-Nung Chen}.} \bibinfo{year}{2018}\natexlab{}.
\newblock \showarticletitle{Slot-gated modeling for joint slot filling and
  intent prediction}. In \bibinfo{booktitle}{\emph{Proceedings of the 2018
  Conference of the North American Chapter of the Association for Computational
  Linguistics: Human Language Technologies, Volume 2 (Short Papers)}}.
  \bibinfo{publisher}{Association for Computational Linguistics},
  \bibinfo{address}{New Orleans, USA}, \bibinfo{pages}{753--757}.
\newblock


\bibitem[\protect\citeauthoryear{Guo, Tur, Yih, and Zweig}{Guo
  et~al\mbox{.}}{2014}]%
        {guo2014joint}
\bibfield{author}{\bibinfo{person}{Daniel Guo}, \bibinfo{person}{Gokhan Tur},
  \bibinfo{person}{Wen-tau Yih}, {and} \bibinfo{person}{Geoffrey Zweig}.}
  \bibinfo{year}{2014}\natexlab{}.
\newblock \showarticletitle{Joint semantic utterance classification and slot
  filling with recursive neural networks}. In \bibinfo{booktitle}{\emph{2014
  IEEE Spoken Language Technology Workshop (SLT)}}. IEEE,
  \bibinfo{publisher}{IEEE}, \bibinfo{address}{South Lake Tahoe, USA},
  \bibinfo{pages}{554--559}.
\newblock


\bibitem[\protect\citeauthoryear{Gupta, Hewitt, and Kirchhoff}{Gupta
  et~al\mbox{.}}{2019a}]%
        {gupta2019simple}
\bibfield{author}{\bibinfo{person}{Arshit Gupta}, \bibinfo{person}{John
  Hewitt}, {and} \bibinfo{person}{Katrin Kirchhoff}.}
  \bibinfo{year}{2019}\natexlab{a}.
\newblock \showarticletitle{Simple, Fast, Accurate Intent Classification and
  Slot Labeling for Goal-Oriented Dialogue Systems}. In
  \bibinfo{booktitle}{\emph{Proceedings of the 20th Annual SIGdial Meeting on
  Discourse and Dialogue}}. \bibinfo{publisher}{Association for Computational
  Linguistics}, \bibinfo{address}{Stockholm, Sweden}, \bibinfo{pages}{46--55}.
\newblock
\urldef\tempurl%
\url{https://doi.org/10.18653/v1/W19-5906}
\showDOI{\tempurl}


\bibitem[\protect\citeauthoryear{Gupta, Zhang, Lalwani, and Diab}{Gupta
  et~al\mbox{.}}{2019b}]%
        {gupta2019casa}
\bibfield{author}{\bibinfo{person}{Arshit Gupta}, \bibinfo{person}{Peng Zhang},
  \bibinfo{person}{Garima Lalwani}, {and} \bibinfo{person}{Mona Diab}.}
  \bibinfo{year}{2019}\natexlab{b}.
\newblock \showarticletitle{{CASA}-{NLU}: Context-Aware Self-Attentive Natural
  Language Understanding for Task-Oriented Chatbots}. In
  \bibinfo{booktitle}{\emph{Proceedings of the 2019 Conference on Empirical
  Methods in Natural Language Processing and the 9th International Joint
  Conference on Natural Language Processing (EMNLP-IJCNLP)}}.
  \bibinfo{publisher}{Association for Computational Linguistics},
  \bibinfo{address}{Hong Kong, China}, \bibinfo{pages}{1285--1290}.
\newblock
\urldef\tempurl%
\url{https://doi.org/10.18653/v1/D19-1127}
\showDOI{\tempurl}


\bibitem[\protect\citeauthoryear{Hakkani-T{\"u}r, T{\"u}r, Celikyilmaz, Chen,
  Gao, Deng, and Wang}{Hakkani-T{\"u}r et~al\mbox{.}}{2016}]%
        {hakkani2016multi}
\bibfield{author}{\bibinfo{person}{Dilek Hakkani-T{\"u}r},
  \bibinfo{person}{G{\"o}khan T{\"u}r}, \bibinfo{person}{Asli Celikyilmaz},
  \bibinfo{person}{Yun-Nung Chen}, \bibinfo{person}{Jianfeng Gao},
  \bibinfo{person}{Li Deng}, {and} \bibinfo{person}{Ye-Yi Wang}.}
  \bibinfo{year}{2016}\natexlab{}.
\newblock \showarticletitle{Multi-domain joint semantic frame parsing using
  bi-directional RNN-LSTM}. In \bibinfo{booktitle}{\emph{Interspeech}}.
  \bibinfo{publisher}{{ISCA}}, \bibinfo{address}{San Francisco, USA},
  \bibinfo{pages}{715--719}.
\newblock


\bibitem[\protect\citeauthoryear{Hasanuzzaman, Saha, Dias, and
  Ferrari}{Hasanuzzaman et~al\mbox{.}}{2015}]%
        {Hasanuzzaman_2015}
\bibfield{author}{\bibinfo{person}{Mohammed Hasanuzzaman},
  \bibinfo{person}{Sriparna Saha}, \bibinfo{person}{Gaël Dias}, {and}
  \bibinfo{person}{St{\'{e}}phane Ferrari}.} \bibinfo{year}{2015}\natexlab{}.
\newblock \showarticletitle{Understanding Temporal Query Intent}. In
  \bibinfo{booktitle}{\emph{Proceedings of the 38th International {ACM} {SIGIR}
  Conference on Research and Development in Information Retrieval - {SIGIR}
  {\textquotesingle}15}}. \bibinfo{publisher}{{ACM} Press},
  \bibinfo{address}{Santiago, Chile}, \bibinfo{pages}{823--826}.
\newblock
\urldef\tempurl%
\url{https://doi.org/10.1145/2766462.2767792}
\showDOI{\tempurl}


\bibitem[\protect\citeauthoryear{Hashemi, Asiaee, and Kraft}{Hashemi
  et~al\mbox{.}}{2016}]%
        {hashemi2016query}
\bibfield{author}{\bibinfo{person}{Homa~B Hashemi}, \bibinfo{person}{Amir
  Asiaee}, {and} \bibinfo{person}{Reiner Kraft}.}
  \bibinfo{year}{2016}\natexlab{}.
\newblock \showarticletitle{Query intent detection using convolutional neural
  networks}. In \bibinfo{booktitle}{\emph{International Conference on Web
  Search and Data Mining, Workshop on Query Understanding}}.
  \bibinfo{publisher}{Association for Computational Linguistics},
  \bibinfo{address}{San Francisco, USA}.
\newblock


\bibitem[\protect\citeauthoryear{Hemphill, Godfrey, and Doddington}{Hemphill
  et~al\mbox{.}}{1990}]%
        {hemphill1990atis}
\bibfield{author}{\bibinfo{person}{Charles~T. Hemphill},
  \bibinfo{person}{John~J. Godfrey}, {and} \bibinfo{person}{George~R.
  Doddington}.} \bibinfo{year}{1990}\natexlab{}.
\newblock \showarticletitle{The ATIS Spoken Language Systems Pilot Corpus}. In
  \bibinfo{booktitle}{\emph{Proceedings of the Workshop on Speech and Natural
  Language}} \emph{(\bibinfo{series}{HLT ’90})}.
  \bibinfo{publisher}{Association for Computational Linguistics},
  \bibinfo{address}{Hidden Valley, Pennsylvania, USA},
  \bibinfo{pages}{96–101}.
\newblock
\urldef\tempurl%
\url{https://doi.org/10.3115/116580.116613}
\showDOI{\tempurl}


\bibitem[\protect\citeauthoryear{Hirschman}{Hirschman}{1992}]%
        {hirschman1992atis2}
\bibfield{author}{\bibinfo{person}{Lynette Hirschman}.}
  \bibinfo{year}{1992}\natexlab{}.
\newblock \showarticletitle{Multi-Site Data Collection for a Spoken Language
  Corpus - MAD COW}. In \bibinfo{booktitle}{\emph{Second International
  Conference on Spoken Language Processing (ICSLP'92)}}.
  \bibinfo{publisher}{International Speech Communication Association},
  \bibinfo{address}{Banff, Canada}, \bibinfo{pages}{903--906}.
\newblock


\bibitem[\protect\citeauthoryear{Hou, Li, Li, and Lin}{Hou
  et~al\mbox{.}}{2019}]%
        {hou2019taskoriented}
\bibfield{author}{\bibinfo{person}{Lixian Hou}, \bibinfo{person}{Yanling Li},
  \bibinfo{person}{Chengcheng Li}, {and} \bibinfo{person}{Min Lin}.}
  \bibinfo{year}{2019}\natexlab{}.
\newblock \showarticletitle{Review of Research on Task-Oriented Spoken Language
  Understanding}.
\newblock \bibinfo{journal}{\emph{Journal of Physics: Conference Series}}
  \bibinfo{volume}{1267} (\bibinfo{date}{Jul} \bibinfo{year}{2019}),
  \bibinfo{pages}{012023}.
\newblock
\urldef\tempurl%
\url{https://doi.org/10.1088/1742-6596/1267/1/012023}
\showDOI{\tempurl}


\bibitem[\protect\citeauthoryear{{Jeong} and {Lee}}{{Jeong} and {Lee}}{2008}]%
        {jeong2008triangular}
\bibfield{author}{\bibinfo{person}{Minwoo {Jeong}} {and}
  \bibinfo{person}{Gary~Geunbae {Lee}}.} \bibinfo{year}{2008}\natexlab{}.
\newblock \showarticletitle{Triangular-Chain Conditional Random Fields}.
\newblock \bibinfo{journal}{\emph{IEEE Transactions on Audio, Speech, and
  Language Processing}} \bibinfo{volume}{16}, \bibinfo{number}{7}
  (\bibinfo{year}{2008}), \bibinfo{pages}{1287--1302}.
\newblock


\bibitem[\protect\citeauthoryear{Jung, Lee, and Kim}{Jung
  et~al\mbox{.}}{2018}]%
        {jung2018learning}
\bibfield{author}{\bibinfo{person}{Sangkeun Jung}, \bibinfo{person}{Jinsik
  Lee}, {and} \bibinfo{person}{Jiwon Kim}.} \bibinfo{year}{2018}\natexlab{}.
\newblock \showarticletitle{Learning to Embed Semantic Correspondence for
  Natural Language Understanding}. In \bibinfo{booktitle}{\emph{Proceedings of
  the 22nd Conference on Computational Natural Language Learning}}.
  \bibinfo{publisher}{Association for Computational Linguistics},
  \bibinfo{address}{Brussels, Belgium}, \bibinfo{pages}{131--140}.
\newblock
\urldef\tempurl%
\url{https://doi.org/10.18653/v1/K18-1013}
\showDOI{\tempurl}


\bibitem[\protect\citeauthoryear{Kanhabua, Ngoc~Nguyen, and Nejdl}{Kanhabua
  et~al\mbox{.}}{2015}]%
        {kanhabua2015learning}
\bibfield{author}{\bibinfo{person}{Nattiya Kanhabua}, \bibinfo{person}{Tu
  Ngoc~Nguyen}, {and} \bibinfo{person}{Wolfgang Nejdl}.}
  \bibinfo{year}{2015}\natexlab{}.
\newblock \showarticletitle{Learning to Detect Event-Related Queries for Web
  Search}. In \bibinfo{booktitle}{\emph{Proceedings of the 24th International
  Conference on World Wide Web}} (Florence, Italy) \emph{(\bibinfo{series}{WWW
  ’15 Companion})}. \bibinfo{publisher}{Association for Computing Machinery},
  \bibinfo{address}{New York, NY, USA}, \bibinfo{pages}{1339–1344}.
\newblock
\showISBNx{9781450334730}
\urldef\tempurl%
\url{https://doi.org/10.1145/2740908.2741698}
\showDOI{\tempurl}


\bibitem[\protect\citeauthoryear{Kim and Kim}{Kim and Kim}{2018}]%
        {kim2018ood}
\bibfield{author}{\bibinfo{person}{Joo-Kyung Kim} {and}
  \bibinfo{person}{Young-Bum Kim}.} \bibinfo{year}{2018}\natexlab{}.
\newblock \showarticletitle{Joint Learning of Domain Classification and
  Out-of-Domain Detection with Dynamic Class Weighting for Satisficing False
  Acceptance Rates}. In \bibinfo{booktitle}{\emph{Proc. Interspeech 2018}}.
  \bibinfo{publisher}{{ISCA}}, \bibinfo{address}{Hyderabad, India},
  \bibinfo{pages}{556--560}.
\newblock
\urldef\tempurl%
\url{https://doi.org/10.21437/Interspeech.2018-1581}
\showDOI{\tempurl}


\bibitem[\protect\citeauthoryear{Kim, Jha, Williams, Marin, and Zitouni}{Kim
  et~al\mbox{.}}{2019}]%
        {kim2019slot}
\bibfield{author}{\bibinfo{person}{Kunho Kim}, \bibinfo{person}{Rahul Jha},
  \bibinfo{person}{Kyle Williams}, \bibinfo{person}{Alex Marin}, {and}
  \bibinfo{person}{Imed Zitouni}.} \bibinfo{year}{2019}\natexlab{}.
\newblock \showarticletitle{Slot Tagging for Task Oriented Spoken Language
  Understanding in Human-to-Human Conversation Scenarios}. In
  \bibinfo{booktitle}{\emph{Proceedings of the 23rd Conference on Computational
  Natural Language Learning (CoNLL)}}. \bibinfo{publisher}{Association for
  Computational Linguistics}, \bibinfo{address}{Hong Kong, China},
  \bibinfo{pages}{757--767}.
\newblock
\urldef\tempurl%
\url{https://doi.org/10.18653/v1/K19-1071}
\showDOI{\tempurl}


\bibitem[\protect\citeauthoryear{{Kim}, {Lee}, and {Stratos}}{{Kim}
  et~al\mbox{.}}{2017}]%
        {kim2017onenet}
\bibfield{author}{\bibinfo{person}{Young-Bum {Kim}}, \bibinfo{person}{Sungjin
  {Lee}}, {and} \bibinfo{person}{Karl {Stratos}}.}
  \bibinfo{year}{2017}\natexlab{}.
\newblock \showarticletitle{ONENET: Joint domain, intent, slot prediction for
  spoken language understanding}. In \bibinfo{booktitle}{\emph{2017 IEEE
  Automatic Speech Recognition and Understanding Workshop (ASRU)}}.
  \bibinfo{publisher}{{IEEE}}, \bibinfo{address}{Okinawa, Japan},
  \bibinfo{pages}{547--553}.
\newblock


\bibitem[\protect\citeauthoryear{Korpusik, Liu, and Glass}{Korpusik
  et~al\mbox{.}}{2019}]%
        {korpusik2019comparison}
\bibfield{author}{\bibinfo{person}{Mandy Korpusik}, \bibinfo{person}{Zoe Liu},
  {and} \bibinfo{person}{James Glass}.} \bibinfo{year}{2019}\natexlab{}.
\newblock \showarticletitle{{A Comparison of Deep Learning Methods for Language
  Understanding}}. In \bibinfo{booktitle}{\emph{Proc. Interspeech 2019}}.
  \bibinfo{publisher}{{ISCA}}, \bibinfo{address}{Graz, Austria},
  \bibinfo{pages}{849--853}.
\newblock
\urldef\tempurl%
\url{https://doi.org/10.21437/Interspeech.2019-1262}
\showDOI{\tempurl}


\bibitem[\protect\citeauthoryear{Krone, Zhang, and Diab}{Krone
  et~al\mbox{.}}{2020}]%
        {krone2020learning}
\bibfield{author}{\bibinfo{person}{Jason Krone}, \bibinfo{person}{Yi Zhang},
  {and} \bibinfo{person}{Mona Diab}.} \bibinfo{year}{2020}\natexlab{}.
\newblock \bibinfo{title}{Learning to Classify Intents and Slot Labels Given a
  Handful of Examples}.
\newblock
\newblock
\showeprint[arxiv]{2004.10793}~[cs.CL]


\bibitem[\protect\citeauthoryear{Kurata, Xiang, Zhou, and Yu}{Kurata
  et~al\mbox{.}}{2016}]%
        {kurata2016leveraging}
\bibfield{author}{\bibinfo{person}{Gakuto Kurata}, \bibinfo{person}{Bing
  Xiang}, \bibinfo{person}{Bowen Zhou}, {and} \bibinfo{person}{Mo Yu}.}
  \bibinfo{year}{2016}\natexlab{}.
\newblock \showarticletitle{Leveraging Sentence-level Information with Encoder
  {LSTM} for Semantic Slot Filling}. In \bibinfo{booktitle}{\emph{Proceedings
  of the 2016 Conference on Empirical Methods in Natural Language Processing}}.
  \bibinfo{publisher}{Association for Computational Linguistics},
  \bibinfo{address}{Austin, Texas}, \bibinfo{pages}{2077--2083}.
\newblock
\urldef\tempurl%
\url{https://doi.org/10.18653/v1/D16-1223}
\showDOI{\tempurl}


\bibitem[\protect\citeauthoryear{{Lee}, {Kim}, {Sarikaya}, and {Kim}}{{Lee}
  et~al\mbox{.}}{2018}]%
        {lee2018coupled}
\bibfield{author}{\bibinfo{person}{Jihwan {Lee}}, \bibinfo{person}{Dongchan
  {Kim}}, \bibinfo{person}{Ruhi {Sarikaya}}, {and} \bibinfo{person}{Young-Bum
  {Kim}}.} \bibinfo{year}{2018}\natexlab{}.
\newblock \showarticletitle{Coupled Representation Learning for Domains,
  Intents and Slots in Spoken Language Understanding}. In
  \bibinfo{booktitle}{\emph{2018 IEEE Spoken Language Technology Workshop
  (SLT)}}. \bibinfo{publisher}{{IEEE}}, \bibinfo{address}{Athens, Greece},
  \bibinfo{pages}{714--719}.
\newblock


\bibitem[\protect\citeauthoryear{Li, Kong, and Zhao}{Li et~al\mbox{.}}{2018a}]%
        {li2018joint}
\bibfield{author}{\bibinfo{person}{Changliang Li}, \bibinfo{person}{Cunliang
  Kong}, {and} \bibinfo{person}{Yan Zhao}.} \bibinfo{year}{2018}\natexlab{a}.
\newblock \showarticletitle{A Joint Multi-Task Learning Framework for Spoken
  Language Understanding}. In \bibinfo{booktitle}{\emph{2018 IEEE International
  Conference on Acoustics, Speech and Signal Processing (ICASSP)}}.
  \bibinfo{publisher}{{IEEE}}, \bibinfo{address}{Calgary, Canada},
  \bibinfo{pages}{6054--6058}.
\newblock


\bibitem[\protect\citeauthoryear{Li, Li, and Qi}{Li et~al\mbox{.}}{2018b}]%
        {li2018self}
\bibfield{author}{\bibinfo{person}{Changliang Li}, \bibinfo{person}{Liang Li},
  {and} \bibinfo{person}{Ji Qi}.} \bibinfo{year}{2018}\natexlab{b}.
\newblock \showarticletitle{A Self-Attentive Model with Gate Mechanism for
  Spoken Language Understanding}. In \bibinfo{booktitle}{\emph{Proceedings of
  the 2018 Conference on Empirical Methods in Natural Language Processing}}.
  \bibinfo{publisher}{Association for Computational Linguistics},
  \bibinfo{address}{Brussels, Belgium}, \bibinfo{pages}{3824--3833}.
\newblock
\urldef\tempurl%
\url{https://doi.org/10.18653/v1/D18-1417}
\showDOI{\tempurl}


\bibitem[\protect\citeauthoryear{Li, Zhao, and Yu}{Li et~al\mbox{.}}{2019}]%
        {li2019conditional}
\bibfield{author}{\bibinfo{person}{Changliang Li}, \bibinfo{person}{Yan Zhao},
  {and} \bibinfo{person}{Dong Yu}.} \bibinfo{year}{2019}\natexlab{}.
\newblock \showarticletitle{Conditional Joint Model for Spoken Dialogue
  System}. In \bibinfo{booktitle}{\emph{Cognitive Computing -- ICCC 2019}},
  \bibfield{editor}{\bibinfo{person}{Ruifeng Xu}, \bibinfo{person}{Jianzong
  Wang}, {and} \bibinfo{person}{Liang-Jie Zhang}} (Eds.).
  \bibinfo{publisher}{Springer International Publishing},
  \bibinfo{address}{Cham}, \bibinfo{pages}{26--36}.
\newblock
\showISBNx{978-3-030-23407-2}


\bibitem[\protect\citeauthoryear{Lin and Xu}{Lin and Xu}{2019}]%
        {Lin_2019}
\bibfield{author}{\bibinfo{person}{Ting-En Lin} {and} \bibinfo{person}{Hua
  Xu}.} \bibinfo{year}{2019}\natexlab{}.
\newblock \showarticletitle{Deep Unknown Intent Detection with Margin Loss}. In
  \bibinfo{booktitle}{\emph{Proceedings of the 57th Annual Meeting of the
  Association for Computational Linguistics}}. \bibinfo{publisher}{Association
  for Computational Linguistics}, \bibinfo{address}{Florence, Italy},
  \bibinfo{pages}{5491--5496}.
\newblock
\urldef\tempurl%
\url{https://doi.org/10.18653/v1/P19-1548}
\showDOI{\tempurl}


\bibitem[\protect\citeauthoryear{Liu and Lane}{Liu and Lane}{2015}]%
        {liu2015recurrent}
\bibfield{author}{\bibinfo{person}{Bing Liu} {and} \bibinfo{person}{Ian Lane}.}
  \bibinfo{year}{2015}\natexlab{}.
\newblock \showarticletitle{Recurrent Neural Network Structured Output
  Prediction for Spoken Language Understanding}. In
  \bibinfo{booktitle}{\emph{Proceedings of {NIPS} Workshop on Machine Learning
  for Spoken Language Understanding and Interactions}}.
  \bibinfo{publisher}{Association for Computational Linguistics},
  \bibinfo{address}{Montreal, Canada}.
\newblock


\bibitem[\protect\citeauthoryear{Liu and Lane}{Liu and Lane}{2016a}]%
        {liu2016attention}
\bibfield{author}{\bibinfo{person}{Bing Liu} {and} \bibinfo{person}{Ian Lane}.}
  \bibinfo{year}{2016}\natexlab{a}.
\newblock \showarticletitle{Attention-Based Recurrent Neural Network Models for
  Joint Intent Detection and Slot Filling}. In
  \bibinfo{booktitle}{\emph{Interspeech 2016}}. \bibinfo{publisher}{{ISCA}},
  \bibinfo{address}{San Francisco, USA}, \bibinfo{pages}{685--689}.
\newblock
\urldef\tempurl%
\url{https://doi.org/10.21437/Interspeech.2016-1352}
\showDOI{\tempurl}


\bibitem[\protect\citeauthoryear{Liu and Lane}{Liu and Lane}{2016b}]%
        {liu2016joint}
\bibfield{author}{\bibinfo{person}{Bing Liu} {and} \bibinfo{person}{Ian Lane}.}
  \bibinfo{year}{2016}\natexlab{b}.
\newblock \showarticletitle{Joint Online Spoken Language Understanding and
  Language Modeling with Recurrent Neural Networks}. In
  \bibinfo{booktitle}{\emph{Proceedings of the SIGDIAL 2016 Conference}}.
  \bibinfo{publisher}{Association for Computational Linguistics},
  \bibinfo{address}{Los Angeles, USA}, \bibinfo{pages}{22--30}.
\newblock
\showeprint[arxiv]{1609.01462}
\urldef\tempurl%
\url{http://arxiv.org/abs/1609.01462}
\showURL{%
\tempurl}


\bibitem[\protect\citeauthoryear{Liu, Li, and Lin}{Liu et~al\mbox{.}}{2019a}]%
        {liu2019intentdetection}
\bibfield{author}{\bibinfo{person}{Jiao Liu}, \bibinfo{person}{Yanling Li},
  {and} \bibinfo{person}{Min Lin}.} \bibinfo{year}{2019}\natexlab{a}.
\newblock \showarticletitle{Review of Intent Detection Methods in the
  Human-Machine Dialogue System}.
\newblock \bibinfo{journal}{\emph{Journal of Physics: Conference Series}}
  \bibinfo{volume}{1267} (\bibinfo{date}{Jul} \bibinfo{year}{2019}),
  \bibinfo{pages}{012059}.
\newblock
\urldef\tempurl%
\url{https://doi.org/10.1088/1742-6596/1267/1/012059}
\showDOI{\tempurl}


\bibitem[\protect\citeauthoryear{Liu, Meng, Zhang, Zhou, Chen, and Xu}{Liu
  et~al\mbox{.}}{2019b}]%
        {liu2019cm}
\bibfield{author}{\bibinfo{person}{Yijin Liu}, \bibinfo{person}{Fandong Meng},
  \bibinfo{person}{Jinchao Zhang}, \bibinfo{person}{Jie Zhou},
  \bibinfo{person}{Yufeng Chen}, {and} \bibinfo{person}{Jinan Xu}.}
  \bibinfo{year}{2019}\natexlab{b}.
\newblock \showarticletitle{{CM}-Net: A Novel Collaborative Memory Network for
  Spoken Language Understanding}. In \bibinfo{booktitle}{\emph{Proceedings of
  the 2019 Conference on Empirical Methods in Natural Language Processing and
  the 9th International Joint Conference on Natural Language Processing
  (EMNLP-IJCNLP)}}. \bibinfo{publisher}{Association for Computational
  Linguistics}, \bibinfo{address}{Hong Kong, China},
  \bibinfo{pages}{1051--1060}.
\newblock
\urldef\tempurl%
\url{https://doi.org/10.18653/v1/D19-1097}
\showDOI{\tempurl}


\bibitem[\protect\citeauthoryear{Liu, Shin, Xu, Winata, Xu, Madotto, and
  Fung}{Liu et~al\mbox{.}}{2019c}]%
        {liu2019zerocross}
\bibfield{author}{\bibinfo{person}{Zihan Liu}, \bibinfo{person}{Jamin Shin},
  \bibinfo{person}{Yan Xu}, \bibinfo{person}{Genta~Indra Winata},
  \bibinfo{person}{Peng Xu}, \bibinfo{person}{Andrea Madotto}, {and}
  \bibinfo{person}{Pascale Fung}.} \bibinfo{year}{2019}\natexlab{c}.
\newblock \showarticletitle{Zero-shot Cross-lingual Dialogue Systems with
  Transferable Latent Variables}. In \bibinfo{booktitle}{\emph{Proceedings of
  the 2019 Conference on Empirical Methods in Natural Language Processing and
  the 9th International Joint Conference on Natural Language Processing
  (EMNLP-IJCNLP)}}. \bibinfo{publisher}{Association for Computational
  Linguistics}, \bibinfo{address}{Hong Kong, China},
  \bibinfo{pages}{1297--1303}.
\newblock
\urldef\tempurl%
\url{https://doi.org/10.18653/v1/D19-1129}
\showDOI{\tempurl}


\bibitem[\protect\citeauthoryear{Louvan and Magnini}{Louvan and
  Magnini}{2018}]%
        {louvan2018exploring}
\bibfield{author}{\bibinfo{person}{Samuel Louvan} {and}
  \bibinfo{person}{Bernardo Magnini}.} \bibinfo{year}{2018}\natexlab{}.
\newblock \showarticletitle{Exploring Named Entity Recognition As an Auxiliary
  Task for Slot Filling in Conversational Language Understanding}. In
  \bibinfo{booktitle}{\emph{Proceedings of the 2018 {EMNLP} Workshop {SCAI}:
  The 2nd International Workshop on Search-Oriented Conversational {AI}}}.
  \bibinfo{publisher}{Association for Computational Linguistics},
  \bibinfo{address}{Brussels, Belgium}, \bibinfo{pages}{74--80}.
\newblock
\urldef\tempurl%
\url{https://doi.org/10.18653/v1/W18-5711}
\showDOI{\tempurl}


\bibitem[\protect\citeauthoryear{Louvan and Magnini}{Louvan and
  Magnini}{2019}]%
        {louvan2019leveraging}
\bibfield{author}{\bibinfo{person}{Samuel Louvan} {and}
  \bibinfo{person}{Bernardo Magnini}.} \bibinfo{year}{2019}\natexlab{}.
\newblock \showarticletitle{Leveraging Non-Conversational Tasks for Low
  Resource Slot Filling: Does it help?}. In
  \bibinfo{booktitle}{\emph{Proceedings of the 20th Annual SIGdial Meeting on
  Discourse and Dialogue}}. \bibinfo{publisher}{Association for Computational
  Linguistics}, \bibinfo{address}{Stockholm, Sweden}, \bibinfo{pages}{85--91}.
\newblock
\urldef\tempurl%
\url{https://doi.org/10.18653/v1/W19-5911}
\showDOI{\tempurl}


\bibitem[\protect\citeauthoryear{Ma, Zhao, Huang, Xiang, and Zhou}{Ma
  et~al\mbox{.}}{2017}]%
        {ma2017jointly}
\bibfield{author}{\bibinfo{person}{Mingbo Ma}, \bibinfo{person}{Kai Zhao},
  \bibinfo{person}{Liang Huang}, \bibinfo{person}{Bing Xiang}, {and}
  \bibinfo{person}{Bowen Zhou}.} \bibinfo{year}{2017}\natexlab{}.
\newblock \showarticletitle{Jointly Trained Sequential Labeling and
  Classification by Sparse Attention Neural Networks}. In
  \bibinfo{booktitle}{\emph{Interspeech}}. \bibinfo{publisher}{{ISCA}},
  \bibinfo{address}{Stockholm, Sweden}, \bibinfo{pages}{3334--3338}.
\newblock


\bibitem[\protect\citeauthoryear{Masumura, Shinohara, Higashinaka, and
  Aono}{Masumura et~al\mbox{.}}{2018}]%
        {Masumura_2018}
\bibfield{author}{\bibinfo{person}{Ryo Masumura}, \bibinfo{person}{Yusuke
  Shinohara}, \bibinfo{person}{Ryuichiro Higashinaka}, {and}
  \bibinfo{person}{Yushi Aono}.} \bibinfo{year}{2018}\natexlab{}.
\newblock \showarticletitle{Adversarial Training for Multi-task and
  Multi-lingual Joint Modeling of Utterance Intent Classification}. In
  \bibinfo{booktitle}{\emph{Proceedings of the 2018 Conference on Empirical
  Methods in Natural Language Processing}}. \bibinfo{publisher}{Association for
  Computational Linguistics}, \bibinfo{address}{Brussels, Belgium},
  \bibinfo{pages}{633--639}.
\newblock
\urldef\tempurl%
\url{https://doi.org/10.18653/v1/d18-1064}
\showDOI{\tempurl}


\bibitem[\protect\citeauthoryear{Mesnil, Dauphin, Yao, Bengio, Deng,
  Hakkani-Tur, He, Heck, Tur, Yu, and Zweig}{Mesnil et~al\mbox{.}}{2015}]%
        {mesnil2015recurrent}
\bibfield{author}{\bibinfo{person}{Gr\'egoire Mesnil}, \bibinfo{person}{Yann
  Dauphin}, \bibinfo{person}{Kaisheng Yao}, \bibinfo{person}{Yoshua Bengio},
  \bibinfo{person}{Li Deng}, \bibinfo{person}{Dilek Hakkani-Tur},
  \bibinfo{person}{Xiaodong He}, \bibinfo{person}{Larry Heck},
  \bibinfo{person}{Gokhan Tur}, \bibinfo{person}{Dong Yu}, {and}
  \bibinfo{person}{Geoffrey Zweig}.} \bibinfo{year}{2015}\natexlab{}.
\newblock \showarticletitle{Using Recurrent Neural Networks for Slot Filling in
  Spoken Language Understanding}.
\newblock \bibinfo{journal}{\emph{IEEE/ACM Trans. Audio, Speech and Language
  Processing}} \bibinfo{volume}{23}, \bibinfo{number}{3} (\bibinfo{date}{March}
  \bibinfo{year}{2015}), \bibinfo{pages}{530–539}.
\newblock
\showISSN{2329-9290}
\urldef\tempurl%
\url{https://doi.org/10.1109/TASLP.2014.2383614}
\showDOI{\tempurl}


\bibitem[\protect\citeauthoryear{Mesnil, He, Deng, and Bengio}{Mesnil
  et~al\mbox{.}}{2013}]%
        {mesnil2013investigation}
\bibfield{author}{\bibinfo{person}{Gr\'egoire Mesnil},
  \bibinfo{person}{Xiaodong He}, \bibinfo{person}{Li Deng}, {and}
  \bibinfo{person}{Yoshua Bengio}.} \bibinfo{year}{2013}\natexlab{}.
\newblock \showarticletitle{Investigation of recurrent-neural-network
  architectures and learning methods for spoken language understanding}. In
  \bibinfo{booktitle}{\emph{Interspeech}}. \bibinfo{publisher}{ISCA},
  \bibinfo{address}{Lyon, France}, \bibinfo{pages}{3771--3775}.
\newblock


\bibitem[\protect\citeauthoryear{Mohasseb, Bader-El-Den, and Cocea}{Mohasseb
  et~al\mbox{.}}{2018}]%
        {Mohasseb_2018}
\bibfield{author}{\bibinfo{person}{Alaa Mohasseb}, \bibinfo{person}{Mohamed
  Bader-El-Den}, {and} \bibinfo{person}{Mihaela Cocea}.}
  \bibinfo{year}{2018}\natexlab{}.
\newblock \showarticletitle{Classification of factoid questions intent using
  grammatical features}.
\newblock \bibinfo{journal}{\emph{{ICT} Express}} \bibinfo{volume}{4},
  \bibinfo{number}{4} (\bibinfo{date}{Dec} \bibinfo{year}{2018}),
  \bibinfo{pages}{239--242}.
\newblock
\urldef\tempurl%
\url{https://doi.org/10.1016/j.icte.2018.10.004}
\showDOI{\tempurl}


\bibitem[\protect\citeauthoryear{Ni, Li, Li, and Chang}{Ni
  et~al\mbox{.}}{2020}]%
        {ni2020NLUIoT}
\bibfield{author}{\bibinfo{person}{Pin Ni}, \bibinfo{person}{Yuming Li},
  \bibinfo{person}{Gangmin Li}, {and} \bibinfo{person}{Victor Chang}.}
  \bibinfo{year}{2020}\natexlab{}.
\newblock \showarticletitle{Natural language understanding approaches based on
  joint task of intent detection and slot filling for IoT voice interaction}.
\newblock \bibinfo{journal}{\emph{Neural Computing and Applications}}
  \bibinfo{volume}{32} (\bibinfo{year}{2020}).
\newblock
\showISBNx{1433-3058}
\urldef\tempurl%
\url{https://doi.org/10.1007/s00521-020-04805-x}
\showDOI{\tempurl}


\bibitem[\protect\citeauthoryear{Niu and Penn}{Niu and Penn}{2019}]%
        {niu2019rationally}
\bibfield{author}{\bibinfo{person}{Jingcheng Niu} {and} \bibinfo{person}{Gerald
  Penn}.} \bibinfo{year}{2019}\natexlab{}.
\newblock \showarticletitle{Rationally Reappraising {ATIS}-based Dialogue
  Systems}. In \bibinfo{booktitle}{\emph{Proceedings of the 57th Annual Meeting
  of the Association for Computational Linguistics}}.
  \bibinfo{publisher}{Association for Computational Linguistics},
  \bibinfo{address}{Florence, Italy}, \bibinfo{pages}{5503--5507}.
\newblock


\bibitem[\protect\citeauthoryear{Okur, Kumar, Sahay, Esme, and Nachman}{Okur
  et~al\mbox{.}}{2019}]%
        {okur2019natural}
\bibfield{author}{\bibinfo{person}{Eda Okur}, \bibinfo{person}{Shachi~H Kumar},
  \bibinfo{person}{Saurav Sahay}, \bibinfo{person}{Asli~Arslan Esme}, {and}
  \bibinfo{person}{Lama Nachman}.} \bibinfo{year}{2019}\natexlab{}.
\newblock \bibinfo{title}{Natural Language Interactions in Autonomous Vehicles:
  Intent Detection and Slot Filling from Passenger Utterances}.
\newblock
\newblock
\showeprint[arxiv]{1904.10500}~[cs.CL]


\bibitem[\protect\citeauthoryear{Pallett, Dahlgren, Fiscus, Fisher, Garofolo,
  and Tjaden}{Pallett et~al\mbox{.}}{1992}]%
        {pallett1992atis1}
\bibfield{author}{\bibinfo{person}{David~S. Pallett}, \bibinfo{person}{Nancy~L.
  Dahlgren}, \bibinfo{person}{Jonathan~G. Fiscus}, \bibinfo{person}{William~M.
  Fisher}, \bibinfo{person}{John~S. Garofolo}, {and} \bibinfo{person}{Brett~C.
  Tjaden}.} \bibinfo{year}{1992}\natexlab{}.
\newblock \showarticletitle{DARPA February 1992 ATIS Benchmark Test Results}.
  In \bibinfo{booktitle}{\emph{Proceedings of the Workshop on Speech and
  Natural Language}} (Harriman, New York) \emph{(\bibinfo{series}{HLT ’91})}.
  \bibinfo{publisher}{Association for Computational Linguistics},
  \bibinfo{address}{USA}, \bibinfo{pages}{15–27}.
\newblock
\showISBNx{1558602720}
\urldef\tempurl%
\url{https://doi.org/10.3115/1075527.1075532}
\showDOI{\tempurl}


\bibitem[\protect\citeauthoryear{Pan, Zhang, Ren, Hou, Li, Liang, and Liu}{Pan
  et~al\mbox{.}}{2018}]%
        {pan2018multiple}
\bibfield{author}{\bibinfo{person}{Lingfeng Pan}, \bibinfo{person}{Yi Zhang},
  \bibinfo{person}{Feiliang Ren}, \bibinfo{person}{Yining Hou},
  \bibinfo{person}{Yan Li}, \bibinfo{person}{Xiaobo Liang}, {and}
  \bibinfo{person}{Yongkang Liu}.} \bibinfo{year}{2018}\natexlab{}.
\newblock \showarticletitle{A Multiple Utterances Based Neural Network Model
  for Joint Intent Detection and Slot Filling}. In
  \bibinfo{booktitle}{\emph{Proceedings of the Evaluation Tasks at the China
  Conference on Knowledge Graph and Semantic Computing (CCKS 2018)}}.
  \bibinfo{publisher}{CEUR-WS.org}, \bibinfo{address}{Tianjin, China},
  \bibinfo{pages}{25--33}.
\newblock


\bibitem[\protect\citeauthoryear{Peng and Yao}{Peng and Yao}{2015}]%
        {peng2015recurrent}
\bibfield{author}{\bibinfo{person}{Baolin Peng} {and} \bibinfo{person}{Kaisheng
  Yao}.} \bibinfo{year}{2015}\natexlab{}.
\newblock \bibinfo{title}{Recurrent Neural Networks with External Memory for
  Language Understanding}.
\newblock
\newblock
\showeprint[arxiv]{1506.00195}~[cs.CL]


\bibitem[\protect\citeauthoryear{Pentyala, Liu, and Dreyer}{Pentyala
  et~al\mbox{.}}{2019}]%
        {pentyala2019multi}
\bibfield{author}{\bibinfo{person}{Shiva Pentyala}, \bibinfo{person}{Mengwen
  Liu}, {and} \bibinfo{person}{Markus Dreyer}.}
  \bibinfo{year}{2019}\natexlab{}.
\newblock \showarticletitle{Multi-Task Networks with Universe, Group, and Task
  Feature Learning}. In \bibinfo{booktitle}{\emph{Proceedings of the 57th
  Annual Meeting of the Association for Computational Linguistics}}.
  \bibinfo{publisher}{Association for Computational Linguistics},
  \bibinfo{address}{Florence, Italy}, \bibinfo{pages}{820--830}.
\newblock
\urldef\tempurl%
\url{https://doi.org/10.18653/v1/P19-1079}
\showDOI{\tempurl}


\bibitem[\protect\citeauthoryear{Purohit, Dong, Shalin, Thirunarayan, and
  Sheth}{Purohit et~al\mbox{.}}{2015}]%
        {Purohit_2015}
\bibfield{author}{\bibinfo{person}{Hemant Purohit}, \bibinfo{person}{Guozhu
  Dong}, \bibinfo{person}{Valerie Shalin}, \bibinfo{person}{Krishnaprasad
  Thirunarayan}, {and} \bibinfo{person}{Amit Sheth}.}
  \bibinfo{year}{2015}\natexlab{}.
\newblock \showarticletitle{Intent Classification of Short-Text on Social
  Media}. In \bibinfo{booktitle}{\emph{2015 {IEEE} International Conference on
  Smart City/{SocialCom}/{SustainCom} ({SmartCity})}}.
  \bibinfo{publisher}{{IEEE}}, \bibinfo{address}{Chengdu, China},
  \bibinfo{pages}{222--228}.
\newblock
\urldef\tempurl%
\url{https://doi.org/10.1109/smartcity.2015.75}
\showDOI{\tempurl}


\bibitem[\protect\citeauthoryear{Qin, Che, Li, Wen, and Liu}{Qin
  et~al\mbox{.}}{2019}]%
        {qin2019stack}
\bibfield{author}{\bibinfo{person}{Libo Qin}, \bibinfo{person}{Wanxiang Che},
  \bibinfo{person}{Yangming Li}, \bibinfo{person}{Haoyang Wen}, {and}
  \bibinfo{person}{Ting Liu}.} \bibinfo{year}{2019}\natexlab{}.
\newblock \showarticletitle{A Stack-Propagation Framework with Token-Level
  Intent Detection for Spoken Language Understanding}. In
  \bibinfo{booktitle}{\emph{Proceedings of the 2019 Conference on Empirical
  Methods in Natural Language Processing and the 9th International Joint
  Conference on Natural Language Processing, {EMNLP-IJCNLP} 2019, Hong Kong,
  China, November 3-7, 2019}}, \bibfield{editor}{\bibinfo{person}{Kentaro
  Inui}, \bibinfo{person}{Jing Jiang}, \bibinfo{person}{Vincent Ng}, {and}
  \bibinfo{person}{Xiaojun Wan}} (Eds.). \bibinfo{publisher}{Association for
  Computational Linguistics}, \bibinfo{address}{Hong Kong},
  \bibinfo{pages}{2078--2087}.
\newblock
\urldef\tempurl%
\url{https://doi.org/10.18653/v1/D19-1214}
\showDOI{\tempurl}


\bibitem[\protect\citeauthoryear{Qin, Ni, Zhang, and Che}{Qin
  et~al\mbox{.}}{2020a}]%
        {qin2020cosdaml}
\bibfield{author}{\bibinfo{person}{Libo Qin}, \bibinfo{person}{Minheng Ni},
  \bibinfo{person}{Yue Zhang}, {and} \bibinfo{person}{Wanxiang Che}.}
  \bibinfo{year}{2020}\natexlab{a}.
\newblock \showarticletitle{CoSDA-ML: Multi-Lingual Code-Switching Data
  Augmentation for Zero-Shot Cross-Lingual NLP}. In
  \bibinfo{booktitle}{\emph{Proceedings of the Twenty-Ninth International Joint
  Conference on Artificial Intelligence, {IJCAI-20}}},
  \bibfield{editor}{\bibinfo{person}{Christian Bessiere}} (Ed.).
  \bibinfo{publisher}{International Joint Conferences on Artificial
  Intelligence Organization}, \bibinfo{address}{Yokohama, Japan},
  \bibinfo{pages}{3853--3860}.
\newblock
\urldef\tempurl%
\url{https://doi.org/10.24963/ijcai.2020/533}
\showDOI{\tempurl}
\newblock
\shownote{Main track.}


\bibitem[\protect\citeauthoryear{Qin, Xu, Che, and Liu}{Qin
  et~al\mbox{.}}{2020b}]%
        {qin2020agif}
\bibfield{author}{\bibinfo{person}{Libo Qin}, \bibinfo{person}{Xiao Xu},
  \bibinfo{person}{Wanxiang Che}, {and} \bibinfo{person}{Ting Liu}.}
  \bibinfo{year}{2020}\natexlab{b}.
\newblock \showarticletitle{{AGIF}: An Adaptive Graph-Interactive Framework for
  Joint Multiple Intent Detection and Slot Filling}. In
  \bibinfo{booktitle}{\emph{Findings of the Association for Computational
  Linguistics: EMNLP 2020}}. \bibinfo{publisher}{Association for Computational
  Linguistics}, \bibinfo{address}{Online}, \bibinfo{pages}{1807--1816}.
\newblock
\urldef\tempurl%
\url{https://doi.org/10.18653/v1/2020.findings-emnlp.163}
\showDOI{\tempurl}


\bibitem[\protect\citeauthoryear{Qiu, Chen, Jia, and Zhang}{Qiu
  et~al\mbox{.}}{2018}]%
        {Qiu_2018}
\bibfield{author}{\bibinfo{person}{Lirong Qiu}, \bibinfo{person}{Yida Chen},
  \bibinfo{person}{Haoran Jia}, {and} \bibinfo{person}{Zhen Zhang}.}
  \bibinfo{year}{2018}\natexlab{}.
\newblock \showarticletitle{Query Intent Recognition Based on Multi-Class
  Features}.
\newblock \bibinfo{journal}{\emph{{IEEE} Access}}  \bibinfo{volume}{6}
  (\bibinfo{year}{2018}), \bibinfo{pages}{52195--52204}.
\newblock
\urldef\tempurl%
\url{https://doi.org/10.1109/access.2018.2869585}
\showDOI{\tempurl}


\bibitem[\protect\citeauthoryear{Ravuri and Stolcke}{Ravuri and
  Stolcke}{2015}]%
        {ravuri2015recurrent}
\bibfield{author}{\bibinfo{person}{Suman Ravuri} {and} \bibinfo{person}{Andreas
  Stolcke}.} \bibinfo{year}{2015}\natexlab{}.
\newblock \showarticletitle{Recurrent Neural Network and LSTM Models for
  Lexical Utterance Classification}. In \bibinfo{booktitle}{\emph{Proc.
  Interspeech}}. \bibinfo{publisher}{ISCA - International Speech Communication
  Association}, \bibinfo{address}{Dresden, Germany}, \bibinfo{pages}{135--139}.
\newblock
\urldef\tempurl%
\url{https://www.microsoft.com/en-us/research/publication/recurrent-neural-network-and-lstm-models-for-lexical-utterance-classification/}
\showURL{%
\tempurl}


\bibitem[\protect\citeauthoryear{Ray, Shen, and Jin}{Ray et~al\mbox{.}}{2018}]%
        {ray2018robust}
\bibfield{author}{\bibinfo{person}{Avik Ray}, \bibinfo{person}{Yilin Shen},
  {and} \bibinfo{person}{Hongxia Jin}.} \bibinfo{year}{2018}\natexlab{}.
\newblock \showarticletitle{Robust Spoken Language Understanding via
  Paraphrasing}. In \bibinfo{booktitle}{\emph{Proc. Interspeech 2018}}.
  \bibinfo{publisher}{{ISCA}}, \bibinfo{address}{Hyderabad, India},
  \bibinfo{pages}{3454--3458}.
\newblock
\urldef\tempurl%
\url{https://doi.org/10.21437/Interspeech.2018-2358}
\showDOI{\tempurl}


\bibitem[\protect\citeauthoryear{Ray, Shen, and Jin}{Ray et~al\mbox{.}}{2019}]%
        {ray2019interative}
\bibfield{author}{\bibinfo{person}{Avik Ray}, \bibinfo{person}{Yilin Shen},
  {and} \bibinfo{person}{Hongxia Jin}.} \bibinfo{year}{2019}\natexlab{}.
\newblock \showarticletitle{Iterative Delexicalization for Improved Spoken
  Language Understanding}. In \bibinfo{booktitle}{\emph{Interspeech}}.
  \bibinfo{publisher}{{ISCA}}, \bibinfo{address}{Graz, Austria},
  \bibinfo{pages}{1183--1187}.
\newblock
\urldef\tempurl%
\url{https://doi.org/10.21437/Interspeech.2019-2955}
\showDOI{\tempurl}


\bibitem[\protect\citeauthoryear{Raymond and Riccardi}{Raymond and
  Riccardi}{2007}]%
        {raymond2007generative}
\bibfield{author}{\bibinfo{person}{Christian Raymond} {and}
  \bibinfo{person}{Giuseppe Riccardi}.} \bibinfo{year}{2007}\natexlab{}.
\newblock \showarticletitle{Generative and discriminative algorithms for spoken
  language understanding}. In \bibinfo{booktitle}{\emph{Interspeech}}.
  \bibinfo{publisher}{{ISCA}}, \bibinfo{address}{Antwerp, Belgium},
  \bibinfo{pages}{1605--1608}.
\newblock


\bibitem[\protect\citeauthoryear{{Ren} and {Xue}}{{Ren} and {Xue}}{2020}]%
        {ren2020siamese}
\bibfield{author}{\bibinfo{person}{Fuji {Ren}} {and} \bibinfo{person}{Siyuan
  {Xue}}.} \bibinfo{year}{2020}\natexlab{}.
\newblock \showarticletitle{Intention Detection Based on Siamese Neural Network
  With Triplet Loss}.
\newblock \bibinfo{journal}{\emph{IEEE Access}}  \bibinfo{volume}{8}
  (\bibinfo{year}{2020}), \bibinfo{pages}{82242--82254}.
\newblock


\bibitem[\protect\citeauthoryear{Sarikaya, Hinton, and Ramabhadran}{Sarikaya
  et~al\mbox{.}}{2011}]%
        {Sarikaya_2011}
\bibfield{author}{\bibinfo{person}{Ruhi Sarikaya}, \bibinfo{person}{Geoffrey~E.
  Hinton}, {and} \bibinfo{person}{Bhuvana Ramabhadran}.}
  \bibinfo{year}{2011}\natexlab{}.
\newblock \showarticletitle{Deep belief nets for natural language
  call-routing}. In \bibinfo{booktitle}{\emph{2011 {IEEE} International
  Conference on Acoustics, Speech and Signal Processing ({ICASSP})}}.
  \bibinfo{publisher}{{IEEE}}, \bibinfo{address}{Prague, Czech Republic},
  \bibinfo{pages}{5680--5683}.
\newblock
\urldef\tempurl%
\url{https://doi.org/10.1109/icassp.2011.5947649}
\showDOI{\tempurl}


\bibitem[\protect\citeauthoryear{Schuster, Gupta, Shah, and Lewis}{Schuster
  et~al\mbox{.}}{2019}]%
        {schuster2019crosslingual}
\bibfield{author}{\bibinfo{person}{Sebastian Schuster}, \bibinfo{person}{Sonal
  Gupta}, \bibinfo{person}{Rushin Shah}, {and} \bibinfo{person}{Mike Lewis}.}
  \bibinfo{year}{2019}\natexlab{}.
\newblock \showarticletitle{Cross-lingual Transfer Learning for Multilingual
  Task Oriented Dialog}. In \bibinfo{booktitle}{\emph{Proceedings of the 2019
  Conference of the North {A}merican Chapter of the Association for
  Computational Linguistics: Human Language Technologies, Volume 1 (Long and
  Short Papers)}}. \bibinfo{publisher}{Association for Computational
  Linguistics}, \bibinfo{address}{Minneapolis, Minnesota},
  \bibinfo{pages}{3795--3805}.
\newblock
\urldef\tempurl%
\url{https://doi.org/10.18653/v1/N19-1380}
\showDOI{\tempurl}


\bibitem[\protect\citeauthoryear{Serban, Lowe, Henderson, Charlin, and
  Pineau}{Serban et~al\mbox{.}}{2018}]%
        {serban2018corpora}
\bibfield{author}{\bibinfo{person}{Iulian~Vlad Serban}, \bibinfo{person}{Ryan
  Lowe}, \bibinfo{person}{Peter Henderson}, \bibinfo{person}{Laurent Charlin},
  {and} \bibinfo{person}{Joelle Pineau}.} \bibinfo{year}{2018}\natexlab{}.
\newblock \showarticletitle{A Survey of Available Corpora For Building
  Data-Driven Dialogue Systems: The Journal Version}.
\newblock \bibinfo{journal}{\emph{Dialogue \& Discourse}} \bibinfo{volume}{9},
  \bibinfo{number}{1} (\bibinfo{year}{2018}), \bibinfo{pages}{1--49}.
\newblock


\bibitem[\protect\citeauthoryear{Shen, Chen, and Jin}{Shen
  et~al\mbox{.}}{2019a}]%
        {shen2019interpreting}
\bibfield{author}{\bibinfo{person}{Yilin Shen}, \bibinfo{person}{Wenhu Chen},
  {and} \bibinfo{person}{Hongxia Jin}.} \bibinfo{year}{2019}\natexlab{a}.
\newblock \showarticletitle{Interpreting and Improving Deep Neural SLU Models
  via Vocabulary Importance}. In \bibinfo{booktitle}{\emph{Interspeech}}.
  \bibinfo{publisher}{{ISCA}}, \bibinfo{address}{Graz, Austria},
  \bibinfo{pages}{1328--1332}.
\newblock
\urldef\tempurl%
\url{https://doi.org/10.21437/Interspeech.2019-3184}
\showDOI{\tempurl}


\bibitem[\protect\citeauthoryear{Shen, Zeng, and Jin}{Shen
  et~al\mbox{.}}{2019b}]%
        {shen2019progressive}
\bibfield{author}{\bibinfo{person}{Yilin Shen}, \bibinfo{person}{Xiangyu Zeng},
  {and} \bibinfo{person}{Hongxia Jin}.} \bibinfo{year}{2019}\natexlab{b}.
\newblock \showarticletitle{A Progressive Model to Enable Continual Learning
  for Semantic Slot Filling}. In \bibinfo{booktitle}{\emph{Proceedings of the
  2019 Conference on Empirical Methods in Natural Language Processing and the
  9th International Joint Conference on Natural Language Processing
  (EMNLP-IJCNLP)}}. \bibinfo{publisher}{Association for Computational
  Linguistics}, \bibinfo{address}{Hong Kong, China},
  \bibinfo{pages}{1279--1284}.
\newblock
\urldef\tempurl%
\url{https://doi.org/10.18653/v1/D19-1126}
\showDOI{\tempurl}


\bibitem[\protect\citeauthoryear{Shen, Zeng, Wang, and Jin}{Shen
  et~al\mbox{.}}{2018}]%
        {shen2018user}
\bibfield{author}{\bibinfo{person}{Yilin Shen}, \bibinfo{person}{Xiangyu Zeng},
  \bibinfo{person}{Yu Wang}, {and} \bibinfo{person}{Hongxia Jin}.}
  \bibinfo{year}{2018}\natexlab{}.
\newblock \showarticletitle{User Information Augmented Semantic Frame Parsing
  Using Progressive Neural Networks}. In \bibinfo{booktitle}{\emph{Interspeech
  2018, 19th Annual Conference of the International Speech Communication
  Association, Hyderabad, India, 2-6 September 2018}},
  \bibfield{editor}{\bibinfo{person}{B.~Yegnanarayana}} (Ed.).
  \bibinfo{publisher}{{ISCA}}, \bibinfo{address}{Hyderabad, India},
  \bibinfo{pages}{3464--3468}.
\newblock
\urldef\tempurl%
\url{https://doi.org/10.21437/Interspeech.2018-1149}
\showDOI{\tempurl}


\bibitem[\protect\citeauthoryear{{Shi}, {Yao}, {Chen}, {Pan}, {Hwang}, and
  {Peng}}{{Shi} et~al\mbox{.}}{2015}]%
        {shi2015contextual}
\bibfield{author}{\bibinfo{person}{Yangyang {Shi}}, \bibinfo{person}{Kaisheng
  {Yao}}, \bibinfo{person}{Hu {Chen}}, \bibinfo{person}{Yi-Cheng {Pan}},
  \bibinfo{person}{Mei-Yuh {Hwang}}, {and} \bibinfo{person}{Baolin {Peng}}.}
  \bibinfo{year}{2015}\natexlab{}.
\newblock \showarticletitle{Contextual spoken language understanding using
  recurrent neural networks}. In \bibinfo{booktitle}{\emph{2015 IEEE
  International Conference on Acoustics, Speech and Signal Processing
  (ICASSP)}}. \bibinfo{publisher}{{IEEE}}, \bibinfo{address}{Brisbane,
  Australia}, \bibinfo{pages}{5271--5275}.
\newblock


\bibitem[\protect\citeauthoryear{Shin, Yoo, and Lee}{Shin
  et~al\mbox{.}}{2018}]%
        {shin2018slot}
\bibfield{author}{\bibinfo{person}{Youhyun Shin}, \bibinfo{person}{Kang Yoo},
  {and} \bibinfo{person}{Sang-goo Lee}.} \bibinfo{year}{2018}\natexlab{}.
\newblock \showarticletitle{Slot Filling with Delexicalized Sentence
  Generation}. In \bibinfo{booktitle}{\emph{Interspeech 2018, 19th Annual
  Conference of the International Speech Communication Association, Hyderabad,
  India, 2-6 September 2018}}. \bibinfo{publisher}{{ISCA}},
  \bibinfo{address}{Hyderabad, India}, \bibinfo{pages}{2082--2086}.
\newblock
\urldef\tempurl%
\url{https://doi.org/10.21437/Interspeech.2018-1808}
\showDOI{\tempurl}


\bibitem[\protect\citeauthoryear{Shridhar, Dash, Sahu, Pihlgren, Alonso,
  Pondenkandath, Kovacs, Simistira, and Liwicki}{Shridhar
  et~al\mbox{.}}{2019}]%
        {Shridhar_2019}
\bibfield{author}{\bibinfo{person}{Kumar Shridhar}, \bibinfo{person}{Ayushman
  Dash}, \bibinfo{person}{Amit Sahu}, \bibinfo{person}{Gustav~Grund Pihlgren},
  \bibinfo{person}{Pedro Alonso}, \bibinfo{person}{Vinaychandran
  Pondenkandath}, \bibinfo{person}{Gyorgy Kovacs}, \bibinfo{person}{Foteini
  Simistira}, {and} \bibinfo{person}{Marcus Liwicki}.}
  \bibinfo{year}{2019}\natexlab{}.
\newblock \showarticletitle{Subword Semantic Hashing for Intent Classification
  on Small Datasets}. In \bibinfo{booktitle}{\emph{2019 International Joint
  Conference on Neural Networks ({IJCNN})}}. \bibinfo{publisher}{{IEEE}},
  \bibinfo{address}{Budapest, Hungary}, \bibinfo{pages}{1--6}.
\newblock
\urldef\tempurl%
\url{https://doi.org/10.1109/ijcnn.2019.8852420}
\showDOI{\tempurl}


\bibitem[\protect\citeauthoryear{Siddhant, Goyal, and Metallinou}{Siddhant
  et~al\mbox{.}}{2019}]%
        {siddhant2018unsupervised}
\bibfield{author}{\bibinfo{person}{Aditya Siddhant},
  \bibinfo{person}{Anuj~Kumar Goyal}, {and} \bibinfo{person}{Angeliki
  Metallinou}.} \bibinfo{year}{2019}\natexlab{}.
\newblock \showarticletitle{Unsupervised Transfer Learning for Spoken Language
  Understanding in Intelligent Agents}. In
  \bibinfo{booktitle}{\emph{Proceedings of the AAAI Conference on Artificial
  Intelligence}}, Vol.~\bibinfo{volume}{33}. \bibinfo{publisher}{AAAI Press},
  \bibinfo{address}{Honolulu, USA}, \bibinfo{pages}{4959--4966}.
\newblock
\urldef\tempurl%
\url{https://doi.org/10.1609/aaai.v33i01.33014959}
\showDOI{\tempurl}


\bibitem[\protect\citeauthoryear{{Staliūnaitė} and
  {Iacobacci}}{{Staliūnaitė} and {Iacobacci}}{2020}]%
        {staliunaite2020capsule}
\bibfield{author}{\bibinfo{person}{Ieva {Staliūnaitė}} {and}
  \bibinfo{person}{Ignacio {Iacobacci}}.} \bibinfo{year}{2020}\natexlab{}.
\newblock \showarticletitle{Auxiliary Capsules for Natural Language
  Understanding}. In \bibinfo{booktitle}{\emph{ICASSP 2020 - 2020 IEEE
  International Conference on Acoustics, Speech and Signal Processing
  (ICASSP)}}. \bibinfo{publisher}{{IEEE}}, \bibinfo{address}{Barcelona, Spain},
  \bibinfo{pages}{8154--8158}.
\newblock


\bibitem[\protect\citeauthoryear{Tam, Shi, Chen, and Hwang}{Tam
  et~al\mbox{.}}{2015}]%
        {tam2015rnnbased}
\bibfield{author}{\bibinfo{person}{Yik-Cheung Tam}, \bibinfo{person}{Yangyang
  Shi}, \bibinfo{person}{Hunk Chen}, {and} \bibinfo{person}{Mei-Yuh Hwang}.}
  \bibinfo{year}{2015}\natexlab{}.
\newblock \showarticletitle{RNN-based labeled data generation for spoken
  language understanding}. In \bibinfo{booktitle}{\emph{Interspeech}}.
  \bibinfo{publisher}{{ISCA}}, \bibinfo{address}{Dresden, Germany},
  \bibinfo{pages}{125--129}.
\newblock


\bibitem[\protect\citeauthoryear{Tang, Ji, and Zhou}{Tang
  et~al\mbox{.}}{2020}]%
        {tang2020endtoendmask}
\bibfield{author}{\bibinfo{person}{Hao Tang}, \bibinfo{person}{Donghong Ji},
  {and} \bibinfo{person}{Qiji Zhou}.} \bibinfo{year}{2020}\natexlab{}.
\newblock \showarticletitle{End-to-end masked graph-based CRF for joint slot
  filling and intent detection}.
\newblock \bibinfo{journal}{\emph{Neurocomputing}}  \bibinfo{volume}{413}
  (\bibinfo{year}{2020}), \bibinfo{pages}{348--359}.
\newblock
\showISBNx{0925-2312}
\urldef\tempurl%
\url{https://doi.org/10.1016/j.neucom.2019.06.113}
\showDOI{\tempurl}


\bibitem[\protect\citeauthoryear{{Thi Do} and {Gaspers}}{{Thi Do} and
  {Gaspers}}{2019}]%
        {thido2019cross}
\bibfield{author}{\bibinfo{person}{Quynh~Ngoc {Thi Do}} {and}
  \bibinfo{person}{Judith {Gaspers}}.} \bibinfo{year}{2019}\natexlab{}.
\newblock \showarticletitle{Cross-lingual Transfer Learning for Spoken Language
  Understanding}. In \bibinfo{booktitle}{\emph{ICASSP 2019 - 2019 IEEE
  International Conference on Acoustics, Speech and Signal Processing
  (ICASSP)}}. \bibinfo{publisher}{{IEEE}}, \bibinfo{address}{Brighton, United
  Kingdom of Great Britain and Northern Ireland}, \bibinfo{pages}{5956--5960}.
\newblock


\bibitem[\protect\citeauthoryear{Tur, Celikyilmaz, He, Hakkani-T{\"u}r, and
  Deng}{Tur et~al\mbox{.}}{2018}]%
        {tur2018deeplearning}
\bibfield{author}{\bibinfo{person}{Gokhan Tur}, \bibinfo{person}{Asli
  Celikyilmaz}, \bibinfo{person}{Xiaodong He}, \bibinfo{person}{Dilek
  Hakkani-T{\"u}r}, {and} \bibinfo{person}{Li Deng}.}
  \bibinfo{year}{2018}\natexlab{}.
\newblock \showarticletitle{Deep Learning in Conversational Language
  Understanding}.
\newblock In \bibinfo{booktitle}{\emph{Deep Learning in Natural Language
  Processing}}, \bibfield{editor}{\bibinfo{person}{Li~Deng} {and}
  \bibinfo{person}{Yang Liu}} (Eds.). \bibinfo{publisher}{Springer Singapore},
  \bibinfo{address}{Singapore}, \bibinfo{pages}{23--48}.
\newblock
\showISBNx{978-981-10-5209-5}
\urldef\tempurl%
\url{https://doi.org/10.1007/978-981-10-5209-5_2}
\showDOI{\tempurl}


\bibitem[\protect\citeauthoryear{Tur and {De Mori}}{Tur and {De Mori}}{2011}]%
        {tur2011spoken}
\bibfield{author}{\bibinfo{person}{Gokhan Tur} {and} \bibinfo{person}{Renato
  {De Mori}}.} \bibinfo{year}{2011}\natexlab{}.
\newblock \bibinfo{booktitle}{\emph{Spoken Language Understanding: Systems for
  Extracting Semantic Information from Speech}}.
\newblock \bibinfo{publisher}{John Wiley and Sons}, \bibinfo{address}{New York,
  USA}.
\newblock
\urldef\tempurl%
\url{https://www.microsoft.com/en-us/research/publication/spoken-language-understanding-systems-for-extracting-semantic-information-from-speech/}
\showURL{%
\tempurl}


\bibitem[\protect\citeauthoryear{Tur, Hakkani-T{\"u}r, and Heck}{Tur
  et~al\mbox{.}}{2010}]%
        {tur2010left}
\bibfield{author}{\bibinfo{person}{Gokhan Tur}, \bibinfo{person}{Dilek
  Hakkani-T{\"u}r}, {and} \bibinfo{person}{Larry Heck}.}
  \bibinfo{year}{2010}\natexlab{}.
\newblock \showarticletitle{What is left to be understood in ATIS?}. In
  \bibinfo{booktitle}{\emph{2010 IEEE Spoken Language Technology Workshop}}.
  IEEE, \bibinfo{publisher}{{IEEE}}, \bibinfo{address}{Berkeley, USA},
  \bibinfo{pages}{19--24}.
\newblock


\bibitem[\protect\citeauthoryear{Tur, Hakkani-Tur, Heck, and Parthasarathy}{Tur
  et~al\mbox{.}}{2011}]%
        {tur2011sentence}
\bibfield{author}{\bibinfo{person}{Gokhan Tur}, \bibinfo{person}{Dilek
  Hakkani-Tur}, \bibinfo{person}{Larry Heck}, {and} \bibinfo{person}{S.
  Parthasarathy}.} \bibinfo{year}{2011}\natexlab{}.
\newblock \showarticletitle{Sentence simplification for spoken language
  understanding}. In \bibinfo{booktitle}{\emph{2011 {IEEE} International
  Conference on Acoustics, Speech and Signal Processing ({ICASSP})}}.
  \bibinfo{publisher}{{IEEE}}, \bibinfo{address}{Prague, Czech Republic},
  \bibinfo{pages}{5628 -- 5631}.
\newblock


\bibitem[\protect\citeauthoryear{Vaswani, Shazeer, Parmar, Uszkoreit, Jones,
  Gomez, Kaiser, and Polosukhin}{Vaswani et~al\mbox{.}}{2017}]%
        {vaswani2017attention}
\bibfield{author}{\bibinfo{person}{Ashish Vaswani}, \bibinfo{person}{Noam
  Shazeer}, \bibinfo{person}{Niki Parmar}, \bibinfo{person}{Jakob Uszkoreit},
  \bibinfo{person}{Llion Jones}, \bibinfo{person}{Aidan~N Gomez},
  \bibinfo{person}{{\L}ukasz Kaiser}, {and} \bibinfo{person}{Illia
  Polosukhin}.} \bibinfo{year}{2017}\natexlab{}.
\newblock \showarticletitle{Attention is all you need}.
\newblock In \bibinfo{booktitle}{\emph{Advances in neural information
  processing systems 30}}. \bibinfo{publisher}{Curran Associates, Inc.},
  \bibinfo{address}{Long Beach, USA}, \bibinfo{pages}{5998--6008}.
\newblock
\urldef\tempurl%
\url{http://papers.nips.cc/paper/7181-attention-is-all-you-need.pdf}
\showURL{%
\tempurl}


\bibitem[\protect\citeauthoryear{Veyseh, Dernoncourt, and Nguyen}{Veyseh
  et~al\mbox{.}}{2019}]%
        {veyseh2019improving}
\bibfield{author}{\bibinfo{person}{Amir Pouran~Ben Veyseh},
  \bibinfo{person}{Franck Dernoncourt}, {and} \bibinfo{person}{Thien~Huu
  Nguyen}.} \bibinfo{year}{2019}\natexlab{}.
\newblock \bibinfo{title}{Improving Slot Filling by Utilizing Contextual
  Information}.
\newblock
\newblock
\showeprint[arxiv]{1911.01680}~[cs.CL]


\bibitem[\protect\citeauthoryear{Vu}{Vu}{2016}]%
        {vu2016sequential}
\bibfield{author}{\bibinfo{person}{Thang Vu}.} \bibinfo{year}{2016}\natexlab{}.
\newblock \showarticletitle{Sequential Convolutional Neural Networks for Slot
  Filling in Spoken Language Understanding}. In
  \bibinfo{booktitle}{\emph{Interspeech}}. \bibinfo{publisher}{{ISCA}},
  \bibinfo{address}{San Francisco, USA}, \bibinfo{pages}{3250--3254}.
\newblock
\urldef\tempurl%
\url{https://doi.org/10.21437/Interspeech.2016-395}
\showDOI{\tempurl}


\bibitem[\protect\citeauthoryear{Vu, Gupta, Adel, and Schütze}{Vu
  et~al\mbox{.}}{2016}]%
        {vu2016bidirectional}
\bibfield{author}{\bibinfo{person}{Thang Vu}, \bibinfo{person}{Pankaj Gupta},
  \bibinfo{person}{Heike Adel}, {and} \bibinfo{person}{Hinrich Schütze}.}
  \bibinfo{year}{2016}\natexlab{}.
\newblock \showarticletitle{Bi-directional recurrent neural network with
  ranking loss for spoken language understanding}. In
  \bibinfo{booktitle}{\emph{2016 IEEE International Conference on Acoustics,
  Speech and Signal Processing (ICASSP)}}. \bibinfo{publisher}{{IEEE}},
  \bibinfo{address}{Shanghai, China}, \bibinfo{pages}{6060--6064}.
\newblock
\urldef\tempurl%
\url{https://doi.org/10.1109/ICASSP.2016.7472841}
\showDOI{\tempurl}


\bibitem[\protect\citeauthoryear{Wang, Huang, and Hu}{Wang
  et~al\mbox{.}}{2020b}]%
        {wang2020sasgbc}
\bibfield{author}{\bibinfo{person}{Congrui Wang}, \bibinfo{person}{Zhen Huang},
  {and} \bibinfo{person}{Minghao Hu}.} \bibinfo{year}{2020}\natexlab{b}.
\newblock \showarticletitle{SASGBC: Improving Sequence Labeling Performance for
  Joint Learning of Slot Filling and Intent Detection}. In
  \bibinfo{booktitle}{\emph{Proceedings of 2020 the 6th International
  Conference on Computing and Data Engineering}} (Sanya, China)
  \emph{(\bibinfo{series}{ICCDE 2020})}. \bibinfo{publisher}{Association for
  Computing Machinery}, \bibinfo{address}{New York, NY, USA},
  \bibinfo{pages}{29–33}.
\newblock
\showISBNx{9781450376730}
\urldef\tempurl%
\url{https://doi.org/10.1145/3379247.3379266}
\showDOI{\tempurl}


\bibitem[\protect\citeauthoryear{Wang and Yuan}{Wang and Yuan}{2016}]%
        {wang2016recent}
\bibfield{author}{\bibinfo{person}{Xiaojie Wang} {and} \bibinfo{person}{Caixia
  Yuan}.} \bibinfo{year}{2016}\natexlab{}.
\newblock \showarticletitle{Recent Advances on Human-Computer Dialogue}.
\newblock \bibinfo{journal}{\emph{CAAI Transactions on Intelligence
  Technology}} \bibinfo{volume}{1}, \bibinfo{number}{4} (\bibinfo{date}{Oct}
  \bibinfo{year}{2016}), \bibinfo{pages}{303--312}.
\newblock
\urldef\tempurl%
\url{https://doi.org/10.1016/j.trit.2016.12.004}
\showURL{%
\tempurl}


\bibitem[\protect\citeauthoryear{Wang, Deng, Shen, and Jin}{Wang
  et~al\mbox{.}}{2020a}]%
        {wang2020new}
\bibfield{author}{\bibinfo{person}{Yu Wang}, \bibinfo{person}{Yue Deng},
  \bibinfo{person}{Yilin Shen}, {and} \bibinfo{person}{Hongxia Jin}.}
  \bibinfo{year}{2020}\natexlab{a}.
\newblock \showarticletitle{A New Concept of Multiple Neural Networks Structure
  Using Convex Combination}.
\newblock \bibinfo{journal}{\emph{IEEE Transactions on Neural Networks and
  Learning Systems}}  \bibinfo{volume}{31} (\bibinfo{year}{2020}),
  \bibinfo{pages}{1--12}.
\newblock


\bibitem[\protect\citeauthoryear{Wang, He, Fan, Zhou, and Tu}{Wang
  et~al\mbox{.}}{2019a}]%
        {wang2019effective}
\bibfield{author}{\bibinfo{person}{Yufan Wang}, \bibinfo{person}{Tingting He},
  \bibinfo{person}{Rui Fan}, \bibinfo{person}{Wenji Zhou}, {and}
  \bibinfo{person}{Xinhui Tu}.} \bibinfo{year}{2019}\natexlab{a}.
\newblock \showarticletitle{Effective Utilization of External Knowledge and
  History Context in Multi-turn Spoken Language Understanding Model}. In
  \bibinfo{booktitle}{\emph{2019 IEEE International Conference on Big Data (Big
  Data)}}. \bibinfo{publisher}{{IEEE}}, \bibinfo{address}{Los Angeles, USA},
  \bibinfo{pages}{960--967}.
\newblock


\bibitem[\protect\citeauthoryear{Wang, Huang, He, and Tu}{Wang
  et~al\mbox{.}}{2019b}]%
        {Wang_2019}
\bibfield{author}{\bibinfo{person}{Yufan Wang}, \bibinfo{person}{Jiawei Huang},
  \bibinfo{person}{Tingting He}, {and} \bibinfo{person}{Xinhui Tu}.}
  \bibinfo{year}{2019}\natexlab{b}.
\newblock \showarticletitle{Dialogue intent classification with
  character-{CNN}-{BGRU} networks}.
\newblock \bibinfo{journal}{\emph{Multimedia Tools and Applications}}
  \bibinfo{volume}{79}, \bibinfo{number}{7-8} (\bibinfo{date}{Jun}
  \bibinfo{year}{2019}), \bibinfo{pages}{4553--4572}.
\newblock
\urldef\tempurl%
\url{https://doi.org/10.1007/s11042-019-7678-1}
\showDOI{\tempurl}


\bibitem[\protect\citeauthoryear{Wang, Patel, and Jin}{Wang
  et~al\mbox{.}}{2018a}]%
        {wang2018new}
\bibfield{author}{\bibinfo{person}{Yu Wang}, \bibinfo{person}{Abhishek Patel},
  {and} \bibinfo{person}{Hongxia Jin}.} \bibinfo{year}{2018}\natexlab{a}.
\newblock \showarticletitle{A New Concept of Deep Reinforcement Learning based
  Augmented General Tagging System}. In \bibinfo{booktitle}{\emph{Proceedings
  of the 27th International Conference on Computational Linguistics}}.
  \bibinfo{publisher}{Association for Computational Linguistics},
  \bibinfo{address}{Santa Fe, New Mexico, USA}, \bibinfo{pages}{1683--1693}.
\newblock
\urldef\tempurl%
\url{https://www.aclweb.org/anthology/C18-1143}
\showURL{%
\tempurl}


\bibitem[\protect\citeauthoryear{Wang, Shen, and Jin}{Wang
  et~al\mbox{.}}{2018b}]%
        {wang2018bimodel}
\bibfield{author}{\bibinfo{person}{Yu Wang}, \bibinfo{person}{Yilin Shen},
  {and} \bibinfo{person}{Hongxia Jin}.} \bibinfo{year}{2018}\natexlab{b}.
\newblock \bibinfo{title}{A Bi-model based RNN Semantic Frame Parsing Model for
  Intent Detection and Slot Filling}.
\newblock
\newblock
\showeprint[arxiv]{1812.10235}~[cs.CL]


\bibitem[\protect\citeauthoryear{Wang, Tang, and He}{Wang
  et~al\mbox{.}}{2018c}]%
        {wang2018chinese}
\bibfield{author}{\bibinfo{person}{Yufan Wang}, \bibinfo{person}{Li Tang},
  {and} \bibinfo{person}{Tingting He}.} \bibinfo{year}{2018}\natexlab{c}.
\newblock \showarticletitle{Attention-Based CNN-BLSTM Networks for Joint Intent
  Detection and Slot Filling}. In \bibinfo{booktitle}{\emph{Chinese
  Computational Linguistics and Natural Language Processing Based on Naturally
  Annotated Big Data}}, \bibfield{editor}{\bibinfo{person}{Maosong Sun},
  \bibinfo{person}{Ting Liu}, \bibinfo{person}{Xiaojie Wang},
  \bibinfo{person}{Zhiyuan Liu}, {and} \bibinfo{person}{Yang Liu}} (Eds.).
  \bibinfo{publisher}{Springer International Publishing},
  \bibinfo{address}{Cham}, \bibinfo{pages}{250--261}.
\newblock
\showISBNx{978-3-030-01716-3}


\bibitem[\protect\citeauthoryear{Wang}{Wang}{2010}]%
        {wang2010strategies}
\bibfield{author}{\bibinfo{person}{Ye-Yi Wang}.}
  \bibinfo{year}{2010}\natexlab{}.
\newblock \showarticletitle{Strategies for statistical spoken language
  understanding with small amount of data - an empirical study}. In
  \bibinfo{booktitle}{\emph{Interspeech}}. \bibinfo{publisher}{{ISCA}},
  \bibinfo{address}{Makuhari, Japan}, \bibinfo{pages}{2498--2501}.
\newblock


\bibitem[\protect\citeauthoryear{Wang, Deng, and Acero}{Wang
  et~al\mbox{.}}{2005}]%
        {wang2005spoken}
\bibfield{author}{\bibinfo{person}{Ye-Yi Wang}, \bibinfo{person}{Li Deng},
  {and} \bibinfo{person}{Alex Acero}.} \bibinfo{year}{2005}\natexlab{}.
\newblock \showarticletitle{Spoken Language Understanding: An Introduction to
  the Statistical Framework}.
\newblock \bibinfo{journal}{\emph{IEEE Signal Processing Magazine}}
  \bibinfo{volume}{22}, \bibinfo{number}{5} (\bibinfo{date}{January}
  \bibinfo{year}{2005}), \bibinfo{pages}{16--31}.
\newblock
\urldef\tempurl%
\url{https://www.microsoft.com/en-us/research/publication/spoken-language-understanding-an-introduction-to-the-statistical-framework/}
\showURL{%
\tempurl}


\bibitem[\protect\citeauthoryear{Wen, Wang, Dong, and Chen}{Wen
  et~al\mbox{.}}{2018}]%
        {wen2018jointly}
\bibfield{author}{\bibinfo{person}{Liyun Wen}, \bibinfo{person}{Xiaojie Wang},
  \bibinfo{person}{Zhenjiang Dong}, {and} \bibinfo{person}{Hong Chen}.}
  \bibinfo{year}{2018}\natexlab{}.
\newblock \showarticletitle{Jointly Modeling Intent Identification and Slot
  Filling with Contextual and Hierarchical Information}. In
  \bibinfo{booktitle}{\emph{Natural Language Processing and Chinese
  Computing}}, \bibfield{editor}{\bibinfo{person}{Xuanjing Huang},
  \bibinfo{person}{Jing Jiang}, \bibinfo{person}{Dongyan Zhao},
  \bibinfo{person}{Yansong Feng}, {and} \bibinfo{person}{Yu~Hong}} (Eds.).
  \bibinfo{publisher}{Springer International Publishing},
  \bibinfo{address}{Cham}, \bibinfo{pages}{3--15}.
\newblock
\showISBNx{978-3-319-73618-1}


\bibitem[\protect\citeauthoryear{Xia, Zhang, Yan, Chang, and Yu}{Xia
  et~al\mbox{.}}{2018}]%
        {Xia_2018}
\bibfield{author}{\bibinfo{person}{Congying Xia}, \bibinfo{person}{Chenwei
  Zhang}, \bibinfo{person}{Xiaohui Yan}, \bibinfo{person}{Yi Chang}, {and}
  \bibinfo{person}{Philip Yu}.} \bibinfo{year}{2018}\natexlab{}.
\newblock \showarticletitle{Zero-shot User Intent Detection via Capsule Neural
  Networks}. In \bibinfo{booktitle}{\emph{Proceedings of the 2018 Conference on
  Empirical Methods in Natural Language Processing}}.
  \bibinfo{publisher}{Association for Computational Linguistics},
  \bibinfo{address}{Brussels, Belgium}, \bibinfo{pages}{3090--3099}.
\newblock
\urldef\tempurl%
\url{https://doi.org/10.18653/v1/d18-1348}
\showDOI{\tempurl}


\bibitem[\protect\citeauthoryear{Xie, Gao, Ding, and Hao}{Xie
  et~al\mbox{.}}{2018}]%
        {Xie_2018}
\bibfield{author}{\bibinfo{person}{Wenxiu Xie}, \bibinfo{person}{Dongfa Gao},
  \bibinfo{person}{Ruoyao Ding}, {and} \bibinfo{person}{Tianyong Hao}.}
  \bibinfo{year}{2018}\natexlab{}.
\newblock \showarticletitle{A Feature-Enriched Method for User Intent
  Classification by Leveraging Semantic Tag Expansion}.
\newblock In \bibinfo{booktitle}{\emph{Natural Language Processing and Chinese
  Computing}}. \bibinfo{publisher}{Springer International Publishing},
  \bibinfo{address}{Cham}, \bibinfo{pages}{224--234}.
\newblock
\urldef\tempurl%
\url{https://doi.org/10.1007/978-3-319-99501-4_19}
\showDOI{\tempurl}


\bibitem[\protect\citeauthoryear{Xu, Li, Zhang, Cui, Sun, and Zhou}{Xu
  et~al\mbox{.}}{2020}]%
        {xu2020model}
\bibfield{author}{\bibinfo{person}{Cong Xu}, \bibinfo{person}{Qing Li},
  \bibinfo{person}{Dezheng Zhang}, \bibinfo{person}{Jiarui Cui},
  \bibinfo{person}{Zhenqi Sun}, {and} \bibinfo{person}{Hao Zhou}.}
  \bibinfo{year}{2020}\natexlab{}.
\newblock \showarticletitle{A model with length-variable attention for spoken
  language understanding}.
\newblock \bibinfo{journal}{\emph{Neurocomputing}}  \bibinfo{volume}{379}
  (\bibinfo{year}{2020}), \bibinfo{pages}{197--202}.
\newblock
\showISBNx{0925-2312}
\urldef\tempurl%
\url{https://doi.org/10.1016/j.neucom.2019.07.112}
\showDOI{\tempurl}


\bibitem[\protect\citeauthoryear{{Xu} and {Sarikaya}}{{Xu} and
  {Sarikaya}}{2013}]%
        {xu2013convolutional}
\bibfield{author}{\bibinfo{person}{Puyang {Xu}} {and} \bibinfo{person}{Ruhi
  {Sarikaya}}.} \bibinfo{year}{2013}\natexlab{}.
\newblock \showarticletitle{Convolutional neural network based triangular CRF
  for joint intent detection and slot filling}. In
  \bibinfo{booktitle}{\emph{2013 IEEE Workshop on Automatic Speech Recognition
  and Understanding}}. \bibinfo{publisher}{{IEEE}},
  \bibinfo{address}{Vancouver, Canada}, \bibinfo{pages}{78--83}.
\newblock


\bibitem[\protect\citeauthoryear{Xu and Sarikaya}{Xu and Sarikaya}{2013}]%
        {xu2013exploiting}
\bibfield{author}{\bibinfo{person}{Puyang Xu} {and} \bibinfo{person}{Ruhi
  Sarikaya}.} \bibinfo{year}{2013}\natexlab{}.
\newblock \showarticletitle{Exploiting Shared Information for Multi-Intent
  Natural Language Sentence Classification}. In
  \bibinfo{booktitle}{\emph{Interspeech}}. \bibinfo{publisher}{ISCA},
  \bibinfo{address}{Lyon, France}, \bibinfo{pages}{3785--3789}.
\newblock


\bibitem[\protect\citeauthoryear{{Yang}, {Chen}, {Hakkani-Tür}, {Crook}, {Li},
  {Gao}, and {Deng}}{{Yang} et~al\mbox{.}}{2017}]%
        {yang2017endtoend}
\bibfield{author}{\bibinfo{person}{Xuesong {Yang}}, \bibinfo{person}{Yun-Nung
  {Chen}}, \bibinfo{person}{Dilek {Hakkani-Tür}}, \bibinfo{person}{Paul
  {Crook}}, \bibinfo{person}{Xiujun {Li}}, \bibinfo{person}{Jianfeng {Gao}},
  {and} \bibinfo{person}{Li {Deng}}.} \bibinfo{year}{2017}\natexlab{}.
\newblock \showarticletitle{End-to-end joint learning of natural language
  understanding and dialogue manager}. In \bibinfo{booktitle}{\emph{2017 IEEE
  International Conference on Acoustics, Speech and Signal Processing
  (ICASSP)}}. \bibinfo{publisher}{{IEEE}}, \bibinfo{address}{New Orleans, USA},
  \bibinfo{pages}{5690--5694}.
\newblock


\bibitem[\protect\citeauthoryear{{Yao}, {Peng}, {Zhang}, {Yu}, {Zweig}, and
  {Shi}}{{Yao} et~al\mbox{.}}{2014a}]%
        {yao2014spoken}
\bibfield{author}{\bibinfo{person}{Kaisheng {Yao}}, \bibinfo{person}{Baolin
  {Peng}}, \bibinfo{person}{Yu {Zhang}}, \bibinfo{person}{Dong {Yu}},
  \bibinfo{person}{Geoffrey {Zweig}}, {and} \bibinfo{person}{Yangyang {Shi}}.}
  \bibinfo{year}{2014}\natexlab{a}.
\newblock \showarticletitle{Spoken language understanding using long short-term
  memory neural networks}. In \bibinfo{booktitle}{\emph{2014 IEEE Spoken
  Language Technology Workshop (SLT)}}. \bibinfo{publisher}{{IEEE}},
  \bibinfo{address}{South Lake Tahoe, USA}, \bibinfo{pages}{189--194}.
\newblock
\showISSN{null}
\urldef\tempurl%
\url{https://doi.org/10.1109/SLT.2014.7078572}
\showDOI{\tempurl}


\bibitem[\protect\citeauthoryear{{Yao}, {Peng}, {Zweig}, {Yu}, {Li}, and
  {Gao}}{{Yao} et~al\mbox{.}}{2014b}]%
        {yao2014recurrent}
\bibfield{author}{\bibinfo{person}{Kaisheng {Yao}}, \bibinfo{person}{Baolin
  {Peng}}, \bibinfo{person}{Geoffrey {Zweig}}, \bibinfo{person}{Dong {Yu}},
  \bibinfo{person}{Xiaolong {Li}}, {and} \bibinfo{person}{Feng {Gao}}.}
  \bibinfo{year}{2014}\natexlab{b}.
\newblock \showarticletitle{Recurrent conditional random field for language
  understanding}. In \bibinfo{booktitle}{\emph{2014 IEEE International
  Conference on Acoustics, Speech and Signal Processing (ICASSP)}}.
  \bibinfo{publisher}{{IEEE}}, \bibinfo{address}{Florence, Italy},
  \bibinfo{pages}{4077--4081}.
\newblock


\bibitem[\protect\citeauthoryear{Yao, Zweig, Hwang, Shi, and Yu}{Yao
  et~al\mbox{.}}{2013}]%
        {yao2013recurrent}
\bibfield{author}{\bibinfo{person}{Kaisheng Yao}, \bibinfo{person}{Geoffrey
  Zweig}, \bibinfo{person}{Mei-Yuh Hwang}, \bibinfo{person}{Yangyang Shi},
  {and} \bibinfo{person}{Dong Yu}.} \bibinfo{year}{2013}\natexlab{}.
\newblock \showarticletitle{Recurrent Neural Networks for Language
  Understanding}. In \bibinfo{booktitle}{\emph{Interspeech}}.
  \bibinfo{publisher}{ISCA}, \bibinfo{address}{Lyon, France},
  \bibinfo{pages}{2524--2528}.
\newblock
\urldef\tempurl%
\url{https://doi.org/10.13140/2.1.2755.3285}
\showDOI{\tempurl}


\bibitem[\protect\citeauthoryear{Yilmaz and Toraman}{Yilmaz and
  Toraman}{2020}]%
        {yilmaz2020kloos}
\bibfield{author}{\bibinfo{person}{Eyup~Halit Yilmaz} {and}
  \bibinfo{person}{Cagri Toraman}.} \bibinfo{year}{2020}\natexlab{}.
\newblock \showarticletitle{KLOOS: KL Divergence-Based Out-of-Scope Intent
  Detection in Human-to-Machine Conversations}. In
  \bibinfo{booktitle}{\emph{Proceedings of the 43rd International ACM SIGIR
  Conference on Research and Development in Information Retrieval}} (Virtual
  Event, China) \emph{(\bibinfo{series}{SIGIR ’20})}.
  \bibinfo{publisher}{Association for Computing Machinery},
  \bibinfo{address}{New York, NY, USA}, \bibinfo{pages}{2105–2108}.
\newblock
\showISBNx{9781450380164}
\urldef\tempurl%
\url{https://doi.org/10.1145/3397271.3401318}
\showDOI{\tempurl}


\bibitem[\protect\citeauthoryear{Yu, Wang, and Deng}{Yu et~al\mbox{.}}{2011}]%
        {yu2011sequence}
\bibfield{author}{\bibinfo{person}{Dong Yu}, \bibinfo{person}{Shizhen Wang},
  {and} \bibinfo{person}{li Deng}.} \bibinfo{year}{2011}\natexlab{}.
\newblock \showarticletitle{Sequential Labeling Using Deep-Structured
  Conditional Random Fields}.
\newblock \bibinfo{journal}{\emph{Selected Topics in Signal Processing, IEEE
  Journal of}}  \bibinfo{volume}{4} (\bibinfo{date}{01} \bibinfo{year}{2011}),
  \bibinfo{pages}{965 -- 973}.
\newblock
\urldef\tempurl%
\url{https://doi.org/10.1109/JSTSP.2010.2075990}
\showDOI{\tempurl}


\bibitem[\protect\citeauthoryear{{Yu}, {Shen}, {Zhu}, and {Chen}}{{Yu}
  et~al\mbox{.}}{2018}]%
        {yu2018novel}
\bibfield{author}{\bibinfo{person}{Shuai {Yu}}, \bibinfo{person}{Lei {Shen}},
  \bibinfo{person}{Pengcheng {Zhu}}, {and} \bibinfo{person}{Jiansong {Chen}}.}
  \bibinfo{year}{2018}\natexlab{}.
\newblock \showarticletitle{ACJIS: A Novel Attentive Cross Approach For Joint
  Intent Detection And Slot Filling}. In \bibinfo{booktitle}{\emph{2018
  International Joint Conference on Neural Networks (IJCNN)}}.
  \bibinfo{publisher}{{IEEE}}, \bibinfo{address}{Rio de Janeiro, Brazil},
  \bibinfo{pages}{1--7}.
\newblock


\bibitem[\protect\citeauthoryear{{Yulan He} and {Young}}{{Yulan He} and
  {Young}}{2003}]%
        {he2003datadriven}
\bibfield{author}{\bibinfo{person}{{Yulan He}} {and} \bibinfo{person}{Steve
  {Young}}.} \bibinfo{year}{2003}\natexlab{}.
\newblock \showarticletitle{A data-driven spoken language understanding
  system}. In \bibinfo{booktitle}{\emph{2003 IEEE Workshop on Automatic Speech
  Recognition and Understanding (IEEE Cat. No.03EX721)}}.
  \bibinfo{publisher}{IEEE}, \bibinfo{address}{Piscataway, USA},
  \bibinfo{pages}{583--588}.
\newblock


\bibitem[\protect\citeauthoryear{Zhang, Fan, Du, and Yu}{Zhang
  et~al\mbox{.}}{2016}]%
        {zhang2016mining}
\bibfield{author}{\bibinfo{person}{Chenwei Zhang}, \bibinfo{person}{Wei Fan},
  \bibinfo{person}{Nan Du}, {and} \bibinfo{person}{Philip~S. Yu}.}
  \bibinfo{year}{2016}\natexlab{}.
\newblock \showarticletitle{Mining User Intentions from Medical Queries: A
  Neural Network Based Heterogeneous Jointly Modeling Approach}. In
  \bibinfo{booktitle}{\emph{Proceedings of the 25th International Conference on
  World Wide Web}} (Montr\'{e}al, Qu\'{e}bec, Canada)
  \emph{(\bibinfo{series}{WWW ’16})}. \bibinfo{publisher}{International World
  Wide Web Conferences Steering Committee}, \bibinfo{address}{Republic and
  Canton of Geneva, CHE}, \bibinfo{pages}{1373–1384}.
\newblock
\showISBNx{9781450341431}
\urldef\tempurl%
\url{https://doi.org/10.1145/2872427.2874810}
\showDOI{\tempurl}


\bibitem[\protect\citeauthoryear{Zhang, Li, Du, Fan, and Yu}{Zhang
  et~al\mbox{.}}{2019a}]%
        {zhang2019capsule}
\bibfield{author}{\bibinfo{person}{Chenwei Zhang}, \bibinfo{person}{Yaliang
  Li}, \bibinfo{person}{Nan Du}, \bibinfo{person}{Wei Fan}, {and}
  \bibinfo{person}{Philip Yu}.} \bibinfo{year}{2019}\natexlab{a}.
\newblock \showarticletitle{Joint Slot Filling and Intent Detection via Capsule
  Neural Networks}. In \bibinfo{booktitle}{\emph{Proceedings of the 57th Annual
  Meeting of the Association for Computational Linguistics}}.
  \bibinfo{publisher}{Association for Computational Linguistics},
  \bibinfo{address}{Florence, Italy}, \bibinfo{pages}{5259--5267}.
\newblock
\urldef\tempurl%
\url{https://doi.org/10.18653/v1/P19-1519}
\showDOI{\tempurl}


\bibitem[\protect\citeauthoryear{Zhang, Fang, Cao, Liu, Chen, and Tan}{Zhang
  et~al\mbox{.}}{2018}]%
        {zhang2018attention}
\bibfield{author}{\bibinfo{person}{Dongjie Zhang}, \bibinfo{person}{Zheng
  Fang}, \bibinfo{person}{Yanan Cao}, \bibinfo{person}{Yanbing Liu},
  \bibinfo{person}{Xiaojun Chen}, {and} \bibinfo{person}{Jianlong Tan}.}
  \bibinfo{year}{2018}\natexlab{}.
\newblock \showarticletitle{Attention-Based RNN Model for Joint Extraction of
  Intent and Word Slot Based on a Tagging Strategy}. In
  \bibinfo{booktitle}{\emph{Artificial Neural Networks and Machine Learning --
  ICANN 2018}}, \bibfield{editor}{\bibinfo{person}{V{\v{e}}ra
  K{\r{u}}rkov{\'a}}, \bibinfo{person}{Yannis Manolopoulos},
  \bibinfo{person}{Barbara Hammer}, \bibinfo{person}{Lazaros Iliadis}, {and}
  \bibinfo{person}{Ilias Maglogiannis}} (Eds.). \bibinfo{publisher}{Springer
  International Publishing}, \bibinfo{address}{Cham},
  \bibinfo{pages}{178--188}.
\newblock
\showISBNx{978-3-030-01424-7}


\bibitem[\protect\citeauthoryear{Zhang, Ma, Zhang, Yan, and Wang}{Zhang
  et~al\mbox{.}}{2020b}]%
        {zhang2020graph}
\bibfield{author}{\bibinfo{person}{Linhao Zhang}, \bibinfo{person}{Dehong Ma},
  \bibinfo{person}{Xiaodong Zhang}, \bibinfo{person}{Xiaohui Yan}, {and}
  \bibinfo{person}{Hou-Feng Wang}.} \bibinfo{year}{2020}\natexlab{b}.
\newblock \showarticletitle{Graph LSTM with Context-Gated Mechanism for Spoken
  Language Understanding}. In \bibinfo{booktitle}{\emph{AAAI 2020}}.
  \bibinfo{publisher}{AAAI Press}, \bibinfo{address}{New York, USA}.
\newblock


\bibitem[\protect\citeauthoryear{Zhang and Wang}{Zhang and Wang}{2019}]%
        {zhang2019using}
\bibfield{author}{\bibinfo{person}{Linhao Zhang} {and} \bibinfo{person}{Houfeng
  Wang}.} \bibinfo{year}{2019}\natexlab{}.
\newblock \showarticletitle{Using Bidirectional Transformer-CRF for Spoken
  Language Understanding}. In \bibinfo{booktitle}{\emph{Natural Language
  Processing and Chinese Computing}}, \bibfield{editor}{\bibinfo{person}{Jie
  Tang}, \bibinfo{person}{Min-Yen Kan}, \bibinfo{person}{Dongyan Zhao},
  \bibinfo{person}{Sujian Li}, {and} \bibinfo{person}{Hongying Zan}} (Eds.).
  \bibinfo{publisher}{Springer International Publishing},
  \bibinfo{address}{Cham}, \bibinfo{pages}{130--141}.
\newblock
\showISBNx{978-3-030-32233-5}


\bibitem[\protect\citeauthoryear{{Zhang}, {Jiang}, {He}, {Zhao}, and
  {Fang}}{{Zhang} et~al\mbox{.}}{2019}]%
        {zhang2019novel}
\bibfield{author}{\bibinfo{person}{Shuyou {Zhang}}, \bibinfo{person}{Junjie
  {Jiang}}, \bibinfo{person}{Zaixing {He}}, \bibinfo{person}{Xinyue {Zhao}},
  {and} \bibinfo{person}{Jinhui {Fang}}.} \bibinfo{year}{2019}\natexlab{}.
\newblock \showarticletitle{A Novel Slot-Gated Model Combined With a Key Verb
  Context Feature for Task Request Understanding by Service Robots}.
\newblock \bibinfo{journal}{\emph{IEEE Access}}  \bibinfo{volume}{7}
  (\bibinfo{year}{2019}), \bibinfo{pages}{105937--105947}.
\newblock


\bibitem[\protect\citeauthoryear{Zhang and Wang}{Zhang and Wang}{2016}]%
        {zhang2016joint}
\bibfield{author}{\bibinfo{person}{Xiaodong Zhang} {and}
  \bibinfo{person}{Houfeng Wang}.} \bibinfo{year}{2016}\natexlab{}.
\newblock \showarticletitle{A Joint Model of Intent Determination and Slot
  Filling for Spoken Language Understanding}. In
  \bibinfo{booktitle}{\emph{Proceedings of the Twenty-Fifth International Joint
  Conference on Artificial Intelligence (IJCAI-16)}}. \bibinfo{publisher}{AAAI
  Press}, \bibinfo{address}{New York, USA}, \bibinfo{pages}{2993--2999}.
\newblock


\bibitem[\protect\citeauthoryear{Zhang, Huang, and Wang}{Zhang
  et~al\mbox{.}}{2020a}]%
        {zhang2020deeptime}
\bibfield{author}{\bibinfo{person}{Zhen Zhang}, \bibinfo{person}{Hao Huang},
  {and} \bibinfo{person}{Kai Wang}.} \bibinfo{year}{2020}\natexlab{a}.
\newblock \showarticletitle{Using Deep Time Delay Neural Network for Slot
  Filling in Spoken Language Understanding}.
\newblock \bibinfo{journal}{\emph{Symmetry}} \bibinfo{volume}{12},
  \bibinfo{number}{6} (\bibinfo{date}{06} \bibinfo{year}{2020}).
\newblock


\bibitem[\protect\citeauthoryear{Zhang, Zhang, Chen, and Zhang}{Zhang
  et~al\mbox{.}}{2019b}]%
        {zhang2019joint}
\bibfield{author}{\bibinfo{person}{Zhichang Zhang}, \bibinfo{person}{Zhenwen
  Zhang}, \bibinfo{person}{Haoyuan Chen}, {and} \bibinfo{person}{Zhiman
  Zhang}.} \bibinfo{year}{2019}\natexlab{b}.
\newblock \showarticletitle{A Joint Learning Framework With BERT for Spoken
  Language Understanding}.
\newblock \bibinfo{journal}{\emph{IEEE Access}}  \bibinfo{volume}{7}
  (\bibinfo{year}{2019}), \bibinfo{pages}{168849--168858}.
\newblock


\bibitem[\protect\citeauthoryear{Zhao and Feng}{Zhao and Feng}{2018}]%
        {zhao2018improving}
\bibfield{author}{\bibinfo{person}{Lin Zhao} {and} \bibinfo{person}{Zhe Feng}.}
  \bibinfo{year}{2018}\natexlab{}.
\newblock \showarticletitle{Improving Slot Filling in Spoken Language
  Understanding with Joint Pointer and Attention}. In
  \bibinfo{booktitle}{\emph{Proceedings of the 56th Annual Meeting of the
  Association for Computational Linguistics (Volume 2: Short Papers)}}.
  \bibinfo{publisher}{Association for Computational Linguistics},
  \bibinfo{address}{Melbourne, Australia}, \bibinfo{pages}{426--431}.
\newblock
\urldef\tempurl%
\url{https://doi.org/10.18653/v1/P18-2068}
\showDOI{\tempurl}


\bibitem[\protect\citeauthoryear{Zhao, E, and Song}{Zhao et~al\mbox{.}}{2018}]%
        {zhao2018joint}
\bibfield{author}{\bibinfo{person}{Xinlu Zhao}, \bibinfo{person}{Haihong E},
  {and} \bibinfo{person}{Meina Song}.} \bibinfo{year}{2018}\natexlab{}.
\newblock \showarticletitle{A Joint Model based on CNN-LSTMs in Dialogue
  Understanding}. In \bibinfo{booktitle}{\emph{2018 International Conference on
  Information Systems and Computer Aided Education (ICISCAE)}}.
  \bibinfo{publisher}{IEEE}, \bibinfo{address}{Piscataway, USA},
  \bibinfo{pages}{471--475}.
\newblock


\bibitem[\protect\citeauthoryear{{Zheng}, {Liu}, and {Hansen}}{{Zheng}
  et~al\mbox{.}}{2017}]%
        {zheng2017intent}
\bibfield{author}{\bibinfo{person}{Yang {Zheng}}, \bibinfo{person}{Yongkang
  {Liu}}, {and} \bibinfo{person}{John H.~L. {Hansen}}.}
  \bibinfo{year}{2017}\natexlab{}.
\newblock \showarticletitle{Intent detection and semantic parsing for
  navigation dialogue language processing}. In \bibinfo{booktitle}{\emph{2017
  IEEE 20th International Conference on Intelligent Transportation Systems
  (ITSC)}}. \bibinfo{publisher}{{IEEE}}, \bibinfo{address}{Yokohama, Japan},
  \bibinfo{pages}{1--6}.
\newblock


\bibitem[\protect\citeauthoryear{Zhou, Wen, Wang, Ma, and Wang}{Zhou
  et~al\mbox{.}}{2016}]%
        {zhou2016hierarchical}
\bibfield{author}{\bibinfo{person}{Qianrong Zhou}, \bibinfo{person}{Liyun Wen},
  \bibinfo{person}{Xiaojie Wang}, \bibinfo{person}{Long Ma}, {and}
  \bibinfo{person}{Yue Wang}.} \bibinfo{year}{2016}\natexlab{}.
\newblock \showarticletitle{A Hierarchical LSTM Model for Joint Tasks}. In
  \bibinfo{booktitle}{\emph{Chinese Computational Linguistics and Natural
  Language Processing Based on Naturally Annotated Big Data}},
  \bibfield{editor}{\bibinfo{person}{Maosong Sun}, \bibinfo{person}{Xuanjing
  Huang}, \bibinfo{person}{Hongfei Lin}, \bibinfo{person}{Zhiyuan Liu}, {and}
  \bibinfo{person}{Yang Liu}} (Eds.). \bibinfo{publisher}{Springer
  International Publishing}, \bibinfo{address}{Cham},
  \bibinfo{pages}{324--335}.
\newblock
\showISBNx{978-3-319-47674-2}


\bibitem[\protect\citeauthoryear{Zhu and Yu}{Zhu and Yu}{2017}]%
        {zhu2017encoder}
\bibfield{author}{\bibinfo{person}{Su Zhu} {and} \bibinfo{person}{Kai Yu}.}
  \bibinfo{year}{2017}\natexlab{}.
\newblock \showarticletitle{Encoder-decoder with focus-mechanism for sequence
  labelling based spoken language understanding}. In
  \bibinfo{booktitle}{\emph{2017 IEEE International Conference on Acoustics,
  Speech and Signal Processing (ICASSP)}}. \bibinfo{publisher}{{IEEE}},
  \bibinfo{address}{New Orleans, USA}, \bibinfo{pages}{5675--5679}.
\newblock
\urldef\tempurl%
\url{https://doi.org/10.1109/ICASSP.2017.7953243}
\showDOI{\tempurl}


\bibitem[\protect\citeauthoryear{Zhu, Zhao, Ma, and Yu}{Zhu
  et~al\mbox{.}}{2020}]%
        {zhu2020prior}
\bibfield{author}{\bibinfo{person}{Su Zhu}, \bibinfo{person}{Zijian Zhao},
  \bibinfo{person}{Rao Ma}, {and} \bibinfo{person}{Kai Yu}.}
  \bibinfo{year}{2020}\natexlab{}.
\newblock \showarticletitle{Prior Knowledge Driven Label Embedding for Slot
  Filling in Natural Language Understanding}.
\newblock \bibinfo{journal}{\emph{IEEE/ACM Transactions on Audio, Speech, and
  Language Processing}}  \bibinfo{volume}{PP} (\bibinfo{date}{03}
  \bibinfo{year}{2020}), \bibinfo{pages}{1--1}.
\newblock
\urldef\tempurl%
\url{https://doi.org/10.1109/TASLP.2020.2980152}
\showDOI{\tempurl}


\end{thebibliography}


\end{document}